\documentclass{adobe_research}

\usepackage[T1]{fontenc}
\usepackage{lmodern}

\usepackage{amsmath,amssymb,amsfonts}

\usepackage{makecell}
\usepackage{tabularx}
\usepackage{color, colortbl}

\usepackage{wrapfig}
\usepackage{placeins}

\usepackage{algorithm}
\usepackage{algpseudocode}

\usepackage{pifont}
\usepackage{xspace}

\usepackage{tikz}
\usetikzlibrary{positioning,calc,patterns,arrows.meta,decorations.pathreplacing,backgrounds,shapes.geometric}
\usepackage{float}

\newcommand{\Lflare}{\mathcal{L}_{\mathrm{FLARE}}}

\usepackage{xspace}
\makeatletter
\DeclareRobustCommand\onedot{\futurelet\@let@token\@onedot}
\def\@onedot{\ifx\@let@token.\else.\null\fi\xspace}

\makeatother

\usepackage{newtxtext}

\newcommand{\cmark}{\ding{51}}
\newcommand{\xmark}{\ding{55}}

\definecolor{axiscol}{HTML}{5E504D}
\definecolor{noteline}{HTML}{E7BDB8}
\definecolor{taskshade}{HTML}{F3F3F3}

\newcommand{\model}{\textcolor{adobered}{\textbf{FLARE}}\xspace}

\title{\model: Diffusion for Hybrid Language Model}
\author[1,2,*]{Yuchen Zhu}
\author[1,*]{Jing Shi}
\author[1]{Chongjian Ge}
\author[1]{Hao Tan}
\author[1]{Yiran Xu}
\author[1]{Wanrong Zhu}
\author[1]{Jason Kuen}
\author[1]{Koustava Goswami}
\author[1]{Rajiv Jain}
\author[2]{Yongxin Chen}
\author[2]{Molei Tao}
\author[1,*\dagger]{Jiuxiang Gu}

\affiliation[1]{\textbf{Adobe Research}}
\affiliation[2]{\textbf{Georgia Institute of Technology}}

\contribution[*]{Core Contributors}
\contribution[\dagger]{Project Lead.}

\abstract{

Autoregressive (AR) large language models (LLMs) have achieved broad practical success, but sequential decoding remains a key bottleneck for low-latency deployment. Recent efficient-inference work has progressed along two axes: reducing the cost of each model invocation through efficient architectures, and reducing serial decoding steps through parallel generation. Hybrid attention backbones address the former, while diffusion language models (dLLMs) pursue the latter via iterative parallel denoising. Combining these advantages remains challenging: AR-to-dLLM conversion often fails to preserve seed-checkpoint capability, and hybrid-attention recurrent states and masking constraints make diffusion training and serving nontrivial. We present \model, a systematic conversion framework for hybrid-attention LLMs. Our analysis identifies transfer data quality as the primary determinant of capability preservation, outweighing loss formulation and attention-mask design. The resulting framework combines a token-equal AR-and-diffusion objective, hardware-aware kernels, and unified inference, enabling one checkpoint to support both AR-style verified decoding and diffusion-style parallel denoising. Starting from strong AR checkpoints with limited post-training data, \model is competitive with leading open-source dLLMs across model scales and delivers consistent throughput gains over open-source dLLM baselines in single-GPU concurrent serving. Our results further suggest that practical dLLMs are limited not only by decoding algorithms, but also by transfer data quality and the training inefficiency of current block-diffusion objectives, motivating joint design of data, objectives, architectures, and inference systems.

}

\date{June 2, 2026}

\adobedata[Project Page]{\url{https://tokflare.github.io}}
\adobedata[Code]{\url{https://github.com/yuchen-zhu-zyc/FLARE}}
\adobedata[Checkpoints]{\url{https://huggingface.co/collections/yuchen-zhu-zyc/flare}}

\begin{document}

\maketitle

\section{Introduction}
\label{sec:introduction}

Modern large language models (LLMs) \citep{singh2025openai, anthropic2026claudeopus47, google2026gemini31pro} have become increasingly central to interactive, agentic, and embodied AI systems, spanning general-purpose assistants, personal agents \citep{openclaw2026openclaw}, embodied AI, robotics control, and autonomous driving \citep{jiang2026fast, yang2026realtime, jiang2025survey, ma2025running}. As these models move toward end-user, edge-device, and real-time service deployment, inference must preserve model capability while satisfying system constraints on latency, throughput, memory footprint, and energy consumption \citep{zheng2024sglang, kwon2023efficient, ggml2026llamacpp}. These constraints are especially pronounced in latency-sensitive applications and directly affect deployability on personal devices, edge hardware, and closed-loop decision-making systems. Consequently, serving efficiency and inference latency have become central bottlenecks for practical AR LLM deployment.

\begin{figure}[!t]
\centering
\vspace{-2em}

\includegraphics[width=1\linewidth]{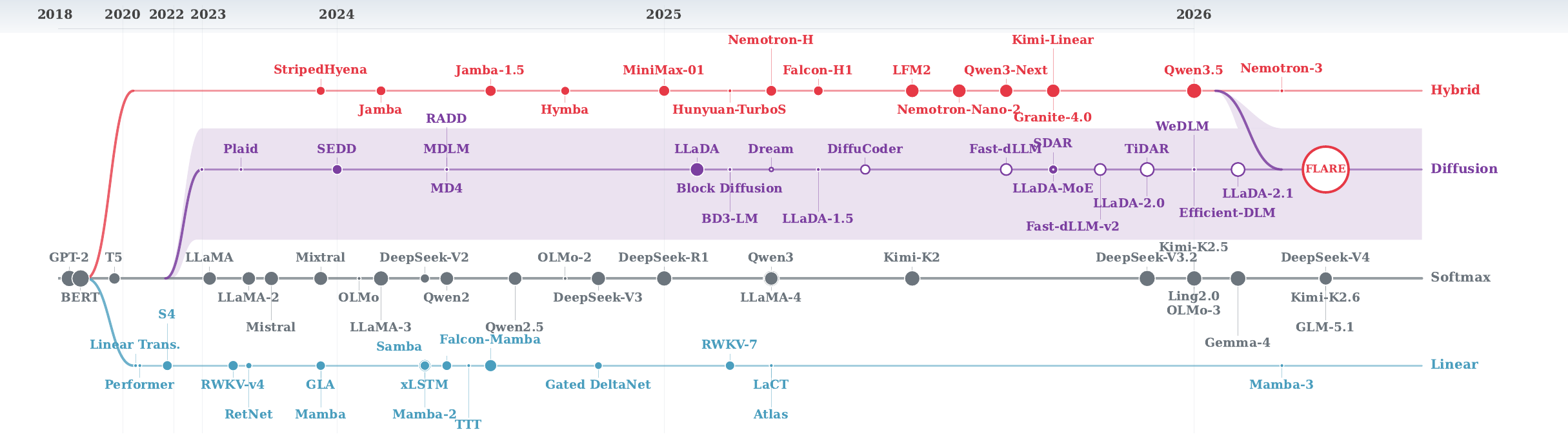}
\caption{Timeline of open-source LLM architectures. \model is a fast and competent LLM as a confluence of algorithmic efficiency (diffusion) and architectural efficiency (hybrid attention). }\label{fig:lm-timeline}
\vspace{-1em}
\end{figure}

Efforts to improve LLM serving efficiency have proceeded along two complementary directions: making each model invocation cheaper, and reducing the number of serial invocations. The first direction relies on architectural changes. Compressed or sparse variants of softmax attention \citep{vaswani2017attention, ainslie2023gqa, liu2024deepseek, liu2025deepseek} and linear-attention or related recurrent-memory modules \citep{katharopoulos2020transformers, schlag2021linear, gu2023mamba, dao2024transformers, yang2024gated} reduce the storage and computation required by long-context and incremental decoding. Since purely linear memory can trade off retrieval fidelity, recent efficient LLMs increasingly use hybrid-attention backbones that combine softmax and linear-attention layers, balancing model capability, cache cost, and serving efficiency \citep{li2025minimax, team2025kimi, basant2025nvidia, blakeman2025nemotron, blakeman2025nvidia, qwen2025qwen3next, qwen2026qwen35}. Hybrid attention lowers per-forward cost, but does not remove the serial nature of AR generation. The second direction reduces serial decoding steps. Multi-token prediction and speculative decoding retain the AR factorization while advancing multiple tokens per round through auxiliary heads, drafting, or parallel verification \citep{gloeckle2024better, li2025eagle, cai2024medusa, leviathan2023fast, kumar2026speculative}. Diffusion large language models (dLLMs) go further by relaxing strict next-token generation and using iterative parallel denoising, offering a distinct route to lower decoding latency \citep{khanna2025mercury, song2025seed, googledeepmind2025geminidiffusion, wang2025diffusion, chen2026dflash}. As illustrated in Figure~\ref{fig:lm-timeline}, these trends have largely evolved in parallel in the open-source LLM ecosystem, with hybrid-attention AR models improving architectural efficiency and dLLMs improving generation-side parallelism.

Research on dLLMs has advanced diffusion-style objectives, masked or uniform diffusion, block diffusion, and large-scale diffusion LMs \citep{lou2023discrete, sahoo2024simple, shi2024simplified, ou2024your, arriola2025block, sahoo2025diffusion, nie2025large, bie2025llada2}, with recent studies beginning to characterize their scaling and data-efficiency behavior \citep{ni2025training, sahoo2026scaling, ni2025diffusion}. In parallel, AR-initialized transfer and related conversion recipes obtain parallel-generation capability from existing AR checkpoints through continued training, shifted-token losses, block-diffusion training, or lightweight post-training \citep{gong2024scaling, fu2025efficient, cheng2025sdar, liu2025tidar, liu2025wedlm, yu2026introspective}. Nevertheless, AR-to-dLLM transfer remains at an early stage: the roles of \emph{transfer-stage data}, \emph{scaling behavior}, and \emph{joint loss--mask design} in capability preservation remain poorly understood. As a result, existing conversion methods often struggle to robustly inherit the capabilities of their AR seed checkpoints. At the same time, higher tokens-per-forward does not automatically translate into higher tokens-per-second; realized serving throughput depends on inference-system support for prefix reuse, KV or recurrent-state caching, and low-overhead parallel decoding \citep{wu2025fast, fu2025efficient, chen2026dflash}. On modern hybrid-attention backbones, these issues become more complex because diffusion training and inference must also handle recurrent-state handoff, mask realization, and serving-path design. How to convert a strong AR model into a capable and serving-efficient hybrid-backbone dLLM under a limited training budget therefore remains an open problem.

\begin{figure}[!t]
\centering
\vspace{-3em}
\includegraphics[width=1\linewidth]{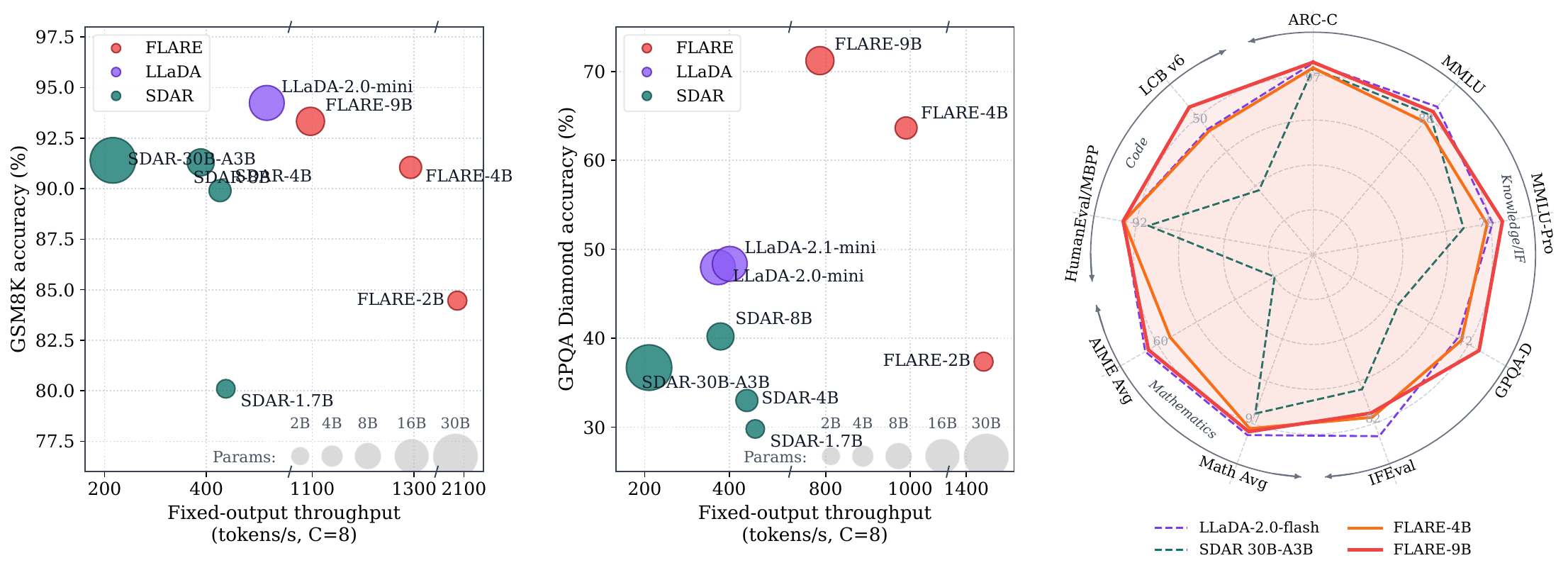}
\caption{\model is a state-of-the-art dLLM with high sampling throughput and near AR performance.}
\label{fig:teaser}
\vspace{-2em}
\end{figure}

Motivated by these challenges, we present \model, a systematic recipe for converting hybrid-attention AR LLMs into capable and serving-efficient dLLMs under a practical training budget. \model examines the full conversion pipeline, from transfer-data construction and loss--mask design to hybrid-backbone training and serving-time inference. Concretely, \model makes three contributions. (1) \textbf{Transfer diagnosis.} To understand the performance degradation observed in prior AR-to-dLLM methods, we conduct controlled ablations over transfer data, objective design, and attention-mask choices, and identify transfer-data quality and distribution match as the dominant factors for preserving AR capability, outweighing loss formulation and mask design. This addresses a factor that has often been under-discussed in prior transfer work \citep{cheng2025sdar, wu2025fast2, fu2025efficient, liu2025tidar}. (2) \textbf{Efficient training.} To enable diffusion training on modern softmax-plus-linear-attention backbones, we develop hardware-aware algorithms for computing linear attention under diffusion-specific visibility patterns \citep{yang2024gated, dao2024transformers, team2025kimi}, realizing these patterns through recurrent-state handoff rather than dense masking. (3) \textbf{Efficient generation.} To turn parallel-generation capability into wall-clock throughput, we build a unified inference system that supports both AR-style verified decoding and diffusion-style parallel denoising from the same checkpoint, extending efficient LLM serving ideas \citep{zheng2024sglang} to hybrid-backbone dLLMs.

Taken together, the study yields four main takeaways. 1) \textbf{Objective and mask design matter for capability preservation:} a token-balanced clean/noisy loss combines AR next-token supervision with block-diffusion denoising, while the document-packed clean/noisy mask enables both causal and bidirectional training signals without cross-document leakage. 2) \textbf{Transfer data explains the remaining variation:} once the core loss--mask design is aligned, transfer-data quality and distribution match dominate the residual performance gap, making AR-SFT a useful low-cost proxy for data selection. 3) \textbf{Hybrid-backbone diffusion requires specialized training:} diffusion visibility on linear-attention components cannot be imposed as a dense mask alone, and \model realizes it through recurrent-state scheduling and hardware-aware training kernels. 4) \textbf{One checkpoint supports two generation regimes:} starting from Qwen3.5 hybrid-attention checkpoints~\citep{qwen2026qwen35}, \model-2B/4B/9B achieves competitive diffusion-LM quality while supporting both AR-style verified speculative decoding and diffusion-style parallel denoising (Figure~\ref{fig:teaser}).

\section{Related Work}
\label{sec:related}

Within the AR factorization, multi-token prediction and speculative decoding relax the one-token-per-forward bottleneck by drafting several future tokens and verifying them with a target model~\citep{gloeckle2024better,li2025eagle,cai2024medusa,leviathan2023fast,kumar2026speculative}. These methods preserve the AR distribution under verification rules, but they still rely on a left-to-right reference model. Diffusion language models instead learn to fill masked positions through iterative denoising, allowing multiple token positions to be committed in one or a few forward passes. This line includes discrete diffusion models~\citep{lou2023discrete,sahoo2024simple,shi2024simplified,ou2024your,sahoo2025diffusion}, large-scale dLLMs such as LLaDA, Dream, Mercury, Seed Diffusion~\citep{nie2025large,bie2025llada2,bie2026llada2,ye2025dream, khanna2025mercury,googledeepmind2025geminidiffusion,song2025seed}, and inference-oriented systems such as DFlash~\citep{chen2026dflash}. Block-diffusion LMs~\citep{arriola2025block} provide a middle ground: blocks are generated sequentially, while positions inside a block can be denoised in parallel. A group of works reduces the training cost of this paradigm by converting or adapting pretrained AR checkpoints into dLLMs~\citep{gong2024scaling,fu2025efficient,wu2025fast2,cheng2025sdar,liu2025tidar,liu2025wedlm,yu2026introspective}. These works make AR-to-diffusion transfer practical, but the sources of transfer degradation remain entangled across objective design, attention masks, transfer data, and evaluation protocol.

A separate line of work reduces the cost of each forward pass. Linear-attention, state-space, and recurrent-memory models replace the growing softmax-attention KV cache with a compact state~\citep{katharopoulos2020transformers,choromanski2020rethinking,schlag2021linear,peng2023rwkv,sun2023retnet,gu2023mamba,dao2024transformers,peng2025rwkv,lahoti2026mamba}. Delta-rule and gated linear-attention variants further improve the recurrent update used by modern efficient LLMs~\citep{yang2024parallelizing,yang2024gated,team2025kimi}. Related test-time-state models such as TTT and Atlas also fit the broader recurrent-memory trend shown in Figure~\ref{fig:lm-timeline}~\citep{sun2024learning,zhang2025test,behrouz2025atlas}. This improves memory scaling, especially for long contexts, but pure linear stacks can still be weaker than softmax attention on some tasks. As a result, recent efficient LLMs increasingly use hybrid backbones that interleave softmax and linear-attention layers, including Samba, Hymba, Zamba, MiniMax, Qwen3-Next/Qwen3.5, Kimi Linear, and Nemotron 3~\citep{ren2024samba,dong2024hymba,glorioso2024zamba,li2025minimax,qwen2025qwen3next,qwen2026qwen35,team2025kimi,basant2025nvidia,blakeman2025nemotronh,blakeman2025nemotron,blakeman2025nvidia}. These architectures reduce cache size and long-context cost while largely retaining AR training and decoding.

Taken together, the timeline in Figure~\ref{fig:lm-timeline} shows that open-source LLM efficiency has advanced along two largely separate threads: generation-side parallelism, including dLLMs, and architecture-side reductions in per-forward cost through hybrid attention. Yet these threads have rarely been joined: dLLM work mostly assumes full-attention backbones, while hybrid LLMs largely retain AR training and decoding. \model studies this missing intersection as AR-to-diffusion post-training for hybrid checkpoints, adding a diffusion-style generation path to the same parameters while preserving causal generation. This requires jointly specifying the objective, clean/noisy mask, transfer data, recurrent-state implementation, and decoding interfaces developed next.

\section{Method}
\label{sec:method}

\subsection{Preliminaries on Attention Mechanisms}
\label{sec:method-prelim}
To ground the subsequent analysis of hybrid-attention backbones and non-causal visibility patterns, we introduce softmax and linear attention under a unified \emph{associative-memory} view~\citep{zhong2025understanding, wang2025test}, which abstracts attention layers as modules that memorize and retrieve. For notation, we denote the per-token queries and keys
$\mathbf{q}_t,\mathbf{k}_t\in\mathbb{R}^{d_k}$ and values
$\mathbf{v}_t\in\mathbb{R}^{d_v}$, received by each layer. It emits an output
$\mathbf{o}_t\in\mathbb{R}^{d_v}$. Under this view, softmax and linear attention share a memory--retrieval interface, but differ in how past $(\mathbf{k}_i,\mathbf{v}_i)$ pairs are stored and queried by $\mathbf{q}_t$.

\noindent \textbf{Softmax attention as explicit key-value memory.}
A softmax layer explicitly stores all visible key-value (KV) pairs $(\mathbf{k}_i,\mathbf{v}_i)$ in a KV cache. Given a query $\mathbf{q}_t$, retrieval is performed by normalizing query-key similarities over a visible index set and taking a weighted sum of the corresponding values. Using a generic visible set $\mathcal{V}_t$, the softmax retrieval is
\begin{equation}
  \mathbf{o}_t
  =
  \sum_{i\in\mathcal{V}_t}
  \frac{\exp(\mathbf{q}_t^{\!\top}\mathbf{k}_i/\sqrt{d_k})}
       {\sum_{j\in\mathcal{V}_t}\exp(\mathbf{q}_t^{\!\top}\mathbf{k}_j/\sqrt{d_k})}
  \mathbf{v}_i,
  \label{eq:softmax-retrieval}
\end{equation}
where $\mathcal{V}_t$ denotes the visible index set for position $t$; for full attention, $\mathcal{V}_t$ contains all positions, while for causal attention it contains positions no later than $t$. This explicit key-value memory provides accurate token-level access and long-range retrieval capability, but its KV-cache footprint grows as $\mathcal{O}(T(d_k+d_v))$; dense full-sequence attention costs $\mathcal{O}(T^2)$, while incremental decoding still requires $\mathcal{O}(T)$ retrieval work per generated token.

\noindent \textbf{Linear attention as recurrent memory.}
Linear attention~\citep{katharopoulos2020transformers, choromanski2020rethinking}
replaces the explicit KV cache with a bounded-size recurrent state
$\mathbf{S}_t\in\mathbb{R}^{d_k\times d_v}$. In its simplest form, the state
accumulates past key-value information through outer-product updates,
\begin{equation}
  \mathbf{S}_t
  =
  \mathbf{S}_{t-1}
  +
  \mathbf{k}_t\mathbf{v}_t^{\!\top},
  \qquad
  \mathbf{o}_t
  =
  \mathbf{S}_t^{\!\top}\mathbf{q}_t .
  \label{eq:linear-retrieval}
\end{equation}
Retrieval is therefore performed from this compressed state rather than through explicit attention over stored key-value pairs. Modern efficient LLMs often instantiate recurrent memory with gating, decay, or delta-rule updates, including Gated DeltaNet~\citep{yang2024gated} and Kimi Delta Attention~\citep{team2025kimi}. These mechanisms share a bounded-state interface but differ in their concrete parameterizations. In this work, we use Gated DeltaNet (GDN) as the representative instantiation for exposition. Its memorization and retrieval rule, the Gated Delta Rule (GDR), is
\begin{equation}
  \mathbf{S}_t
  =
  \alpha_t\,\bigl(\mathbf{I}-\beta_t\,\mathbf{k}_t\mathbf{k}_t^{\!\top}\bigr)\mathbf{S}_{t-1}
  +
  \beta_t\,\mathbf{k}_t\mathbf{v}_t^{\!\top},
  \qquad
  \mathbf{o}_t
  =
  \mathbf{S}_t^{\!\top}\mathbf{q}_t / \sqrt{d_k}.
  \label{eq:gdn-recurrence}
\end{equation}
The state-scheduling principle developed later relies on this recurrent-state interface rather than the specific parameterization of GDN, and can extend to other linear-attention variants. Compared with softmax attention, recurrent-memory attention maintains an $\mathcal{O}(d_kd_v)$ state independent of sequence length; processing a length-$T$ sequence costs $\mathcal{O}(T)$ time, and incremental decoding uses constant-size state updates and retrieval per generated token, at the cost of a bounded and lossy memory.

\noindent \textbf{Hybrid attention backbones.}
The two mechanisms above occupy different points on the spectrum of retrieval fidelity, memory compression, and serving efficiency. Softmax attention preserves explicit key-value memory and supports accurate token-level and long-range retrieval, as in Eq.~\eqref{eq:softmax-retrieval}, while recurrent-memory attention compresses history into bounded states, as in Eq.~\eqref{eq:gdn-recurrence}, reducing the dependence of memory on context length at the cost of lossy retrieval. Hybrid-attention backbones combine these mechanisms at the architecture level: softmax layers provide high-fidelity access to selected context, whereas recurrent-memory layers act as efficient compressed memory. By controlling the ratio and placement of these layer types, recent efficient LLM families seek to preserve model capability while reducing cache size and long-context serving cost~\citep{dong2024hymba, glorioso2024zamba, li2025minimax, qwen2025qwen3next, qwen2026qwen35, blakeman2025nvidia, team2025kimi}. Our main \model instantiation follows this setting. A hybrid backbone contains both explicit key-value memory and recurrent-state memory, so the same non-causal visibility pattern must be implemented through different mechanisms across layer types; the next subsection instantiates this issue for the clean/noisy training mask.

\subsection{Training Objective and Mask Design}
\label{sec:method-training}

Let $\mathbf{x}=(x^{1},\ldots,x^{L})\in\mathcal{V}^{L}$ be a training sample partitioned into $K$ contiguous blocks $\mathbf{x}_{b}=(x^{(b-1)B+1},\ldots,x^{bB})$ of size $B$ ($L=KB$); we write $\mathbf{x}^{<b}$ for all tokens strictly before block $b$. For each block, we sample a masked subset $\mathcal{M}_{b}\subseteq\{(b-1)B+1,\ldots,bB\}$ and form the noisy block $\tilde{\mathbf{x}}_{b}$ by replacing $x^{\ell}$ with \texttt{[MASK]} for $\ell\in\mathcal{M}_{b}$. The complement $\mathcal{M}_{b}^{c}$ defines a second, disjoint noisy view $\tilde{\mathbf{x}}_{b}^{c}$. Based on this block-level corruption process, \model maximizes a token-balanced clean/noisy objective that preserves the AR path while adding diffusion-style supervision. The clean stream follows the original causal order and provides the next-token prediction term $\mathcal{L}_{\mathrm{AR}}$. The noisy stream provides the diffusion term $\mathcal{L}_{\mathrm{diff}}$: the two complementary noisy views predict disjoint token subsets, both conditioned on the same preceding clean context. The overall objective is
\begin{equation}
  \mathcal{L}_{\model}(\theta)
  =
  \mathcal{L}_{\mathrm{AR}}(\theta)
  +
  \mathcal{L}_{\mathrm{diff}}(\theta).
  \label{eq:flare-loss}
\end{equation}
Concretely, the clean-stream term is the usual next-token log-likelihood over the full sequence, and the noisy-stream term sums the denoising log-likelihood over the two complementary masked views:
\begin{align}
\mathcal{L}_{\mathrm{AR}}(\theta)
&=
\sum_{\ell=1}^{L}
\log p_\theta\!\left(x^{\ell}\mid \mathbf{x}^{<\ell}\right), \nonumber \\
\mathcal{L}_{\mathrm{diff}}(\theta)
&=
\sum_{b=1}^{K}
\left[
\underbrace{
\sum_{\ell\in\mathcal{M}_{b}}
\log p_\theta\!\left(x^{\ell}\mid \tilde{\mathbf{x}}_{b},\mathbf{x}^{<b}\right)
}_{\text{primary noisy view}}
+
\underbrace{
\sum_{\ell\in\mathcal{M}_{b}^{c}}
\log p_\theta\!\left(x^{\ell}\mid \tilde{\mathbf{x}}_{b}^{c},\mathbf{x}^{<b}\right)
}_{\text{complementary noisy view}}
\right].
\label{eq:flare-loss-terms}
\end{align}

\begin{wrapfigure}[24]{r}{0.46\textwidth}
\vspace{-4mm}
\centering
\includegraphics[width=\linewidth]{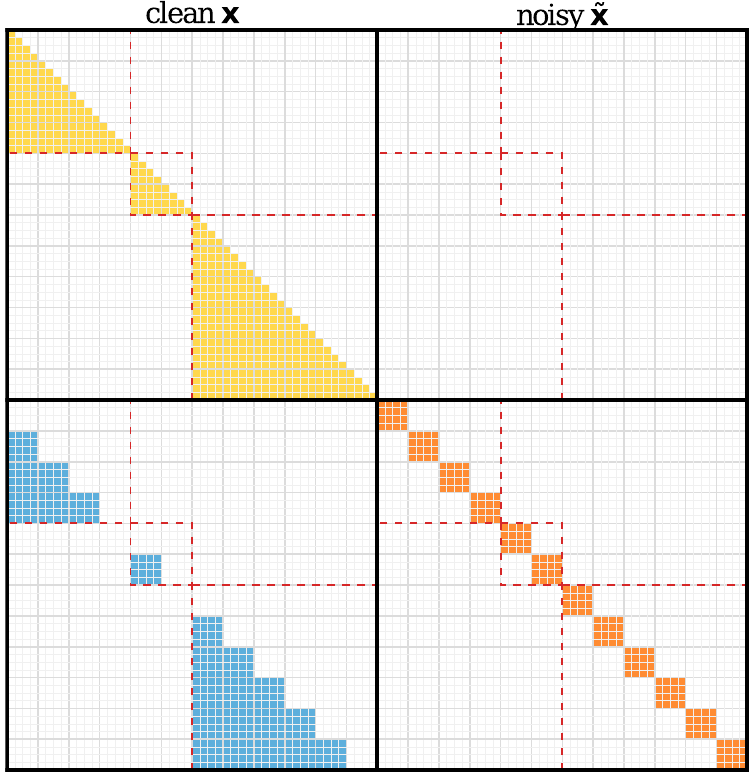}
\caption{\model document-packed clean/noisy mask: causal clean attention (yellow), bidirectional noisy blocks (orange), noisy-to-clean visibility (blue), and document boundaries (red dashed).}
\label{fig:flare-mask-train}
\vspace{-1mm}
\end{wrapfigure}
The clean/noisy construction in Figure~\ref{fig:flare-mask-train} realizes the two terms in Eq.~\eqref{eq:flare-loss-terms}: the causal clean stream supplies $\mathcal{L}_{\mathrm{AR}}$, while the block-bidirectional noisy stream supplies $\mathcal{L}_{\mathrm{diff}}$ under two complementary views. This differs from standard mask-dLLM training~\citep{arriola2025block, nie2025large, cheng2025sdar}, where masked positions are predicted only from noisy contexts. In \model, every token contributes one AR signal and one diffusion signal at unit weight because $\mathcal{M}_{b}$ and $\mathcal{M}_{b}^{c}$ partition each block. This keeps the two supervision sources balanced under varying mask fractions while giving the same checkpoint both causal generation and block-level parallel denoising capabilities.
Under the memory view above, an attention mask specifies which token information is accessible when a query reads from memory. Figure~\ref{fig:flare-mask-train} shows the document-packed clean/noisy visibility pattern used to compute the objective above in a packed forward pass: the clean stream remains token-causal and isolated from the noisy stream, each noisy block is bidirectional within the block and attends to preceding clean context, and document boundaries isolate packed samples to prevent cross-document leakage while avoiding padding overhead. The two memory implementations in a hybrid backbone realize this same pattern differently: in softmax layers, an additive mask $\mathbf{M}\in\{0,-\infty\}^{T\times T}$ inside the softmax of Eq.~\eqref{eq:softmax-retrieval} directly removes forbidden entries from the visible set $\mathcal{V}_t$; in recurrent-memory layers, making a past pair $(\mathbf{k}_i,\mathbf{v}_i)$ inaccessible to a later query means preventing its information from entering or propagating through the relevant state $\mathbf{S}_t$. Thus the same non-causal clean/noisy mask becomes a schedule of writes, state propagation, and state resets rather than a post-hoc masking matrix; the corresponding recurrent-state schedule is derived below.

Algorithmically, this construction is closest to TiDAR~\citep{liu2025tidar} and concurrent work I-DLM~\citep{yu2026introspective}, but differs in two choices that matter for decoding. Unlike TiDAR, we apply logit shift to the noisy-stream diffusion terms to align them with AR semantics. Unlike I-DLM, we keep the noisy stream block-bidirectional and use random rather than fully masked noisy views during training, which preserves Diffusion-Trust decoding in addition to AR-style decoding.

The clean/noisy mask above can be applied as a visibility matrix in softmax layers, but must be translated into a state schedule for linear-attention layers. We use the generic form $\mathbf{S}_\ell=\mathcal{T}_\ell\mathbf{S}_{\ell-1}+\mathcal{W}_\ell$ with per-position output $\mathbf{o}_\ell=f(\mathbf{S}_\ell,\mathbf{q}_\ell)$, which covers Mamba-2~\citep{dao2024transformers}, Gated DeltaNet~\citep{yang2024gated}, and Kimi Delta Attention~\citep{team2025kimi}. We instantiate the construction with GDN below, while the same state-scheduling idea extends to other members of this family by substituting the corresponding transition, write, and output maps. Let $e(b)=bB$; we index clean positions by $t$, noisy positions by $\ell$, and write noisy-side quantities with a tilde. For clarity, the equations below are written for a single document segment; under document packing, the same schedule is applied independently within each segment, with states reset at document boundaries.

For a GDN layer, the clean stream follows the standard causal recurrence of Eq.~\eqref{eq:gdn-recurrence}. For noisy block $b$, the recurrent state is initialized from the preceding clean boundary state, $\tilde{\mathbf{S}}_{e(b-1)}:=\mathbf{S}_{e(b-1)}$, and then updated only with noisy tokens from the same block:
\begin{equation}
\left\{
\begin{aligned}
\mathbf{S}_t
&=
\alpha_t\bigl(\mathbf{I}-\beta_t\mathbf{k}_t\mathbf{k}_t^{\!\top}\bigr)\mathbf{S}_{t-1}
+
\beta_t\mathbf{k}_t\mathbf{v}_t^{\!\top},
&& \text{clean stream}, \\
\tilde{\mathbf{S}}_\ell
&=
\tilde\alpha_\ell\bigl(\mathbf{I}-\tilde\beta_\ell\tilde{\mathbf{k}}_\ell\tilde{\mathbf{k}}_\ell^{\!\top}\bigr)
\tilde{\mathbf{S}}_{\ell-1}
+
\tilde\beta_\ell\tilde{\mathbf{k}}_\ell\tilde{\mathbf{v}}_\ell^{\!\top},
&& \text{noisy block } b,
\end{aligned}
\right.
\label{eq:flare-gdn-schedule}
\end{equation}
where $\ell\in\{e(b-1)+1,\ldots,e(b)\}$. The clean branch uses the standard causal readout
$\mathbf{o}_t=\mathbf{S}_t^{\!\top}\mathbf{q}_t/\sqrt{d_k}$, while the noisy branch reads every position from the shared block-end state
$\tilde{\mathbf{o}}_\ell=\tilde{\mathbf{S}}_{e(b)}^{\!\top}\tilde{\mathbf{q}}_\ell/\sqrt{d_k}$.
This block-end readout realizes bidirectional visibility within the noisy block, and resetting the noisy state to the corresponding clean boundary state prevents information flow across noisy blocks. These recurrences define the intended visibility pattern, but the direct sequential form is not used for training; the chunk-parallel realization and kernel implementation are discussed in Section~\ref{sec:flare-kernel} and Appendix~\ref{app:kernel-routes}.

\subsection{Decoding Interfaces}
\label{sec:method-inference}

The clean/noisy objective in Section~\ref{sec:method-training} gives a trained \model checkpoint two usable output distributions under the same forward interface: clean-stream logits preserve causal AR prediction, while noisy-stream logits denoise a bidirectional block conditioned on clean context. This separation is important for transfer: the seed checkpoint's capability is organized around next-token prediction, and a pure diffusion path would need to replace that causal distribution under limited post-training data. The clean stream therefore anchors the transferred AR capability and provides a verification reference, while the noisy stream adds block-level parallel denoising. This yields two decoding interfaces. In \emph{Diffusion-Trust}, noisy-stream samples are the reference and are committed through block-denoising confidence. In \emph{AR-Trust}, noisy-stream samples serve as drafts and clean-stream logits verify them left-to-right. Algorithms~\ref{alg:diffusion-trust-interface} and~\ref{alg:ar-trust-interface} give the operational view; below, we formalize the corresponding commitment and verification rules.

We use $x$ for the input prompt, $y$ for the generated sequence, $f_\theta$ for the trained checkpoint's forward map, $z$ for pre-softmax logits, and $\sigma_{k,p,T}$ for temperature-scaled top-$k$/top-$p$ sampling. Diffusion-Trust uses noisy-stream distributions $\pi_i$; AR-Trust uses clean-stream target distributions $p_i$, noisy proposal laws $q_i$, and draft tokens $d$. Decoding schematics and path-specific masks are provided in Appendix~\ref{app:inference}, while system-level serving details are discussed in Section~\ref{sec:flare-decoding}.

\begin{center}
\begin{minipage}[t]{0.485\linewidth}
\vspace{0pt}
\begin{tcolorbox}[enhanced,colback=white,colframe=noteline,boxrule=0.35pt,arc=6pt,outer arc=6pt,left=4pt,right=4pt,top=4pt,bottom=4pt,height=7.05cm,valign=top]
\refstepcounter{algorithm}\label{alg:diffusion-trust-interface}
\noindent\textbf{Algorithm~\thealgorithm.} Diffusion-Trust Decoding.\par
\vspace{0.25em}
\small\normalfont\rmfamily
\begin{algorithmic}[1]
\Statex \textbf{Input:} $x$, prefix $y_{<t}$, block size $B$, steps $S$
\Statex \textit{\textcolor{axiscol}{Initialize Active Block}}
\State $y_{t:t+B-1}\leftarrow\texttt{[MASK]}^{B}$,\quad $R_1\leftarrow\{t,\ldots,t+B-1\}$
\Statex \textit{\textcolor{axiscol}{Parallel Block Denoising}}
\For{$s=1,\ldots,S$ until $R_s=\varnothing$}
  \State $\tilde{z}\leftarrow f_{\theta}(x,y_{<t},y_{t:t+B-1})$
  \State $\pi_i\leftarrow\sigma_{k,p,T}(\tilde{z}^{\mathrm{shift}}_i)$;\quad $\hat y_i\sim\pi_i$
  \Statex \hspace{\algorithmicindent}$c_i\leftarrow\operatorname{score}(\pi_i,\hat y_i)$
  \State $A_s\leftarrow\{i\in R_s:c_i\ge\gamma_s\}$; if $s=S$, set $A_s\leftarrow R_s$
  \State Update $y$ and $R_{s+1}$ by Eq.~\eqref{eq:diffusion-trust-unmask}
\EndFor
\Statex \textit{\textcolor{axiscol}{Commit State}}
\State Run token-causal replay on finalized block
\State Append block to prefix and advance the live state
\end{algorithmic}
\end{tcolorbox}
\end{minipage}
\hfill
\begin{minipage}[t]{0.485\linewidth}
\vspace{0pt}
\begin{tcolorbox}[enhanced,colback=white,colframe=noteline,boxrule=0.35pt,arc=6pt,outer arc=6pt,left=4pt,right=4pt,top=4pt,bottom=4pt,height=7.05cm,valign=top]
\refstepcounter{algorithm}\label{alg:ar-trust-interface}
\noindent\textbf{Algorithm~\thealgorithm.} AR-Trust Decoding.\par
\vspace{0.25em}
\small\normalfont\rmfamily
\begin{algorithmic}[1]
\Statex \textbf{Input:} $x$, prefix $y_{<t}$, held proposals $(d_i,q_i)_{i=1}^{K}$
\Statex \textit{\textcolor{axiscol}{Active Window Forward}}
\State $w\leftarrow[y_{<t};d_{1:K};\texttt{[MASK]}^{N-1}]$
\State $(z^{\mathrm{c}},\tilde{z})\leftarrow f_{\theta}(x,w)$,\quad $p_i\leftarrow\sigma_{k,p,T}(z^{\mathrm{c}}_i)$
\Statex \textit{\textcolor{axiscol}{Clean-Stream Verification}}
\For{$i=1,\ldots,K$}
  \State $a_i\sim\operatorname{Bern}\!\left(\min\{1,p_i(d_i)/q_i(d_i)\}\right)$
  \If{$a_i=0$}
    \State Emit correction from Eq.~\eqref{eq:accept-exact}; \textbf{break}
  \EndIf
  \State Emit $d_i$
\EndFor
\State If all drafts are accepted, emit $y^\star\sim p_{K+1}$
\Statex \textit{\textcolor{axiscol}{Proposal Refill}}
\State $(d'_{1:K},q'_{1:K})\leftarrow\operatorname{Draft}(\tilde{z})$
\end{algorithmic}
\end{tcolorbox}
\end{minipage}
\end{center}

We next state the two token-emission rules called by these interfaces.

\noindent\textbf{Diffusion-Trust: block iterative denoising and noisy-stream commitment.} This rule formalizes the commit operation invoked by the parallel denoising loop in Algorithm~\ref{alg:diffusion-trust-interface}. Consider an active block $\mathcal{B}_{t,B}=\{t,\ldots,t+B-1\}$ behind the causal prefix, decoded over steps $s=1,\ldots,S$, with $R_1=\mathcal{B}_{t,B}$. Let $R_s$ be the unresolved positions before step $s$. For each $i\in R_s$, the noisy stream defines $\pi_i=\sigma_{k,p,T}(\tilde{z}^{\mathrm{shift}}_i)$, samples $\hat y_i\sim\pi_i$, and computes a confidence score $c_i=\operatorname{score}(\pi_i,\hat y_i)$. The commit set $A_s$ is selected in parallel across the unresolved positions, and since the noisy stream is trusted in this path, selected candidates are written to $y$ without clean-stream verification:
\begin{subequations}
\label{eq:diffusion-trust-unmask}
\begin{align}
A_s
&=
\begin{cases}
\{i\in R_s:c_i\ge\gamma_s\}, & s<S,\\
R_s, & s=S,
\end{cases}
\label{eq:diffusion-trust-select}\\
y_i^{(s+1)}
&=
\begin{cases}
\hat y_i, & i\in A_s,\\
y_i^{(s)}, & i\in R_s\setminus A_s,
\end{cases}
\qquad
R_{s+1}=R_s\setminus A_s .
\label{eq:diffusion-trust-update}
\end{align}
\end{subequations}
Positions outside $A_s$ stay masked for the next denoising step, and the case $s=S$ commits all unresolved positions so the block is finalized after at most $S$ steps. Eq.~\eqref{eq:diffusion-trust-unmask} covers confidence-based~\citep{nie2025large} and margin-based~\citep{kim2025train} unmasking by changing how $c_i$ is computed. Because intermediate denoising forwards contain provisional block contents, their recurrent states are not written into the live causal state. Once the block is finalized, the causal replay step in Algorithm~\ref{alg:diffusion-trust-interface} commits the finalized tokens to the hybrid backbone state; the full serving loop is given in Appendix~\ref{app:inference-diffusion}.

\begin{samepage}
\noindent\textbf{AR-Trust: speculative verification.} This rule formalizes the clean-stream verification step in Algorithm~\ref{alg:ar-trust-interface}. At verify position $i$, let $z_i$ be the clean-stream logit and define the target distribution $p_i=\sigma_{k,p,T}(z_i)$. Let $q_i$ be the proposal law that first samples the held draft token $d_i$. The reference verifier is
\begin{subequations}
\label{eq:accept-exact}
\begin{align}
d_i&\sim q_i,\qquad
a_i\sim\operatorname{Bern}\!\left(\min\!\left\{1,\frac{p_i(d_i)}{q_i(d_i)}\right\}\right),
\label{eq:accept-exact-draw}\\
y_i&=
\begin{cases}
d_i, & a_i=1,\\
y^\star,\quad
y^\star\sim\operatorname{Norm}_{\mathcal{V}}\!\left([p_i-q_i]_+\right), & a_i=0.
\end{cases}
\label{eq:accept-exact-emit}
\end{align}
\end{subequations}
Here $\operatorname{Bern}(r)$ denotes a Bernoulli draw with success probability $r$, $a_i\in\{0,1\}$ is the accept indicator, $[t]_+=\max\{t,0\}$, and $\operatorname{Norm}_{\mathcal{V}}$ normalizes a nonnegative vector over the vocabulary. The exactness condition is that $q_i$ must be the same proposal law that sampled $d_i$; recomputing it during verification can break distribution equivalence because noisy proposals are generated in parallel under the clean/noisy block mask. Lower-overhead policies may approximate $q_i$ or use argmax drafts, trading strict equivalence for serving efficiency; details are in Appendix~\ref{app:inference-ar}.
\end{samepage}

\begin{wrapfigure}{r}{0.48\textwidth}
\vspace{-0.9em}
\centering
\includegraphics[width=\linewidth]{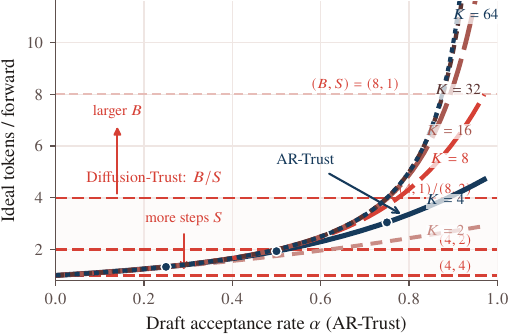}
\vspace{-0.65em}
\caption{\textbf{Analytical decoding-speed regimes.} AR-Trust varies with acceptance rate $\alpha$ and horizon $K$; Diffusion-Trust scales with the block-to-step ratio $B/S$.}
\label{fig:accept-rate-speed}
\vspace{-1.0em}
\end{wrapfigure}

Eqs.~\eqref{eq:diffusion-trust-unmask} and~\eqref{eq:accept-exact} define the two regimes used by Algorithms~\ref{alg:diffusion-trust-interface} and~\ref{alg:ar-trust-interface}. In AR-Trust, Eq.~\eqref{eq:accept-exact} accepts each held draft with probability $\min\{1,p_i(d_i)/q_i(d_i)\}$; following the acceptance-rate analysis of I-DLM~\citep{yu2026introspective}, the plotted $\alpha$ abstracts the average success probability across horizon $K$, so speed improves only when noisy proposals align with the clean target. In Diffusion-Trust, Eq.~\eqref{eq:diffusion-trust-unmask} commits an active block over at most $S$ denoising steps, so finalizing $B$ tokens in $S$ forwards yields the ideal ratio $B/S$. Figure~\ref{fig:accept-rate-speed} summarizes these analytical regimes and isolates the algorithmic control variables: $\alpha$ and $K$ for AR-Trust, and $B$ and $S$ for Diffusion-Trust. System factors such as state replay, cache updates, and kernel scheduling are deliberately excluded here; measured throughput is reported in Section~\ref{sec:flare-decoding}.

\FloatBarrier

\section{Experimental Analysis}
\label{sec:study}

We use this section to identify the key algorithmic and data ingredients of AR-to-dLLM transfer. While existing conversion methods~\citep{cheng2025sdar, wu2025fast2, fu2025efficient} report a range of transfer outcomes, several design choices --- the training objective, the clean/noisy visibility pattern, whether the clean stream is explicitly aligned with AR next-token prediction, logit shift, transfer-data composition, and evaluation protocol --- have not yet been examined systematically and in isolation. We therefore start from a smaller model, Qwen3-1.7B, and conduct a controlled study under a fixed training budget and evaluation suite, varying algorithmic ingredients and transfer data separately, so as to surface the recipe components that drive successful transfer. The resulting recipe is then carried over to stronger hybrid-attention models in Section~\ref{sec:flare}.

All transfer runs initialize from the same pretrained Qwen3-1.7B checkpoint~\citep{yang2025qwen3}. In our SFT format, the prompt tokens act as the leading clean-stream context (visible through attention but excluded from the loss), while the assistant response tokens are partitioned into blocks of size $B$ that simultaneously receive the $\mathcal{L}_{\mathrm{AR}}$ and $\mathcal{L}_{\mathrm{diff}}$ terms defined in Section~\ref{sec:method-training} --- i.e., the clean/noisy dual-stream supervision is applied only on response tokens. All runs use a maximum sequence length of 4096 with the document-level packing mask of Figure~\ref{fig:flare-mask-train}, a global batch size of 256, and 9000 optimizer steps, corresponding to roughly 10B training tokens. We evaluate all checkpoints on the same 12-task suite, covering Math + Reasoning (GSM8K~\citep{cobbe2021training}, MATH-500~\citep{hendrycks2021measuring}, AIME 24/25~\citep{aime2024}, ARC-Challenge~\citep{clark2018think}, GPQA-Diamond~\citep{rein2023gpqa}), Knowledge + IF (MMLU~\citep{hendrycks2020measuring}, MMLU-Pro~\citep{wang2024mmlu}, IFEval~\citep{zhou2023instruction}), and Code (HumanEval~\citep{chen2021evaluating}, MBPP~\citep{austin2021program}, LiveCodeBench~v6~\citep{jain2024livecodebench}). All checkpoints share one decoding protocol; detailed settings are in Appendix~\ref{app:per-benchmark}.

\begin{figure}[t]
\centering
\includegraphics[width=\linewidth]{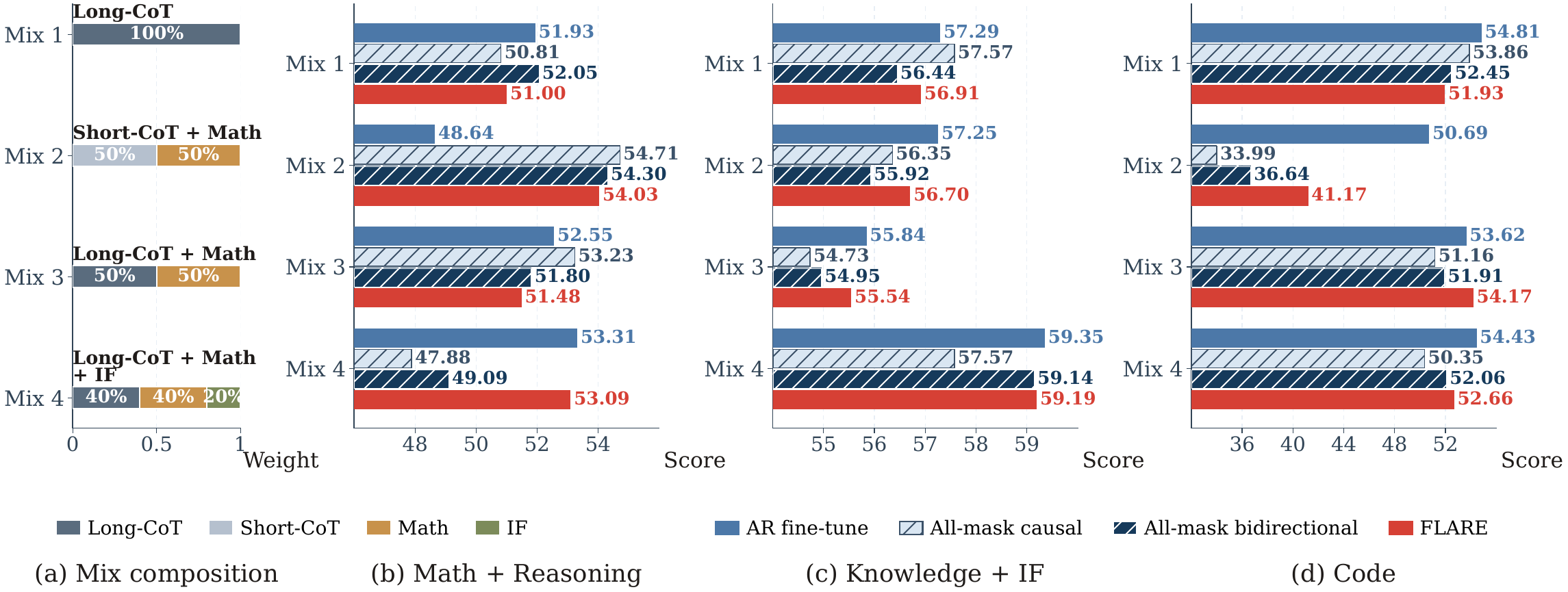}
\vspace{-0.2em}
\caption{Transfer performance across data compositions and capability groups. Panel (a) shows the per-mix sampling weights (mix composition); panels (b)--(d) plot scores for all four transfer recipes under the same Qwen3-1.7B seed and training budget on Math + Reasoning, Knowledge + Instruction Following, and Code, respectively, with a local x-axis range for readability. The trends show that converted dLLMs largely track their AR fine-tuning counterparts under the same data mix, while data composition changes the attainable transfer quality across the three capability groups.}
\label{fig:study-data}
\vspace{-0.5em}
\end{figure}

\begin{figure}[t]
\centering
\vspace{-0.8em}
\includegraphics[width=\linewidth]{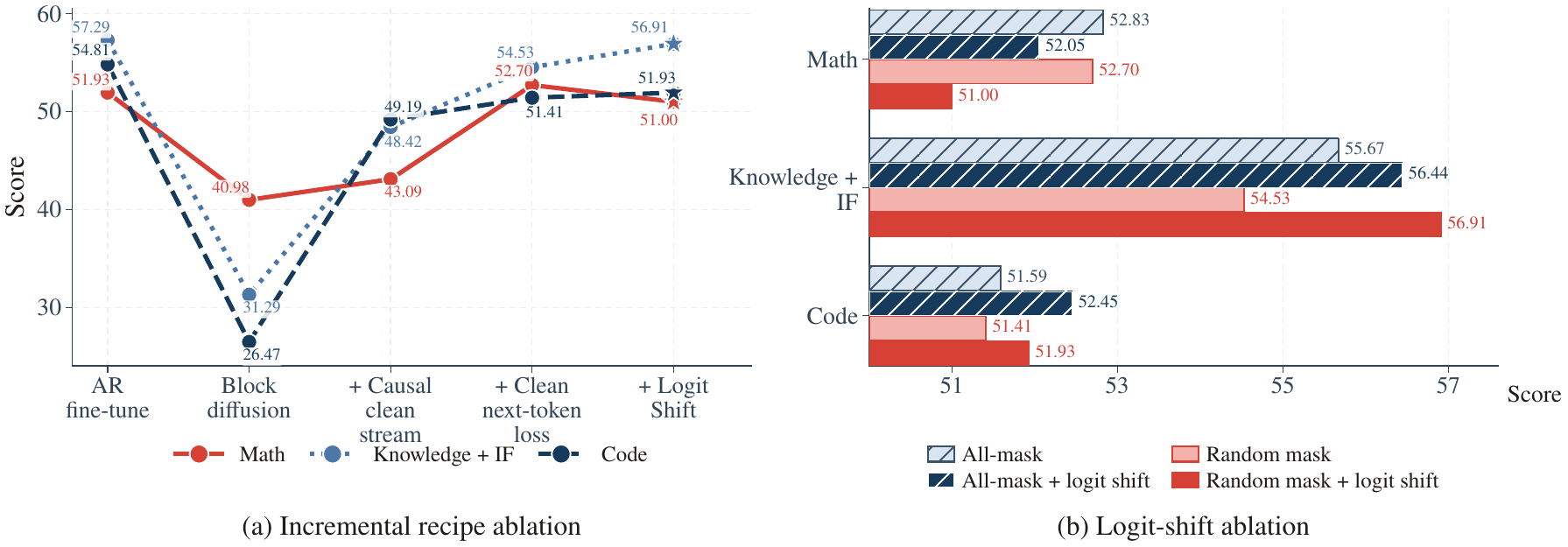}
\vspace{-0.5em}
\caption{Controlled ablations of AR-to-dLLM transfer ingredients on the fixed Mix~1 (\texttt{Long-CoT}) data condition. \textbf{(a)} The left panel follows a cumulative recipe: AR fine-tuning, pure block-diffusion transfer, adding a causal clean stream, adding clean-stream next-token loss, and adding logit-shifted noisy supervision. \textbf{(b)} Logit shift has a limited effect on category-level scores under both all-masked and random noisy masking, while the random noisy mask preserves Diffusion-Trust decoding.}
\label{fig:study-ingredients}
\vspace{0.1em}
\end{figure}

\subsection{Effects of Transfer Data Composition}
\label{sec:study-data}

We first study transfer data because it determines the regime in which the algorithmic ablations should be interpreted. We construct four controlled transfer-data mixes from curated open-source post-training SFT corpora released with the Llama-Nemotron~\citep{bercovich2025llama}, Nemotron Nano~2~\citep{basant2025nvidia}, Nemotron 3 ~\citep{blakeman2025nemotron}, and Nemotron-Cascade-2 ~\citep{yang2026nemotron} model families; the detailed curation, filtering, packing, and mixture-sampling procedure is provided in Appendix~\ref{app:data}. The mixes vary three factors while keeping the seed checkpoint, training budget, and evaluation protocol fixed: reasoning-trace length, math/domain coverage, and instruction-following coverage.

Concretely, the four mixes are built from three reusable source pools, each drawn from publicly released SFT corpora: a long chain-of-thought pool (\textbf{Long-CoT}, primarily Llama-Nemotron-Post-Training~\citep{bercovich2025llama}), a math pool (\textbf{Math}, the math splits of Llama-Nemotron-Post-Training plus Nemotron-Math-Proofs-v1), and an instruction-following pool (\textbf{IF}, the IF-tagged subsets of Nemotron-Cascade-1/-2-SFT-Data and Nemotron-Instruction-Following-Chat-v1/v2~\citep{basant2025nvidia}); a short-reasoning pool (\textbf{Short-CoT}, the English split of Nemotron-Post-Training-Dataset-v2~\citep{basant2025nvidia}) is used only by Mix~2. Specifically, (1) \textbf{Mix~1} (\texttt{Long-CoT}) uses the Long-CoT pool alone and serves as a reasoning-heavy baseline; (2) \textbf{Mix~2} (\texttt{Short-CoT+Math}) combines the Short-CoT pool with the Math pool to separate the effect of trace length from math coverage; (3) \textbf{Mix~3} (\texttt{Long-CoT+Math}) replaces Short-CoT with the Long-CoT pool to test whether stronger math coverage helps without shortening traces; and (4) \textbf{Mix~4} (\texttt{Long-CoT+Math+IF}) further adds the IF pool on top of Long-CoT and Math to test whether broader instruction coverage improves transfer beyond reasoning-centric domains. The four pools span roughly $0.5$--$17$B assistant tokens, and components are combined under per-mix sampling weights that we arrived at empirically through the comparisons reported below; the resulting weights are visualized in Figure~\ref{fig:study-data}(a) and detailed in Table~\ref{tab:app-data-weights} and Appendix~\ref{app:data-mixes}, while Appendix~\ref{app:data-selection} additionally documents an automatic, instance-level data-selection pipeline that we explored as a complementary route.

Figure~\ref{fig:study-data}(b)--(d) compares these four mixes across four transfer recipes under the same Qwen3-1.7B seed and training budget. AR fine-tuning (AR-SFT) denotes continued next-token fine-tuning on response tokens; all-mask causal (AM-Causal) and all-mask bidirectional (AM-Bidir) share the joint AR/diffusion supervision used by \model (formally introduced in Section~\ref{sec:study-objective}) but replace random noisy masking with all-masked noisy views, differing only in whether the noisy stream is token-causal or bidirectional within each block (AM-Causal is close in spirit to concurrent I-DLM~\citep{yu2026introspective}). We read the resulting comparison along two axes. First, on the recipe axis, the all-mask variants tend to trail both AR fine-tuning and \model---most visibly on Code, where they drop sharply on the math-heavy Mix~2 and Mix~4---while AR fine-tuning and \model track each other closely within each mix. The first observation flags an objective-side gap that we dissect under a fixed data mix in Section~\ref{sec:study-objective}; the second indicates that, with a properly aligned clean/noisy objective, AR fine-tuning is a faithful low-cost proxy for screening data mixes before running more expensive dLLM conversion. Second, on the data axis, no single data change is uniformly beneficial: Mix~2 slightly improves Math + Reasoning but substantially weakens Code, whereas Mix~4 gives the strongest Knowledge + Instruction Following results while keeping Math + Reasoning and Code competitive.

Two conclusions follow from the recipe and data axes of Figure~\ref{fig:study-data}. First, conditional on a properly aligned clean/noisy objective, the attainable transfer quality is governed primarily by the transfer data mix rather than by the algorithmic recipe; in particular, converted dLLMs track their AR fine-tuning counterparts under the same mix up to capability-group-specific deviations, so AR fine-tuning provides a low-cost proxy for screening data mixes prior to dLLM conversion. Second, comparisons of objective and attention-mask choices across different mixes are confounded by data quality and are therefore uninformative as algorithmic ablations. These observations motivate the controlled design used in the remainder of Section~\ref{sec:study}: Section~\ref{sec:study-objective} fixes Mix~1 (\texttt{Long-CoT}) and isolates the effects of objective design, clean/noisy visibility, logit shift, and noisy-mask sampling, while Section~\ref{sec:flare} adopts Mix~4 (\texttt{Long-CoT+Math+IF}) for the final hybrid-checkpoint conversions on the basis of its balanced behavior across capability groups in this sweep.

\subsection{Objective and Attention-Mask Ablations}
\label{sec:study-objective}

After isolating the data effect in Section~\ref{sec:study-data}, we fix the transfer data condition to Mix~1 (\texttt{Long-CoT}) and ask which objective and attention-mask ingredients prevent AR-to-dLLM conversion from losing the seed model's capability. All runs use the same document-level packing mask, which keeps variable-length prompt-response samples isolated inside packed batches while avoiding padding overhead. We therefore treat document-level masking as a fixed efficiency condition rather than an ablated factor, and focus on two conversion-specific choices: whether the clean stream retains AR-style token-causal supervision, and whether the noisy stream learns block-diffusion denoising.

Figure~\ref{fig:study-ingredients}(a) studies the clean-stream side through a cumulative recipe. Starting from AR fine-tuning, we replace next-token fine-tuning with a pure block-diffusion objective following the spirit of BD3-LM~\citep{arriola2025block} and SDAR~\citep{cheng2025sdar}, then add back a token-causal AR clean stream, an auxiliary next-token-prediction loss on clean-stream logits, and the logit-shifted noisy-stream supervision used by \model. The results show that pure block-diffusion transfer alone is not capability-preserving: it substantially degrades all three capability groups (averaging $-21.8$ points relative to AR fine-tuning, with the heaviest drops on Code and Knowledge + IF). The largest single-step recovery comes from restoring a token-causal AR clean stream ($+14.0$ points on average), which alone closes most of the Code and Knowledge + IF gap. Adding the clean-stream next-token-prediction loss further pulls Math + Reasoning back to the AR fine-tuning level by anchoring the converted model to the pretrained checkpoint's AR semantics, and logit shift on top yields a near-saturated final score.

Figure~\ref{fig:study-ingredients}(b) isolates the noisy-stream side. At fixed AR-aligned clean-stream supervision, all-masked noisy views and randomly masked diffusion views yield comparable category-level scores, so noisy-mask sampling is not a primary driver of benchmark recovery; its role is instead to determine decoding compatibility, since random masking trains the noisy stream as a genuine denoising path and preserves Diffusion-Trust decoding, while logit shift aligns the block-diffusion logits with inference-time token positions and avoids a wasted block-boundary prediction.

Two conclusions follow from the clean- and noisy-stream axes of Figure~\ref{fig:study-ingredients}. First, capability-preserving AR-to-dLLM conversion is governed primarily by clean-stream alignment, with logit shift and noisy-mask sampling contributing only marginal additional benchmark accuracy. Second, these auxiliary ingredients are retained for decoding compatibility, since they preserve both AR-Trust and Diffusion-Trust sampling paths. Combined with Section~\ref{sec:study-data}, this isolates the residual transfer gap to the quality of the data mix.

\section{Hybrid-Backbone Conversion and Evaluation}
\label{sec:flare}
Section~\ref{sec:study} decomposed AR-to-dLLM transfer into two complementary parts: an AR-aligned clean stream that determines whether capability is preserved, and a noisy-stream design that determines whether the converted checkpoint remains compatible with diffusion-style parallel decoding. This section applies the resulting configuration to strong hybrid-attention backbones and asks three concrete questions: (i) whether the transfer configuration selected by the controlled study scales to Qwen3.5 checkpoints~\citep{qwen2026qwen35} so that the converted models retain the capability of their AR source; (ii) whether two-stream supervision can be executed on a hybrid-attention backbone at hardware efficiency close to pure-AR training; and (iii) whether a single checkpoint can support both AR-Trust and Diffusion-Trust decoding under one unified serving stack. The three questions are addressed in turn in the subsections that follow.

We train \model-2B/4B/9B from the post-trained Qwen3.5-2B/4B/9B AR checkpoints, and compare the converted models against a set of leading dLLM systems: the commercial Mercury-2~\citep{inception2026mercury2}, the LLaDA-2.0/2.1 mini and flash families~\citep{bie2025llada2, bie2026llada2}, and the SDAR family~\citep{cheng2025sdar}. All three \model checkpoints are trained in a single supervised-finetuning stage on Mix~4 (\textbf{\texttt{Long-CoT+Math+IF}}), the data condition selected in Section~\ref{sec:study-data}. We keep the main protocol aligned with Section~\ref{sec:study}: sequence length $4096$, document-level packing masks, global batch size $256$, and $9000$ optimizer steps, corresponding to roughly 10B training tokens. The objective is the token-balanced AR/diffusion loss $\Lflare$ in Eq.~\eqref{eq:flare-loss}, with block size $B=4$. This conversion budget is markedly more modest than prior AR-to-dLLM transfer work, which typically uses 50B--200B tokens~\citep{cheng2025sdar, fu2025efficient}. Evaluation follows the 12-benchmark protocol from Section~\ref{sec:study}; we report \model under both AR-Trust and Diffusion-Trust sampling, with the original Qwen3.5 checkpoints serving as AR source-model references. The LLaDA and SDAR rows are taken from their technical reports, Mercury-2 is evaluated through its official API, and the Qwen3.5 and \model rows are measured under our SGLang-based inference stack~\citep{zheng2024sglang}.

\FloatBarrier

\subsection{Capability Retention Across Model Scales}

We first evaluate whether the converted checkpoints preserve the capability of their AR initialization. Tables~\ref{tab:flare-main-9b} and~\ref{tab:flare-small} report results at 9B and at smaller scales, with the original Qwen3.5 checkpoints serving as source-model references and the dLLM systems listed above as external baselines.

\begin{table}[!htbp]
\centering
\scriptsize
\setlength{\abovecaptionskip}{2pt}
\setlength{\belowcaptionskip}{3pt}
\setlength{\tabcolsep}{2.2pt}
\renewcommand{\arraystretch}{0.92}
\caption{\textbf{Benchmark performance of \model-9B}. \textbf{Bold}: best open-source dLLM; \underline{underline}: second best; \textsuperscript{*}: potentially under-reported.}
\label{tab:flare-main-9b}
\resizebox{\linewidth}{!}{%
\begin{tabular}{lcccccc|ccc}
\toprule
 & \textbf{Mercury-2} & \textbf{LLaDA-2.0} & \textbf{LLaDA-2.1} & \textbf{LLaDA-2.0} & \textbf{LLaDA-2.1} & \textbf{SDAR} & \textbf{\model} & \textbf{\model} & \textbf{Qwen3.5} \\
 &                     & \textbf{-mini}     & \textbf{-mini}     & \textbf{-flash}    & \textbf{-flash}    & \textbf{30B-A3B} & \textbf{-9B} & \textbf{-9B} & \textbf{-9B} \\
\textit{Params}  & \textit{Commercial} & \textit{16B-A1B} & \textit{16B-A1B} & \textit{100B-A5B} & \textit{100B-A5B} & \textit{30B-A3B} & \textit{9B} & \textit{9B} & \textit{9B} \\
\textit{Sampling mode} & \textit{Diffusion} & \textit{Diffusion} & \textit{Diffusion} & \textit{Diffusion} & \textit{Diffusion} & \textit{Diffusion} & \textit{AR-Trust} & \textit{Diffusion-Trust} & \textit{AR} \\
\midrule
\rowcolor{taskshade}
\multicolumn{10}{l}{\textbf{\textit{Knowledge \& Instruction Following}}} \\
ARC-Challenge   & ---   & 93.56          & ---            & \underline{95.93} & ---               & 93.2 & \textbf{96.33} & 95.65          & 97.70 \\
MMLU            & ---   & 80.53          & ---            & \textbf{87.69}    & ---               & 82.8 & \underline{84.80} & 80.75          & 88.21 \\
MMLU-Pro        & ---   & 63.22          & 63.42          & 73.36             & \underline{75.31} & 61.5 & \textbf{77.39} & 74.73          & 81.39 \\
GPQA-Diamond    & 73.00 & 47.98          & 48.36          & 61.98             & \underline{66.67} & 36.7 & \textbf{71.21} & 64.65          & 80.30 \\
IFEval          & ---   & 80.78          & 81.33          & \underline{81.70} & \textbf{83.36}    & 60.6 & 71.35          & 63.22          & 91.31 \\
\midrule
\rowcolor{taskshade}
\multicolumn{10}{l}{\textbf{\textit{Math}}} \\
GSM8K           & 90.62 & \underline{94.24} & ---         & \textbf{96.06}    & ---               & 91.4 & 93.33          & 93.10          & 89.16 \\
MATH-500        & 81.00 & ---            & ---            & ---               & ---               & 77.8 & \textbf{95.20} & \underline{93.60} & 96.60 \\
AIME-24         & 51.10 & ---            & ---            & ---               & ---               & 16.7 & \textbf{63.33} & \underline{60.00} & 65.56 \\
AIME-25         & ---   & 36.67          & 36.67          & \underline{60.00} & \textbf{63.33}    & 10.8 & 54.44          & 53.33          & 60.00 \\
\midrule
\rowcolor{taskshade}
\multicolumn{10}{l}{\textbf{\textit{Code}}} \\
HumanEval       & ---   & 86.59          & ---            & \textbf{94.51}    & ---               & 87.2 & \underline{92.07} & 82.32\textsuperscript{*} & 95.12 \\
MBPP            & ---   & 81.50          & ---            & \underline{88.29} & ---               & 71.6 & \textbf{91.05} & 82.10\textsuperscript{*} & 89.11 \\
LiveCodeBench~v6 & 67.30 & 31.50          & 28.85          & 42.29             & \underline{44.05} & 21.7 & \textbf{49.71} &  5.71\textsuperscript{*} & 49.71 \\
\bottomrule
\end{tabular}
}
\end{table}

\begin{table}[!htbp]
\centering
\scriptsize
\setlength{\abovecaptionskip}{2pt}
\setlength{\belowcaptionskip}{3pt}
\setlength{\tabcolsep}{2.7pt}
\renewcommand{\arraystretch}{0.92}
\caption{\textbf{Benchmark performance of \model-2B and
\model-4B}. \textbf{Bold}: best open-source dLLM; \underline{underline}: second best; \textsuperscript{*}: potentially under-reported.}
\label{tab:flare-small}
\resizebox{\linewidth}{!}{%
\begin{tabular}{lccc|ccc|ccc}
\toprule
 & \textbf{SDAR} & \textbf{SDAR} & \textbf{SDAR} & \textbf{\model} & \textbf{\model} & \textbf{Qwen3.5} & \textbf{\model} & \textbf{\model} & \textbf{Qwen3.5} \\
 & \textbf{1.7B} & \textbf{4B}   & \textbf{8B}   & \textbf{-2B}    & \textbf{-2B}    & \textbf{-2B}     & \textbf{-4B}    & \textbf{-4B}    & \textbf{-4B} \\
\textit{Params} & \textit{1.7B} & \textit{4B} & \textit{8B} & \textit{2B} & \textit{2B} & \textit{2B} & \textit{4B} & \textit{4B} & \textit{4B} \\
\textit{Sampling mode} & \textit{Diffusion} & \textit{Diffusion} & \textit{Diffusion} & \textit{AR-Trust} & \textit{Diffusion-Trust} & \textit{AR} & \textit{AR-Trust} & \textit{Diffusion-Trust} & \textit{AR} \\
\midrule
\rowcolor{taskshade}
\multicolumn{10}{l}{\textbf{\textit{Knowledge \& Instruction Following}}} \\
ARC-Challenge    & 85.4 & 90.5 & 91.9              & 85.07 & 85.84                    & 92.15                    & \underline{93.52} & \textbf{94.62}           & 96.33 \\
MMLU             & 62.9 & 74.9 & 78.6              & 67.60 & 64.14                    & 73.59                    & \underline{78.73} & \textbf{79.54}           & 85.22 \\
MMLU-Pro         & 37.0 & 50.9 & 56.9              & 53.57 & 53.63                    & 59.53                    & \textbf{71.14}    & \underline{70.95}        & 77.88 \\
GPQA-Diamond     & 29.8 & 33.0 & 40.2              & 37.37 & 35.35                    & 62.12                    & \underline{63.64} & \textbf{64.65}           & 80.30 \\
IFEval           & 43.4 & 56.6 & 61.4              & 68.95 & 62.66                    & 79.48                    & \underline{73.20} & \textbf{73.57}           & 90.02 \\
\midrule
\rowcolor{taskshade}
\multicolumn{10}{l}{\textbf{\textit{Math}}} \\
GSM8K            & 80.1 & 89.9 & \underline{91.3}  & 84.46 & 82.79                    & 77.63\textsuperscript{*} & 91.05             & \textbf{91.58}           & 89.16 \\
MATH-500         & 63.2 & 72.8 & 78.6              & 84.40 & 82.20                    & 72.20\textsuperscript{*} & \textbf{94.20}    & \underline{91.60}        & 95.40 \\
AIME-24          & 10.0 & 10.0 & 10.0              & 31.11 & 31.11                    &  8.89\textsuperscript{*} & \textbf{58.89}    & \underline{55.56}        & 63.33 \\
AIME-25          &  2.1 &  7.5 & 10.0              & 26.67 & 26.67                    & 12.22\textsuperscript{*} & \underline{43.33} & \textbf{46.67}           & 48.89 \\
\midrule
\rowcolor{taskshade}
\multicolumn{10}{l}{\textbf{\textit{Code}}} \\
HumanEval        & 61.6 & 72.8 & 78.7              & 64.02 & 50.61\textsuperscript{*} & 48.17                    & \textbf{93.29}    & \underline{83.54}\textsuperscript{*} & 87.80 \\
MBPP             & 61.1 & 65.4 & 72.0              & 68.09 & 55.25\textsuperscript{*} & 53.31                    & \textbf{89.11}    & \underline{77.82}\textsuperscript{*} & 82.49 \\
LiveCodeBench~v6 &  5.7 & 13.1 & \underline{16.6}  & 15.43 &  9.71\textsuperscript{*} & 17.71                    & \textbf{41.71}    & 12.57\textsuperscript{*} & 50.86 \\
\bottomrule
\end{tabular}
}
\end{table}

At the 9B scale, \model-9B matches or exceeds LLaDA-2.1-flash on most shared benchmarks, despite using roughly $1/10$ of its total parameters: GPQA-Diamond ($71.21$ vs.\ $66.67$), MMLU-Pro ($77.39$ vs.\ $75.31$), MBPP ($91.05$ vs.\ $88.29$), and LiveCodeBench~v6 ($49.71$ vs.\ $44.05$). Compared with the commercial Mercury-2 system, \model-9B is clearly stronger on math-reasoning benchmarks (MATH-500 by $+14.2$ points, AIME-24 by $+12.2$ points); we note in fairness a remaining gap on LiveCodeBench~v6 ($49.71$ vs.\ $67.30$), reflecting Mercury-2's additional training advantage on code generation.

More directly, relative to the source-model Qwen3.5-9B, \model-9B retains the majority of the original capability across most tasks: $98.6\%$ on MATH-500 ($95.20$ vs.\ $96.60$), $96.6\%$ on AIME-24 ($63.33$ vs.\ $65.56$), and $95.1\%$ on MMLU-Pro ($77.39$ vs.\ $81.39$), while \emph{exceeding} the original AR checkpoint on MBPP ($91.05$ vs.\ $89.11$). The conversion thus preserves high-level reasoning and coding ability rather than simply improving easier short-form benchmarks. Together, the 9B results support both aspects of question~(i) at the start of this section: parameter efficiency relative to substantially larger dLLM baselines, and capability retention relative to the same-family AR checkpoint.

\FloatBarrier

The same pattern extends to the smaller scales. \model-4B outperforms SDAR-30B-A3B on nearly all reported benchmarks despite using a much smaller dense backbone, with the largest margins on GPQA-Diamond ($63.64$ vs.\ $36.7$), MATH-500 ($94.20$ vs.\ $77.8$), and AIME-24 ($58.89$ vs.\ $16.7$). At the 2B scale, \model-2B improves substantially over the parameter-matched SDAR-1.7B on math and code tasks, including AIME-24 ($31.11$ vs.\ $10.0$) and MATH-500 ($84.40$ vs.\ $63.2$). Relative to the original Qwen3.5 checkpoints at both scales, the converted models retain a large fraction of the inherited capability, further indicating that the method functions as capability-preserving transfer rather than training a new dLLM from scratch.

Each \model row is reported under both AR-Trust and Diffusion-Trust sampling, but the two columns correspond to a single trained checkpoint that only switches its sampling path. The two modes are close on math and on most knowledge benchmarks (with larger AR-Trust advantages on GPQA-Diamond and IFEval at 9B), while Diffusion-Trust is consistently weaker on code-generation tasks. We attribute this gap to two factors: diffusion-style decoding is more sensitive to brittle answer extraction (which fails under minor syntax variations) and long-output truncation under the current code-evaluation scripts, but the parallel-decoding interface also faces an inherent challenge on long, strictly structured outputs such as multi-function code, where each block commits without left-to-right syntactic verification. Results marked with \textsuperscript{*} should therefore be interpreted conservatively. Even with this caveat, the comparison supports the feasibility side of question~(iii): a single \model checkpoint can support both a clean-stream verified causal sampling path and a block-diffusion sampling path, without training separate models for different serving regimes.

A residual gap to the original AR checkpoints remains, especially on instruction following and some coding benchmarks. We attribute this partly to the post-trained nature of the Qwen3.5 source models: continued SFT with external data drawn from a different distribution than the original post-training data can shift the output distribution away from the source checkpoint, consistent with the data-quality findings of Section~\ref{sec:study}. Future conversions may therefore benefit from transfer data more tightly matched to the source AR model's own reasoning and instruction-following style. With capability retention established, the next subsection turns to question~(ii): whether this two-stream supervision can be executed on a hybrid backbone at acceptable training efficiency.

\subsection{Efficient Training on Hybrid Backbones}
\label{sec:flare-kernel}

This subsection addresses question~(ii): whether the transfer configuration can run on a hybrid-attention backbone at hardware efficiency close to pure-AR training. The clean/noisy mask in Section~\ref{sec:method-training} defines a causal clean stream with block-bidirectional noisy streams. In softmax-attention layers this is just a visibility mask over key-value pairs, but in recurrent linear-attention layers visibility is encoded by the state trajectory itself: a noisy block must be initialized from the correct clean boundary state, expose its tokens to one another, and avoid leaking into unrelated blocks or packed documents. The same algorithmic mask thus becomes a state-scheduling problem inside the hybrid backbone.

On the Qwen3.5 backbone, this state-scheduling problem affects two sequence-mixing components in each GDN layer: the GDR recurrence and the width-$W$ causal ShortConv that prepares its inputs, imposing three requirements absent in pure AR training, which sees one token-causal stream. (1)~\textit{Gated Delta Rule}: every noisy block must start from the clean state at its preceding block boundary; these mid-chunk states scale as $L/B$ in a direct implementation and dominate memory at the small block sizes \model uses. (2)~\textit{1D causal ShortConv}: noisy tokens must read both in-block noisy lags and clean-context lags from before the boundary, realized directly via $L/B$ small block-level convolution launches. (3)~\textit{Document packing}: packed examples must stay isolated despite recurrent state propagation and convolutional lag reads.

We compare two realizations of this schedule, sharing clean/noisy visibility semantics but differing in how block-boundary states are produced and consumed. (1)~\emph{Route~I (chunk-then-refine)} computes the clean stream first, materializes every block-boundary clean state $\mathbf{S}_{(b-1)B}$ in HBM, then runs each noisy block as a seeded local recurrence; it serves as a correctness reference and reuses the standard AR kernel structure, but introduces $L/B$-scaling state hand-offs. (2)~\emph{Route~II (fused two-stream)} stores only strided clean-state checkpoints, reconstructs the required boundary state in registers, immediately consumes the corresponding noisy block, and injects block-level gradients back into the clean recurrence without materializing intermediate boundary states; the ShortConv branch fuses lag selection, document-boundary masking, and noisy-block convolution into one kernel, using \texttt{cu\_seqlens}-aware guards to reset noisy states, block cross-document gradient shifts, and zero ShortConv reads crossing document boundaries. Derivations and route-level pseudocode appear in Appendix~\ref{app:kernel-routes}.

\begin{figure}[!htbp]
\centering
\includegraphics[width=\linewidth]{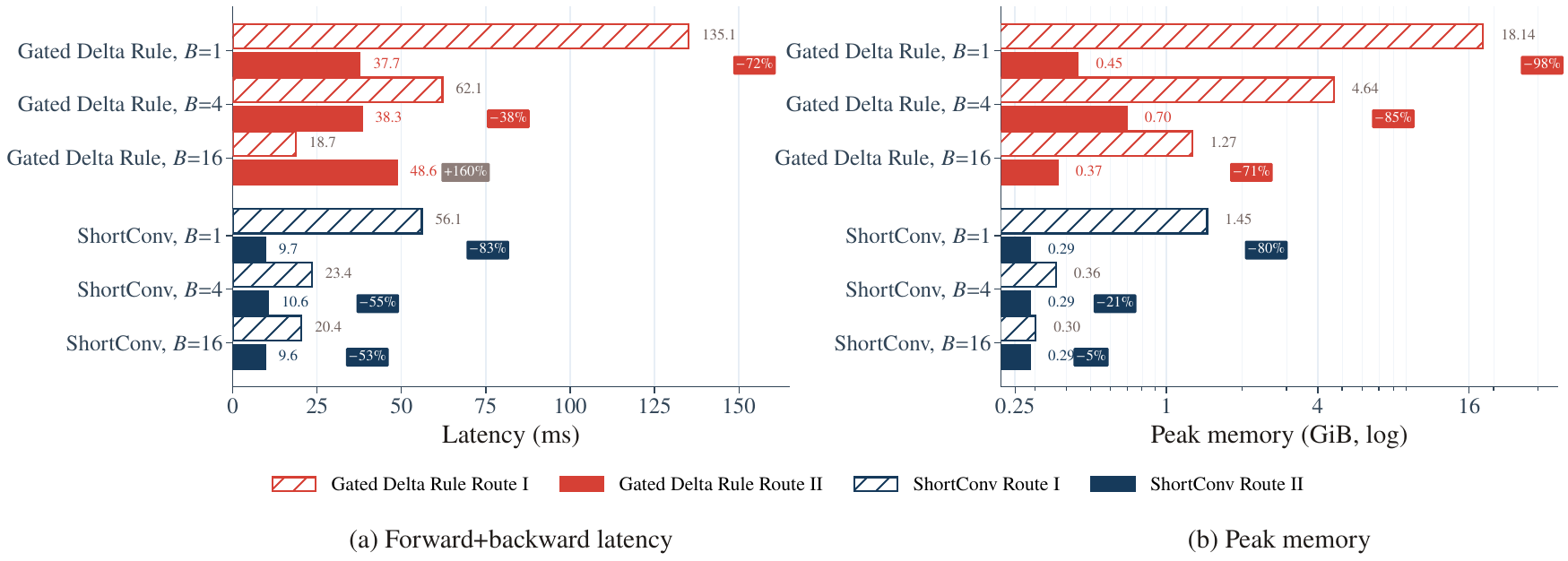}
\caption{\textbf{Route-level kernel microbenchmark on Qwen3.5-2B shapes.} Horizontal paired bars compare Route~I and Route~II for the Gated Delta Rule and ShortConv kernels across diffusion block sizes $B\in\{1,4,16\}$. Labels on Route~II report the relative change from Route~I. Route~II sharply reduces latency and memory in the small-block regime, while Route~I overtakes the Gated Delta Rule at $B{=}16$ as dense chunk-level matmul better saturates tensor cores. Full sweeps are in Appendix~\ref{app:kernel-routes}.}
\label{fig:flare-route-main}
\end{figure}

We first evaluate the schedules at the kernel level. Figure~\ref{fig:flare-route-main} shows Route~II is most effective in the small-block regime diffusion training requires. For the GDR at $B{=}1$, it cuts total latency from $135.10$ ms to $37.69$ ms and peak memory from $18.14$ GiB to $0.45$ GiB. At larger $B$, Route~I can win because its dense chunk-level matmul better saturates tensor cores. Since diffusion training requires small block sizes (\model uses $B=4$), Route~II is a structural requirement rather than an optimization: without it the small-$B$ regime is infeasible in memory. We therefore use Route~II for \model's small diffusion blocks and retain Route~I as the large-block path. For ShortConv, Route~II stays faster across all measured block sizes by avoiding many small block-level launches.

\begin{figure}[!htbp]
\centering
\begin{minipage}[c]{0.49\linewidth}
\centering
\includegraphics[width=\linewidth]{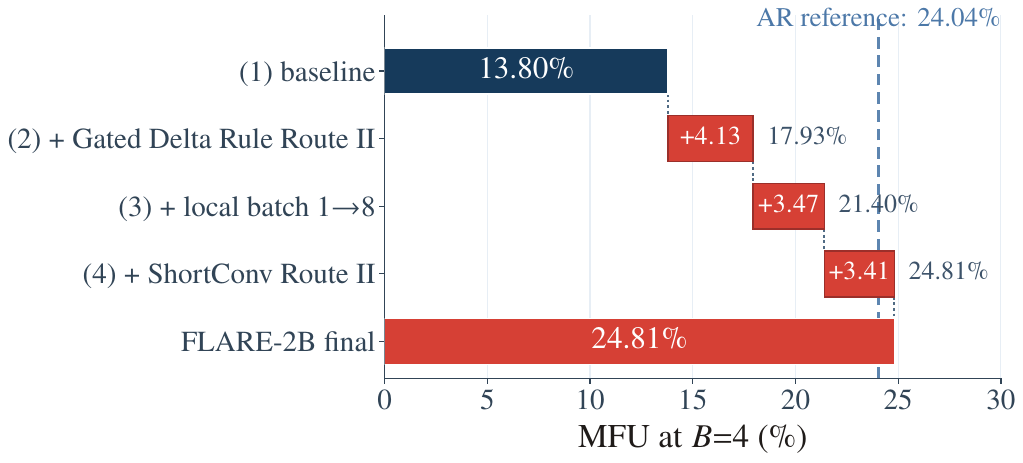}
\end{minipage}\hfill
\begin{minipage}[c]{0.49\linewidth}
\caption{Training-time MFU of \model-2B at $B{=}4$ on $8{\times}$A100-80GB (bf16), as a horizontal waterfall. Bar~(1) is the unoptimized baseline (Route~I for both kernels, local batch $1$); bars~(2)--(4) show the cumulative MFU lift after each kernel-stack change, with per-step gains in percentage points. The final bar is the \model setting, closing the gap to the AR Qwen3.5-2B reference ($24.04\%$). Full sweep and per-block HBM in Tab.~\ref{tab:app-mfu-ablation} of Appendix~\ref{app:mfu-ablation}.}
\label{fig:flare-mfu-main}
\end{minipage}
\end{figure}

Figure~\ref{fig:flare-mfu-main} shows these kernel-level gains carry to end-to-end training, measured as model FLOPs utilization (MFU). Replacing the GDR path with Route~II removes the small-$B$ memory failure and raises MFU at $B{=}4$ from $13.80\%$ to $17.93\%$. The recovered memory headroom supports a larger per-GPU local batch, lifting MFU to $21.40\%$. Adding the fused ShortConv path brings \model-2B to $24.81\%$ MFU at $B{=}4$, \emph{matching and slightly exceeding} the pure-AR Qwen3.5-2B reference ($24.04\%$) on the same hardware. Since two-stream supervision processes more work per step than single-stream AR training, the cost of recurrent-state scheduling is fully absorbed by the hardware-aware implementation rather than becoming the dominant training bottleneck. This affirmatively answers question~(ii); with training efficiency established, the next subsection turns to question~(iii): how to support both decoding paths under one unified serving stack.

\FloatBarrier

\subsection{Unified Serving for AR-Trust and Diffusion-Trust}
\label{sec:flare-decoding}

This subsection addresses question~(iii): whether a single serving stack can support both decoding paths efficiently. Inference is the other stage where the hybrid backbone changes the standard dLLM serving problem. A pure softmax Transformer can treat speculative decoding largely as KV-cache management, while \model must coordinate a softmax KV cache, a GDN recurrent state, and the clean/noisy attention patterns used by the two decoding paths. We implement this in an SGLang-based serving stack~\citep{zheng2024sglang} that runs the AR-Trust and Diffusion-Trust interfaces of Section~\ref{sec:method-inference} from the same checkpoint, with path-specific masks and shared hybrid-attention kernels. Full pseudocode for both paths is given in Appendix~\ref{app:inference}; the hybrid-specific serving machinery is detailed in Appendix~\ref{app:inference-machinery}.

\begin{figure}[!htbp]
\centering
\includegraphics[width=\linewidth]{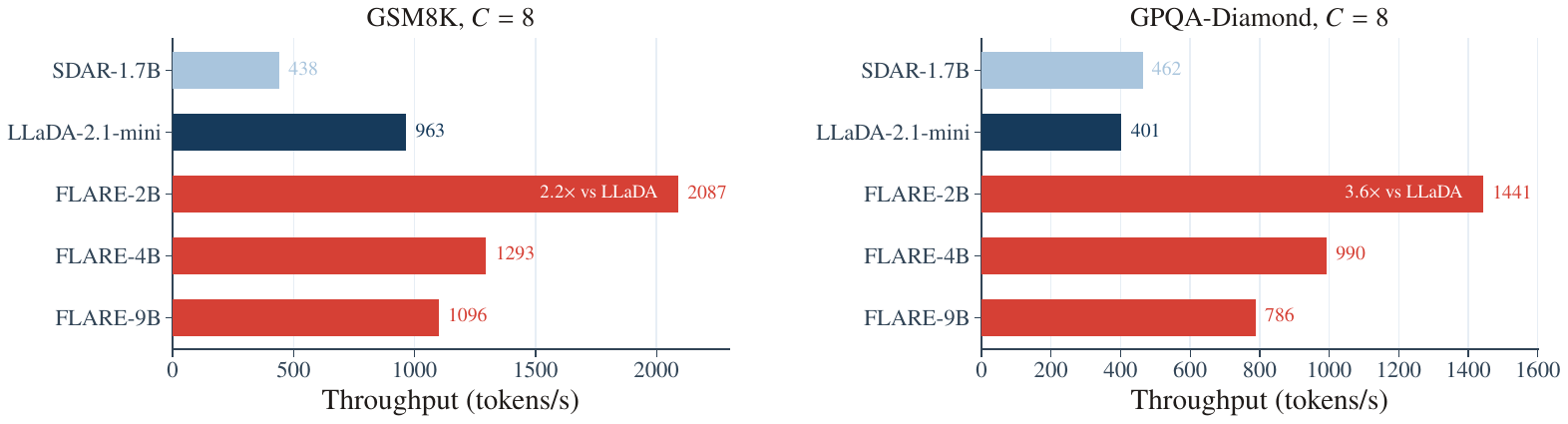}
\caption{High-concurrency fixed-output throughput. Bars report tokens/s at $C{=}8$ on $1{\times}$A100-80GB with bf16 and the SGLang serving stack; \texttt{max\_new\_tokens}$=2048$ and \texttt{ignore\_eos}$=$true. The full $3{\times}3$ grid over benchmarks and $C\in\{1,4,8\}$ is in Table~\ref{tab:flare-tps} in Appendix~\ref{app:per-benchmark}.}
\label{fig:flare-tps-main}
\end{figure}

Figure~\ref{fig:accept-rate-speed} gives the analytical interface-level trade-off between AR-Trust and Diffusion-Trust. Here we focus on the serving mechanisms that turn those regimes into efficient execution on a hybrid-attention backbone; measured tokens-per-second (TPS) is reported later in Figure~\ref{fig:flare-tps-main}.

The serving stack is built around four mechanisms beyond a standard speculative-decoding engine. (1)~\textit{Recurrent-state commit}: accepting only $r$ of $K$ speculative tokens is not just a KV tail trim, since the GDN state after the verify round must also be rewound to $\mathbf{S}^{(r)}$. (2)~\textit{Native dLLM mask modes}: dense custom masks impose unnecessary per-score overhead, so \model encodes the clean/noisy row types compactly and uses a prefix-tile fast path whenever a KV tile lies entirely in the causal prefix. (3)~\textit{Fused verification and top-$k$ kernels}: AR-Trust verification avoids materializing full $[M,V]$ logits or probability tensors, which is important at large vocabularies. (4)~\textit{CUDA-graph replay safety}: graph reuse must match not only tensor shapes but also block size, mask mode, recurrent-state update mode, and logits-output mode; otherwise a graph captured for one decoding path can be incorrectly replayed for another.

The most hybrid-specific of the four is mechanism~(1). To make partial-accept rewind efficient, \model records the intermediate verify-position states inside the same recurrent kernel and commits the accepted offset with one fused gather-scatter kernel (Figure~\ref{fig:accepted-state-commit} in Appendix~\ref{app:inference}). This avoids replaying accepted tokens through every GDN layer after each verify decision. Diffusion-Trust uses a related separation between denoise and commit: denoise passes read the recurrent state but do not write it back, and a final causal state-update pass commits the completed block to the recurrent pool. The reason is that tokens in intermediate denoise rounds may still be revised in subsequent rounds, so writing them back early would contaminate the recurrent state trajectory used by later blocks.

\FloatBarrier

Figure~\ref{fig:flare-tps-main} reports the high-concurrency headline setting, where the system mechanisms above translate the parallel decoding advantage into real tokens-per-second. On a single A100-80GB at $C{=}8$, \model-2B reaches $2{,}087$ tokens/s on GSM8K, giving a $2.2\times$ gain over LLaDA-2.1-mini and a $4.8\times$ gain over SDAR-1.7B. On GPQA-Diamond, it reaches $1{,}441$ tokens/s, a $3.6\times$ gain over LLaDA-2.1-mini. The gap is largest at high concurrency, the regime in which per-step overhead dominates and the fused kernels are most valuable for keeping the hybrid recurrent state out of the critical path. \model-4B and \model-9B are slower in absolute throughput, as expected from their larger backbones, but remain competitive with larger dLLM baselines while providing the capability gains reported above. Full throughput results across three benchmarks and $C\in\{1,4,8\}$ are reported in Table~\ref{tab:flare-tps} in Appendix~\ref{app:per-benchmark}.

Together, the three subsections above answer the three questions posed at the start of this section: (i) the converted \model checkpoints at the 2B/4B/9B scales preserve the capability of their AR sources and exceed substantially larger dLLM baselines under a more constrained conversion budget; (ii) two-stream supervision reaches MFU comparable to, and slightly above, pure-AR training under our hardware-aware kernel implementation; and (iii) a single checkpoint supports both AR-Trust and Diffusion-Trust decoding under one unified SGLang stack, with substantial end-to-end throughput advantages over other dLLMs at high concurrency.

\FloatBarrier

\section{Conclusion, Limitations, and Future Work}
\label{sec:conclusion}

\paragraph{Conclusion.}
We presented \model, a systematic recipe for converting strong hybrid-attention AR models into capable, high-throughput diffusion LLMs. The work contributes several mutually reinforcing designs: a token-balanced clean/noisy \emph{two-stream training objective} that unifies AR next-token supervision and block-diffusion denoising in one forward pass; a document-packed clean/noisy \emph{attention-mask design} that supplies both causal and block-bidirectional training signal without cross-document leakage; a study of \emph{transfer onto hybrid-attention backbones}, where we work out how to realize this objective on mixed softmax/linear-attention architectures through recurrent-state scheduling and matching training kernels, reaching training efficiency comparable to pure-AR training; and a \emph{unified inference system} in which a single checkpoint supports multiple diffusion decoding modes, from parallel denoising to causal verified decoding. On top of these, our controlled study further identifies \emph{transfer-data quality} as one of the key factors for capability preservation. Together, these results show that, under a modest conversion budget of roughly 10B tokens, AR-to-dLLM transfer can both retain the source model's capability and deliver real parallel-decoding throughput, indicating that practical dLLMs are limited less by the decoding algorithm itself than by the joint design of objective, data, and inference system.

\paragraph{Limitations.}
We group the limitations of this work into three points. \textbf{(i) Training overhead.} Block-diffusion training concatenates a clean and a noisy view of every sequence into a single $2L$-length input under a custom attention mask, which roughly doubles the per-step compute and memory relative to size-matched AR training and makes long-context training more expensive; the cost is amplified on hybrid backbones, where linear-attention layers require the noisy stream to read clean intermediate recurrent states at every block boundary --- the root cause of the residual training-efficiency gap in Section~\ref{sec:flare-kernel}. Our dedicated kernels mitigate but do not eliminate this underlying $2\times$ overhead. \textbf{(ii) Residual gap to the source model.} Even with our best data mix, \model still trails its AR source model on several benchmarks. This is consistent with source-distribution shift: continuing SFT on long-CoT traces from \emph{external} teachers (DeepSeek-R1, Qwen3, GPT-OSS-120B) moves the output distribution away from the post-trained Qwen3.5 seed. Notably, more aggressive \emph{filtering} via our automatic IFD-based selection pipeline (Appendix~\ref{app:data-selection}) does not close this gap, indicating that the bottleneck is a distribution mismatch between the transfer data and the source model rather than insufficient data filtering, and pointing to data more tightly aligned with both the source AR teacher and the diffusion objective. \textbf{(iii) Limited scale and post-training scope.} We validate \model only on dense checkpoints below 10B parameters and a single SFT stage; MoE backbones and post-training beyond SFT (e.g., reinforcement learning) remain untested, and whether they preserve capability under the same low conversion budget is open.

\paragraph{Future work.}
These limitations outline open directions for the next generation of dLLM systems~\citep{li2025dlmsurvey}. The most fundamental is to move beyond the concatenated two-stream formulation toward \emph{single-stream training} in which clean and noisy supervision share one forward pass, lowering this overhead along two complementary routes: the backbone side, as in DiffuMamba~\citep{singh2025diffumamba}, and the training-cost side, as in Orthrus~\citep{nguyen2026orthrus} (tuning only $\sim$16\% of parameters). On the data side, the next step is transfer data more tightly aligned with the source-model distribution and the diffusion objective --- e.g., diffusion-friendly traces harvested from the same-family AR teacher and matched to the block structure --- rather than only finer filtering of existing external corpora. On the scaling side, the recipe can be used to up-cycle dense AR checkpoints into MoE-dLLMs, where LLaDA-MoE~\citep{zhu2025lladamoe} already shows feasibility, raising the question of how expert routing interacts with the diffusion mask. Finally, closing the residual quality gap will likely require RL: the per-sequence likelihood of a masked-diffusion model is intractable, so policy-gradient methods need ELBO surrogates~\citep{zhao2025d1, zhu2025enhancing, wang2025spg}, and \model's multiple decoding paths raise the further question of which path to roll out under and how to credit-assign reward across the clean and noisy streams.

\clearpage
\newpage
\bibliographystyle{assets/plainnat}
\bibliography{main}

\newpage
\appendix
\section{Hardware-Aware Kernels for Hybrid-Backbone Diffusion Training}
\label{app:kernel-routes}

This appendix specifies the kernel-level implementation of the
two-stream recurrence defined in
Section~\ref{sec:method-training}: its forward computation on the
hybrid GDN backbone, its gradients, and the shipped
implementation.  A GDN layer contains two sub-components that both
require re-implementation under \model's two-stream training: the
chunkwise-parallel GDR recurrence, and the width-$W$ depthwise
\emph{1D Causal ShortConv} that feeds it.  The rest of the appendix
proceeds as follows.
\begin{itemize}\itemsep3pt
\item \S\ref{app:kernel-problem} states the three challenges that
separate \model's two-stream training from single-stream training.
\item \S\ref{app:gdn-routes} develops the GDR kernels: it reviews
the single-stream chunkwise form, lifts it to two streams via the
block-boundary clean seed and block-level chain rule, derives the
backward primitives, and specifies and compares the two
implementation routes.
\item \S\ref{app:short-conv} maps the same construction onto the
1D Causal ShortConv at lower cost, again as two implementation
routes.
\item \S\ref{app:doc-packed} adds the document-level guards that
keep packed samples isolated under recurrent-state propagation.
\item \S\ref{app:mfu-ablation} reports the end-to-end training-MFU
gains of the shipped kernel stack on \model-2B.
\end{itemize}

\subsection{Why the two-stream case needs a new kernel}
\label{app:kernel-problem}

Three challenges that \model's training setup imposes on both
sub-components of a GDN layer (the GDR
recurrence and the 1D Causal ShortConv) force the design
choices of \S\ref{app:gdn-routes}--\S\ref{app:doc-packed}.
Standard chunkwise-parallel GDR kernels and standard
causal 1D convolution kernels, as shipped by single-stream
linear-attention implementations, provide no mechanism for these
requirements, since each assumes a single token-causal stream.

\paragraph{Challenge 1: clean and noisy streams need asymmetric visibility.}
\model's objective (Eq.~\eqref{eq:flare-loss}) couples two streams:
clean tokens are token-causal over the whole clean stream, whereas a
noisy token at position $\ell$ in block $b$ sees only (i) noisy
tokens in the same block and (ii) clean tokens strictly before the
block start $(b{-}1)B$.  Both sub-components must realize this
asymmetry.  The GDR seeds the noisy recurrence on block $b$ from the
block-boundary clean state $\mathbf{S}_{(b-1)B}$; since this state
lies \emph{inside} a chunk whenever the block size $B$ differs from
the chunk size $C$, a correct implementation must compute $L/B$
mid-chunk clean states that the single-stream kernel never produces.
The 1D Causal ShortConv instead splits its receptive field: a noisy
output at offset $j \in \{0, \ldots, B{-}1\}$ from the block start
reads its first $j{+}1$ lags from the noisy stream and the remaining
$W{-}1{-}j$ lags from the clean stream, whereas a standard causal
convolution reads all $W$ lags from one stream.

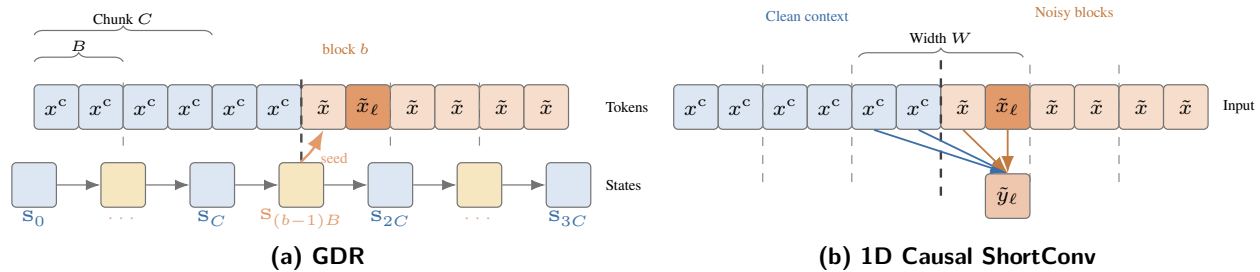
\begin{figure}[H]
\centering
\definecolor{c1clean}{HTML}{DCE7F3}
\definecolor{c1state}{HTML}{F6E7C1}
\definecolor{c1noisy}{HTML}{F7DDCB}
\definecolor{c1active}{HTML}{E29A6B}
\definecolor{c1lagN}{HTML}{C17A40}
\definecolor{c1lagC}{HTML}{3B6EA8}

\tikzset{
  c1tok/.style={draw=black!55, rounded corners=1.5pt, minimum size=0.52cm,
                inner sep=0pt, font=\scriptsize},
  c1brace/.style={decorate, decoration={brace, amplitude=3pt, raise=1pt}, black!55},
}
\begin{subfigure}[t]{0.49\linewidth}\centering
\resizebox{\linewidth}{!}{%
\begin{tikzpicture}[font=\scriptsize, >=Latex, x=0.52cm, y=0.52cm]
  \useasboundingbox (-1.2,-2.7) rectangle (12.6,1.85);
  \foreach \p in {0,1,2,3,4,5} \node[c1tok, fill=c1clean] (t\p) at (\p,0) {$x^{\mathrm{c}}$};
  \foreach \p in {8,9,10,11} \node[c1tok, fill=c1noisy] (t\p) at (\p,0) {$\tilde{x}$};
  \node[c1tok, fill=c1noisy]  (t6) at (6,0) {$\tilde{x}$};
  \node[c1tok, fill=c1active] (t7) at (7,0) {$\tilde{x}_\ell$};
  \foreach \b in {2,8,10} \draw[black!45, thin, dashed] (\b-0.5,0.55) -- (\b-0.5,-0.95);
  \draw[black!70, thick, dashed] (5.5,0.62) -- (5.5,-1.95);
  \node[font=\tiny, text=c1lagN, anchor=south] at (6.5,0.95) {block $b$};
  \node[c1tok, fill=c1clean] (b0) at (-0.5,-1.75) {};
  \node[c1tok, fill=c1state] (m0) at (1.5,-1.75) {};
  \node[c1tok, fill=c1clean] (b1) at (3.5,-1.75) {};
  \node[c1tok, fill=c1state] (m1) at (5.5,-1.75) {};
  \node[c1tok, fill=c1clean] (b2) at (7.5,-1.75) {};
  \node[c1tok, fill=c1state] (m2) at (9.5,-1.75) {};
  \node[c1tok, fill=c1clean] (b3) at (11.5,-1.75) {};
  \foreach \a/\bb in {b0/m0,m0/b1,b1/m1,m1/b2,b2/m2,m2/b3}
    \draw[->,black!55,thin] (\a) -- (\bb);
  \draw[->,c1active,thick] (m1) .. controls +(0,0.55) and +(-0.3,-0.55) .. (t6.south);
  \node[font=\tiny, text=c1active, anchor=west] at (5.7,-1.15) {seed};
  \node[font=\tiny, text=c1lagC] at (-0.5,-2.5) {$\mathbf{S}_{0}$};
  \node[font=\tiny, text=c1lagC] at (3.5,-2.5) {$\mathbf{S}_{C}$};
  \node[font=\tiny, text=c1active] at (5.5,-2.5) {$\mathbf{S}_{(b{-}1)B}$};
  \node[font=\tiny, text=c1lagC] at (7.5,-2.5) {$\mathbf{S}_{2C}$};
  \node[font=\tiny, text=c1lagC] at (11.5,-2.5) {$\mathbf{S}_{3C}$};
  \node[font=\tiny, text=c1active] at (1.5,-2.5) {$\cdots$};
  \node[font=\tiny, text=c1active] at (9.5,-2.5) {$\cdots$};
  \draw[c1brace] (-0.5,0.95) -- node[above=1pt, font=\tiny, black] {$B$} (1.5,0.95);
  \draw[c1brace] (-0.5,1.55) -- node[above=1pt, font=\tiny, black] {Chunk $C$} (3.5,1.55);
  \node[anchor=west, font=\tiny, text=black] at (12.1,0) {Tokens};
  \node[anchor=west, font=\tiny, text=black] at (12.1,-1.75) {States};
\end{tikzpicture}%
}
\subcaption*{\textbf{(a) GDR}}
\label{fig:challenge1-gdr}
\end{subfigure}
\hfill
\begin{subfigure}[t]{0.49\linewidth}\centering
\resizebox{\linewidth}{!}{%
\begin{tikzpicture}[font=\scriptsize, >=Latex, x=0.52cm, y=0.52cm]
  \useasboundingbox (-1.2,-2.7) rectangle (12.6,1.85);
  \foreach \p in {0,1,2,3,4,5} \node[c1tok, fill=c1clean] (p\p) at (\p,0) {$x^{\mathrm{c}}$};
  \node[c1tok, fill=c1noisy]  (p6) at (6,0) {$\tilde{x}$};
  \node[c1tok, fill=c1active] (p7) at (7,0) {$\tilde{x}_\ell$};
  \node[c1tok, fill=c1noisy]  (p8) at (8,0) {$\tilde{x}$};
  \node[c1tok, fill=c1noisy]  (p9) at (9,0) {$\tilde{x}$};
  \node[c1tok, fill=c1noisy]  (p10) at (10,0) {$\tilde{x}$};
  \node[c1tok, fill=c1noisy]  (p11) at (11,0) {$\tilde{x}$};
  \foreach \b in {1.5,3.5,7.5,9.5} \draw[black!45, thin, dashed] (\b,0.95) -- (\b,-1.85);
  \draw[black!75, thick, dashed] (5.5,1.05) -- (5.5,-2.0);
  \node[font=\tiny, text=c1lagC, anchor=south] at (2.5,1.7) {Clean context};
  \node[font=\tiny, text=c1lagN, anchor=south] at (8.5,1.7) {Noisy blocks};
  \node[c1tok, fill=c1active!55] (y) at (7,-2.0) {$\tilde{y}_\ell$};
  \draw[->,semithick,c1lagC] (p4.south) -- (y.north);
  \draw[->,semithick,c1lagC] (p5.south) -- (y.north);
  \draw[->,semithick,c1lagN] (p6.south) -- (y.north);
  \draw[->,semithick,c1lagN] (p7.south) -- (y.north);
  \draw[c1brace] (3.65,1.05) -- node[above=2pt, font=\tiny, black] {Width $W$} (7.35,1.05);
  \node[anchor=west, font=\tiny, text=black] at (11.6,0) {Input};
\end{tikzpicture}%
}
\subcaption*{\textbf{(b) 1D Causal ShortConv}}
\label{fig:challenge1-shortconv}
\end{subfigure}

\caption{\textbf{Challenge~1: the clean/noisy asymmetric visibility
for the two GDN sub-components.}  \textbf{(a)~GDR.}  A single-stream
kernel materializes the clean state only at chunk boundaries
(\textcolor{c1lagC}{blue}, $\mathbf{S}_0,\mathbf{S}_C,\mathbf{S}_{2C},
\ldots$).  Each noisy block~$b$ (e.g.\ the highlighted
$\tilde{x}_\ell$) must instead be seeded from the block-boundary
clean state $\mathbf{S}_{(b-1)B}$ \emph{inside} a chunk
(\textcolor{c1active}{amber}) whenever $B\neq C$ (here $C{=}4$,
$B{=}2$); there are $L/B$ such mid-chunk states and the single-stream
kernel never produces them.  \textbf{(b)~1D Causal ShortConv.}  The
width-$W$ receptive field of a noisy output $\tilde{y}_\ell$ at
offset $j$ (here $W{=}4$, $j{=}1$) splits across the block boundary:
the first $j{+}1$ lags read the \textcolor{c1lagN}{noisy} stream
in-block, the remaining $W{-}1{-}j$ read the \textcolor{c1lagC}{clean}
stream before the boundary, unlike a standard convolution that reads
all $W$ lags from one stream.}
\label{fig:challenge1}
\end{figure}

\paragraph{Challenge 2: document packing requires state resets.}
For throughput, multiple documents are packed into each $L$-token
sequence, with per-document starts encoded as
$\mathbf{cu} = (c_0, c_1, \ldots, c_{N_{\text{doc}}})$; the packed
run must reproduce the per-document single-sequence result.  This
forces a document-level mask to compose with both the two-stream
mask and the chunkwise state each sub-component tracks.  For the GDR, the
noisy recurrence's initial state at each document's first block
must be zeroed rather than inherited from the previous document's
chunk-boundary state, and the backward's cross-chunk scan must zero
its state-gradient hand-off at document boundaries.  For the 1D
Causal ShortConv, any lag read crossing a document boundary must be
masked to zero instead of reading the previous document's trailing
tokens.  Single-stream kernels need none of these guards, since a
plain token-causal mask packed with a document-causal mask already
forbids cross-document dependencies.

\begin{figure}[H]
\centering
\definecolor{c2clean}{HTML}{DCE7F3}
\definecolor{c2noisy}{HTML}{F7DDCB}
\definecolor{c2active}{HTML}{E29A6B}
\definecolor{c2state}{HTML}{F6E7C1}
\definecolor{c2lagN}{HTML}{C17A40}
\definecolor{c2lagC}{HTML}{3B6EA8}
\definecolor{c2cut}{HTML}{C0392B}
\tikzset{
  c2tok/.style={draw=black!55, rounded corners=1.5pt, minimum size=0.52cm,
                inner sep=0pt, font=\scriptsize},
  c2brace/.style={decorate, decoration={brace, amplitude=3pt, raise=1pt}, black!55},
}
\begin{subfigure}[t]{0.49\linewidth}\centering
\resizebox{\linewidth}{!}{%
\begin{tikzpicture}[font=\scriptsize, >=Latex, x=0.52cm, y=0.52cm]
  \useasboundingbox (-1.2,-2.7) rectangle (12.6,1.85);
  \foreach \p in {0,1,2,3,4,5} \node[c2tok, fill=c2noisy] (t\p) at (\p,0) {$\tilde{x}$};
  \node[c2tok, fill=c2active] (t6) at (6,0) {$\tilde{x}$};
  \foreach \p in {7,8,9,10,11} \node[c2tok, fill=c2noisy] (t\p) at (\p,0) {$\tilde{x}$};
  \draw[c2cut, very thick, dashed] (5.5,0.95) -- (5.5,-2.15);
  \node[font=\tiny, text=c2lagC, anchor=south] at (2.5,1.0) {Doc 1};
  \node[font=\tiny, text=c2lagN, anchor=south] at (8.5,1.0) {Doc 2};
  \node[c2tok, fill=c2state] (s5) at (5,-1.7) {};
  \node[c2tok, fill=black!8, draw=black!40] (s6) at (6,-1.7) {$\mathbf{0}$};
  \draw[->,c2cut,thick] (s5.east) -- (6.0,-1.7) node[midway,above=0pt,font=\tiny,c2cut] {};
  \node[c2cut, font=\scriptsize] at (5.5,-1.7) {$\times$};
  \draw[->,c2active,thick] (s6.north) .. controls +(0,0.55) and +(0,-0.55) .. (t6.south);
  \node[font=\tiny, text=c2cut, anchor=west] at (6.4,-2.45) {seed reset to $\mathbf{0}$};
  \node[font=\tiny, text=c2lagC, anchor=east] at (5.6,-2.45) {Doc 1 end state};
\end{tikzpicture}%
}
\subcaption*{\textbf{(a) GDR noisy-seed reset}}
\label{fig:challenge2-gdr}
\end{subfigure}
\hfill
\begin{subfigure}[t]{0.49\linewidth}\centering
\resizebox{\linewidth}{!}{%
\begin{tikzpicture}[font=\scriptsize, >=Latex, x=0.52cm, y=0.52cm]
  \useasboundingbox (-1.2,-2.7) rectangle (12.6,1.85);
  \foreach \p in {0,1,2,3,4,5} \node[c2tok, fill=c2clean] (p\p) at (\p,0) {$x$};
  \node[c2tok, fill=c2active] (p6) at (6,0) {$\tilde{x}_\ell$};
  \foreach \p in {7,8,9,10,11} \node[c2tok, fill=c2noisy] (p\p) at (\p,0) {$\tilde{x}$};
  \draw[c2cut, very thick, dashed] (5.5,0.95) -- (5.5,-2.15);
  \node[font=\tiny, text=c2lagC, anchor=south] at (2.5,1.0) {Doc 1};
  \node[font=\tiny, text=c2lagN, anchor=south] at (8.5,1.0) {Doc 2};
  \node[c2tok, fill=c2active!55] (y) at (6,-2.0) {$\tilde{y}_\ell$};
  \draw[->,semithick,c2lagN] (p6.south) -- (y.north);
  \draw[->,dotted,c2cut] (p5.south) -- (y.north);
  \draw[->,dotted,c2cut] (p4.south) -- (y.north);
  \draw[->,dotted,c2cut] (p3.south) -- (y.north);
  \node[c2cut, font=\scriptsize] at (5.0,-1.1) {$\times$};
  \node[font=\tiny, text=c2cut, anchor=west] at (6.25,-1.15) {cross-doc lags masked};
  \draw[c2brace] (2.65,1.05) -- node[above=1pt, font=\tiny, black] {Width $W$} (6.35,1.05);
\end{tikzpicture}%
}
\subcaption*{\textbf{(b) ShortConv cross-doc lag mask}}
\label{fig:challenge2-shortconv}
\end{subfigure}

\caption{\textbf{Challenge~2: document packing requires state resets
at document boundaries.}  A packed sequence holds Doc~1 and Doc~2
back to back (red boundary).  \textbf{(a)~GDR.}  The noisy recurrence
for Doc~2's first block must be seeded from a \emph{zeroed} state
($\mathbf{0}$) rather than inheriting Doc~1's end-of-document state,
which the cross-document carry ($\times$) would otherwise propagate.
\textbf{(b)~1D Causal ShortConv.}  A noisy output $\tilde{y}_\ell$
near Doc~2's start has lags whose width-$W$ window reaches back
across the boundary; those cross-document lags are masked to zero
($\times$), leaving only the in-document lag.  Single-stream kernels
need none of these guards.}
\label{fig:challenge2}
\end{figure}
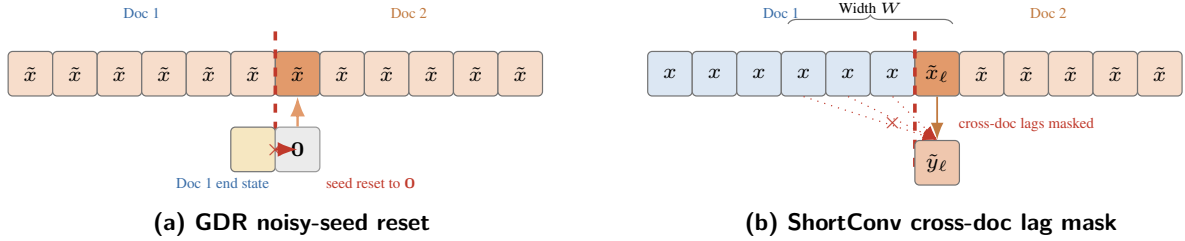

\paragraph{Challenge 3: composing three granularities in the backward.}
Realizing the Challenge~1 visibility is cheap in the forward---
either $L/B$ extra bytes in HBM or $\mathcal{O}(B)$ register-replay
steps per block for the GDR, and negligible for
ShortConv (whose ``state'' is a raw $W{-}1$-token slice).  The
difficulty concentrates in the backward.  A standard chunk-parallel
clean backward carries a \emph{single} cross-chunk gradient
$\mathbf{dh}_{[c]}\equiv \partial\mathcal{L}/\partial\mathbf{S}_{[c]}$
through a matmul scan, delivering all per-token clean gradients at
tensor-core throughput.  In the two-stream case the noisy backward
emits two further objects the chunk-level scan cannot directly
accept: per-block clean-transition gradients
(\S\ref{app:kernel-clean-trans-bwd}) and per-block initial-state
gradients $\mathbf{d}\tilde{\mathbf{S}}_{b,\text{init}}$.  Routing
them into the scan without breaking its parallelism requires
composing three granularities of reverse recurrence---token within
a block, block within a chunk, and chunk within the sequence---in
the correct order (\S\ref{app:kernel-bwd-primitives}).  The
ShortConv backward has only two of these granularities, because its
noisy-side ``state'' is a static clean-stream slice rather than a
recurrence terminus, which makes it cheaper than the GDR
's (\S\ref{app:short-conv-easier}).  The routes of
\S\ref{app:route-refine}--\S\ref{app:route-twostream} are two
schedules of this composition onto a GPU; the resulting
block-by-block backward dataflow is detailed in
Figure~\ref{fig:kernel-grad-flow}.

\paragraph{AR-only training does not trigger these challenges.}
AR-only training of a GDN layer runs a single stream
with uniform token-causal visibility.  Challenge~1 does not arise
because there is a single stream and no block-level granularity.
Challenge~2 reduces to standard document-causal attention masking,
which composes with the token-causal per-stream mask without any
additional kernel-side guard.  Challenge~3 reduces to a single
cross-chunk state-gradient tensor at chunk-level granularity,
which the standard chunkwise-parallel kernels of
\citet{yang2024gated,yang2025qwen3} already handle.
\model's block-seeded noisy-stream objective is what introduces
the machinery described below.

\paragraph{Notation.}
Table~\ref{tab:kernel-notation} collects the symbols used in the
remainder of this appendix.  Positions in a stream are indexed by
$t$ (clean) or $\ell$ (noisy).  The lengths $B$, $C$, $L$ and
derived counts $M$, $N_C$ satisfy $L = C \cdot N_C$ and
$C = B \cdot M$; we use $C{=}64$ throughout.

\begin{table}[!t]
\centering
\footnotesize
\caption{Symbols used in this appendix.  Clean-side quantities are
plain; noisy-side quantities carry a tilde, matching the main text.}
\label{tab:kernel-notation}
\begin{tabularx}{\linewidth}{@{}l X@{}}
\toprule
Symbol & Meaning \\
\midrule
\rowcolor{taskshade}
\multicolumn{2}{@{}l}{\textbf{\textit{Lengths, indices, and tiling}}} \\[1pt]
$L$                & sequence length (one stream) \\
$B$                & diffusion-block size \\
$C$                & chunk size for chunkwise-parallel training; $C=64$ \\
$M = C/B$          & blocks per chunk \\
$N_C = L/C$        & chunks per stream \\
$d_k, d_v$         & per-head key / value dimension \\
$H$                & number of linear-attention heads per layer \\
$b_v$              & kernel tile width in the value dimension ($b_v \le d_v$) \\
$c \in \{0,\ldots,N_C{-}1\}$ & chunk index \\
$b \in \{1,\ldots,L/B\}$     & block index; block $b$ covers positions $(b{-}1)B{+}1,\ldots,bB$ \\
\midrule
\rowcolor{taskshade}
\multicolumn{2}{@{}l}{\textbf{\textit{Per-position quantities}}} \\[1pt]
$x^{\mathrm{c}}_t \in \mathbb{R}^D$ & clean-stream input to the 1D causal ShortConv at position $t$ (\S\ref{app:short-conv}) \\
$\tilde{x}_\ell \in \mathbb{R}^D$ & noisy-stream input to the 1D causal ShortConv at position $\ell$ \\
$\mathbf{q}_t, \mathbf{k}_t, \mathbf{v}_t$ & clean query / key / value at position $t$ \\
$\tilde{\mathbf{q}}_\ell, \tilde{\mathbf{k}}_\ell, \tilde{\mathbf{v}}_\ell$ & noisy query / key / value at position $\ell$ \\
$\tilde g_\ell, \tilde\beta_\ell, \tilde{\boldsymbol\delta}_\ell$ & noisy log-gate / step-size / delta update at position $\ell$ \\
$g_t \in \mathbb{R}$       & scalar log-gate at position $t$ (clean); $\exp(g_t)$ is the multiplicative gate $\alpha_t$ of Eq.~\eqref{eq:gdn-recurrence} \\
$\beta_t \in \mathbb{R}$   & delta-rule step-size at position $t$ (clean) \\
$G_t = \sum_{i \le t} g_i$ & cumulative log-gate at position $t$ (clean) \\
$\exp\,(G_j - G_i)$        & decay factor from position $i$ to position $j$ (clean) \\
$u_t$                      & per-step prediction error $u_t = \mathbf{v}_t - (\mathbf{S}_t^{\text{pre}})^{\!\top}\mathbf{k}_t$ \\
$\boldsymbol\delta_t$      & per-step delta update $\boldsymbol\delta_t = \beta_t u_t$ \\
$\hat{\mathbf{v}}_t$       & chunk-corrected clean value (effective value produced by the chunkwise-parallel pre-step; see \S\ref{app:kernel-single-stream}) \\
\midrule
\rowcolor{taskshade}
\multicolumn{2}{@{}l}{\textbf{\textit{States and gradients}}} \\[1pt]
$\mathbf{S}_t$                 & clean state after position $t$ (after gating and delta update) \\
$\mathbf{S}^{\text{pre}}_t$    & clean state after gating, before the delta update, at position $t$ \\
$\tilde{\mathbf{S}}_\ell$      & noisy state after position $\ell$ \\
$\mathbf{S}_{[c]} \equiv \mathbf{S}_{cC}$ & chunk-$c$-boundary clean state \\
$\tilde{\mathbf{S}}_{b,\text{init}} \equiv \mathbf{S}_{(b-1)B}$ & block-$b$ noisy-recurrence initial state (seeded from clean) \\
$\mathbf{o}_t$                 & clean-stream output at position $t$ (Eq.~\eqref{eq:per-step-fwd}); the full-sequence clean output is the stack $\mathbf{o} = (\mathbf{o}_t)_{t=1}^{L}$ \\
$\tilde{\mathbf{o}}_\ell$      & noisy-stream output at position $\ell$ under the block-end readout (Eq.~\eqref{eq:noisy-block-fwd}); the full-sequence noisy output is $\tilde{\mathbf{o}} = (\tilde{\mathbf{o}}_\ell)_{\ell=1}^{L}$ \\
$\mathbf{O}_{[c]}$             & stacked per-token clean outputs of chunk $c$ (the chunk-level clean output tensor) \\
$\mathbf{do}_t \equiv \partial\mathcal{L}/\partial\mathbf{o}_t$ & output-gradient input to the clean-side backward at position $t$; $\mathbf{do} = (\mathbf{do}_t)_{t=1}^{L}$ denotes the full-sequence clean output gradient \\
$\mathbf{d}\tilde{\mathbf{o}}_\ell \equiv \partial\mathcal{L}/\partial\tilde{\mathbf{o}}_\ell$ & output-gradient input to the noisy-side backward at position $\ell$; $\mathbf{d}\tilde{\mathbf{o}} = (\mathbf{d}\tilde{\mathbf{o}}_\ell)_{\ell=1}^{L}$ denotes the full-sequence noisy output gradient \\
$\mathbf{dS}_t, \mathbf{d}\tilde{\mathbf{S}}_\ell$ & per-position state gradients \\
$\mathbf{dh}_b \equiv \partial\mathcal{L}/\partial\mathbf{S}_{bB}$ & clean state gradient at the end of block $b$ \\
$\mathbf{dh}^{\text{inject}}_{[c]}$ & chunk-$c$ within-chunk contribution to $\partial\mathcal{L}/\partial\mathbf{S}_{[c]}$ \\
\midrule
\rowcolor{taskshade}
\multicolumn{2}{@{}l}{\textbf{\textit{Block-level maps and checkpoint hyperparameters}}} \\[1pt]
$f^{\text{blk}}_b(\mathbf{S})$ & clean-state advance over block $b$ (Eq.~\eqref{eq:f-blk}) \\
$\boldsymbol\Delta_b$          & block $b$'s outer-product write term (Eq.~\eqref{eq:f-blk}) \\
$S \in \{1,\ldots,M\}$         & checkpoint stride: every $S$-th block-boundary clean state is stored (\S\ref{app:route-twostream}) \\
$N_{\text{ckpt}} = M/S$        & clean-state checkpoints per chunk; controls Route~II peak HBM \\
\bottomrule
\end{tabularx}
\end{table}

\subsection{Gated Delta Rule: two-stream training and implementation routes}
\label{app:gdn-routes}

This subsection develops the chunkwise-parallel training routes
for the GDR, the recurrence sub-component of the
GDN layer.  The objects and algorithms introduced here
(chunk-boundary states, WY-corrected values, block-level chain
rule, Route~I and Route~II) constitute the main technical content
of the appendix; \S\ref{app:short-conv} reuses the same structure
for the 1D Causal ShortConv at lower cost.

\subsubsection{Single-stream chunkwise training}
\label{app:kernel-single-stream}

Both routes reuse the standard chunkwise-parallel training form of
a single-stream GDR on the clean side unchanged.  We
restate it here to fix names for the four objects the clean-side
forward produces and both routes reuse: the clean output, the
chunk-boundary states, the chunk-corrected (WY) values, and the
cumulative log-gates.

\paragraph{Per-step recurrence.}
Expanding the GDR of Eq.~\eqref{eq:gdn-recurrence}
into its elementary per-step state and output updates yields the
form used in both optimization and differentiation:
\begin{equation}
\label{eq:per-step-fwd}
\begin{aligned}
  & \mathbf{S}^{\text{pre}}_t = \exp\,(g_t)\,\mathbf{S}_{t-1},
    & \text{(gating decay)}\\
  & u_t = \mathbf{v}_t - (\mathbf{S}^{\text{pre}}_t)^{\!\top}\mathbf{k}_t,
    & \text{(prediction error)}\\
  & \boldsymbol\delta_t = \beta_t\,u_t,
    & \text{(delta update)}\\
  & \mathbf{S}_t = \mathbf{S}^{\text{pre}}_t + \mathbf{k}_t\boldsymbol\delta_t^{\!\top},
    & \text{(outer-product write)}\\
  & \mathbf{o}_t = s\cdot\mathbf{S}_t^{\!\top}\mathbf{q}_t,
    & \text{(output)}
\end{aligned}
\end{equation}
where $\mathbf{S}^{\text{pre}}_t \in \mathbb{R}^{d_k \times d_v}$
is the state after the gating decay but before the delta update,
$u_t \in \mathbb{R}^{d_v}$ is the per-step prediction error between
the target value $\mathbf{v}_t$ and the value currently associated
with $\mathbf{k}_t$ under $\mathbf{S}^{\text{pre}}_t$,
$\boldsymbol\delta_t = \beta_t u_t \in \mathbb{R}^{d_v}$ is the
$\beta$-weighted correction written into the state by the
outer-product update, $\mathbf{S}_t \in \mathbb{R}^{d_k \times d_v}$
is the state after the update, $\mathbf{o}_t \in \mathbb{R}^{d_v}$
is the per-token output, and $s = 1/\sqrt{d_k}$ is the standard
query-key scaling.  The inputs $(g_t, \beta_t, \mathbf{q}_t,
\mathbf{k}_t, \mathbf{v}_t)$ are produced by the linear
projections and ShortConv branches of the GDN
layer and are the same as in Eq.~\eqref{eq:gdn-recurrence}.
\S\ref{app:kernel-per-step-bwd} differentiates the five lines.

\paragraph{Chunkwise-parallel form.}
A token-by-token implementation of Eq.~\eqref{eq:per-step-fwd} is
serial in the sequence dimension and underutilizes tensor cores.
A fully materialized $L \times L$ form reaches tensor cores but
incurs $\mathcal{O}(L^2)$ cost and loses the linear-attention
advantage.  The chunkwise-parallel form recovers both:
split the sequence into $N_C = L/C$ chunks of size $C$
(we set $C = 64$ to match tensor-core tile sizes), process each
chunk with dense matrix multiplication, and carry a single compact
state $\mathbf{S}_{[c]} \in \mathbb{R}^{d_k \times d_v}$ across
chunks.  Two pieces remain: how each chunk is processed in
parallel despite the delta rule's within-chunk dependence, and how
the compact state is advanced across chunks; the next two
paragraphs address them in turn.

\paragraph{Chunk-corrected values.}
The delta-rule correction at each token depends on the partial
state that earlier tokens in the same chunk have already written,
so running Eq.~\eqref{eq:per-step-fwd} token-by-token within a
chunk incurs $C$-step serial depth.  The WY representation of the
delta rule~\citep{yang2024parallelizing,yang2024gated}
removes this depth by replacing each raw value $\mathbf{v}_t$ with
a \emph{chunk-corrected} (or \emph{effective}) value
$\hat{\mathbf{v}}_t$, computed from
$\{(\mathbf{k}_t, \mathbf{v}_t, \beta_t, g_t)\}_{t\in\text{chunk}(c)}$
via a small triangular solve, so that the chunk's contribution to
the state reduces to a single outer-product sum.  We use
$\hat{\mathbf{v}}_t$ throughout; in particular, the chunk-level
state recurrence below is written directly in terms of
$\hat{\mathbf{v}}_t$.

\paragraph{Chunk-level state recurrence.}
Let $G_t = \sum_{i \le t} g_i$ denote the per-token
\emph{cumulative log-gate} on the clean stream, and let
$\mathbf{S}_{[c]} \equiv \mathbf{S}_{cC}$ denote the clean state
at the boundary between chunks $c-1$ and $c$.  In terms of
$\hat{\mathbf{v}}$ and $G$, the single-stream GDR
state recurrence over one chunk $c$ admits the closed form
\begin{equation}
\label{eq:chunk-state}
  \mathbf{S}_{[c+1]}
  \;=\;
  \exp\,\bigl(G_{(c+1)C}-G_{cC}\bigr)\,\mathbf{S}_{[c]}
  \;+\;
  \sum_{t\in\text{chunk}(c)}
    \exp\,\bigl(G_{(c+1)C}-G_t\bigr)\,\mathbf{k}_t\,\hat{\mathbf{v}}_t^{\!\top},
\end{equation}
where $\exp\,(G_{(c+1)C}-G_{cC})$ is the scalar decay factor
produced by compounding the $C$ per-token gates of chunk $c$, and
$\exp\,(G_{(c+1)C}-G_t)$ is the residual decay from position $t$
to the chunk end.  Within-chunk work reduces to a dense sum of
outer products (tensor-core friendly), and the only cross-chunk
serial dependency is the single $\mathbf{S}_{[c]} \to \mathbf{S}_{[c+1]}$
hand-off.  The chunk-level output $\mathbf{O}_{[c]}$ admits a
similar closed form from the chunk's queries, $\mathbf{S}_{[c]}$,
and in-chunk keys; since neither route modifies that kernel we omit
it, though \S\ref{app:kernel-chunk-bwd} differentiates its
state-side contribution.  Stacking $\mathbf{O}_{[c]}$ over all
chunks gives the clean-stream output $\mathbf{o}$ listed below.

\paragraph{Clean-side forward outputs.}
The clean-side forward produces four objects, reused unchanged by
both routes:
\begin{enumerate}\itemsep1pt
\item $\mathbf{o}$, the clean-stream output (the
standard chunkwise attention output);
\item $\{\mathbf{S}_{[c]}\}_{c=0}^{N_C}$, chunk-boundary states
from the scan in Eq.~\eqref{eq:chunk-state};
\item $\{\hat{\mathbf{v}}_t\}_{t=1}^{L}$, chunk-corrected clean
values from the WY step;
\item $\{G_t\}_{t=1}^{L}$, per-position cumulative log-gates.
\end{enumerate}
The route-specific work begins after these outputs: it concerns
only the construction of the block-boundary clean states
$\tilde{\mathbf{S}}_{b,\text{init}}$ that seed the noisy stream,
which \S\ref{app:kernel-seeded-block} takes up next.

\subsubsection{Two-stream recurrence and block-level chain rule}
\label{app:kernel-seeded-block}

The clean side of \S\ref{app:kernel-single-stream} is inherited
unchanged from standard single-stream GDR training;
the \model-specific work supplies the noisy stream with its
block-boundary clean seed.  This subsection defines the seed,
specifies how the noisy recurrence consumes it, and derives the
block-level chain rule that both routes implement.

\paragraph{Clean-transition map over one block.}
Specialising Eq.~\eqref{eq:chunk-state} to a single block of $B$
tokens defines the \emph{clean-transition map}.  For block $b$
spanning positions $(b-1)B+1, \ldots, bB$:
\begin{equation}
\label{eq:f-blk}
  f^{\text{blk}}_b(\mathbf{S})
  \;=\;
  \exp\,\bigl(G_{bB}-G_{(b-1)B}\bigr)\,\mathbf{S}
  \;+\;
  \boldsymbol\Delta_b,
  \qquad
  \boldsymbol\Delta_b
  \;=\!\!\!\sum_{t=(b-1)B+1}^{bB}\!\!\!
    \exp\,\bigl(G_{bB}-G_t\bigr)\,\mathbf{k}_t\,\hat{\mathbf{v}}_t^{\!\top},
\end{equation}
where $f^{\text{blk}}_b : \mathbb{R}^{d_k \times d_v} \to
\mathbb{R}^{d_k \times d_v}$ is the block-$b$ clean-transition map
that advances the clean state by $B$ tokens,
$\exp\,(G_{bB}-G_{(b-1)B})$ is the block's cumulative gate, and
$\boldsymbol\Delta_b \in \mathbb{R}^{d_k \times d_v}$ is the
block's outer-product write in terms of raw clean keys and
chunk-corrected clean values.

\paragraph{Block-boundary clean states.}
Inside chunk $c$, the $M$ block-boundary clean states
$\mathbf{S}_{cM \cdot B + B}, \mathbf{S}_{cM \cdot B + 2B}, \ldots,
\mathbf{S}_{(c+1)C}$ are obtained by composing $f^{\text{blk}}$
starting from the chunk-boundary state $\mathbf{S}_{[c]}$;
applying $f^{\text{blk}}$ all $M$ times recovers
$\mathbf{S}_{[c+1]}$ and Eq.~\eqref{eq:chunk-state} at the next
chunk boundary.  The two routes agree on the definition of these
$L/B$ states and differ only in the stage at which each state is
materialized.

\paragraph{Noisy block forward.}
For block $b$, the noisy recurrence starts from the corresponding
clean block-end state as its initial state, then runs
Eq.~\eqref{eq:per-step-fwd} on the noisy tokens of that block only:
\begin{equation}
\label{eq:noisy-block-fwd}
  \tilde{\mathbf{S}}_{b,\text{init}}
  \;\mathrel{:=}\;
  \mathbf{S}_{(b-1)B},
  \qquad
  \tilde{\mathbf{S}}_\ell
  \;=\;
  \exp\,(\tilde g_\ell)\,\tilde{\mathbf{S}}_{\ell-1}
  \;+\;
  \tilde{\mathbf{k}}_\ell\,\tilde{\boldsymbol\delta}_\ell^{\!\top},
\end{equation}
for $\ell = (b-1)B+1, \ldots, bB$, where
$\tilde{\mathbf{S}}_{b,\text{init}} \in \mathbb{R}^{d_k \times d_v}$
is block $b$'s noisy-recurrence initial state (seeded from the
clean state at the block boundary),
$\tilde{\mathbf{S}}_\ell \in \mathbb{R}^{d_k \times d_v}$ is the
noisy state after noisy position $\ell$, and
$(\tilde g_\ell, \tilde\beta_\ell, \tilde{\mathbf{q}}_\ell,
\tilde{\mathbf{k}}_\ell, \tilde{\mathbf{v}}_\ell)$ are the noisy
per-token inputs (produced by the same linear projections and
ShortConv branches as their clean counterparts, but
applied to the noisy stream).  The noisy per-token delta update
$\tilde{\boldsymbol\delta}_\ell = \tilde\beta_\ell\,\tilde u_\ell
\in \mathbb{R}^{d_v}$ is defined via the noisy prediction error
$\tilde u_\ell = \tilde{\mathbf{v}}_\ell -
(\tilde{\mathbf{S}}^{\text{pre}}_\ell)^{\!\top}\tilde{\mathbf{k}}_\ell$
with $\tilde{\mathbf{S}}^{\text{pre}}_\ell =
\exp\,(\tilde g_\ell)\,\tilde{\mathbf{S}}_{\ell-1}$, reusing the
five-line expansion of Eq.~\eqref{eq:per-step-fwd} on the noisy
stream without restating it.  The noisy output at every position in block
$b$ is read from the block-end state rather than the running state:
$\tilde{\mathbf{o}}_\ell = s\,\tilde{\mathbf{S}}_{bB}^{\!\top}\tilde{\mathbf{q}}_\ell$
for all $\ell \in \{(b{-}1)B{+}1, \ldots, bB\}$.  This single-shared
readout state reproduces the bidirectional noisy-to-noisy region of
the training mask (Eq.~\eqref{eq:flare-gdn-schedule}).  Between blocks
the noisy state is reset; within a block it advances exactly $B$
steps.  Noisy blocks are mutually independent because each is
seeded from its own clean state; all block-to-block information
flow passes through the clean stream.

\paragraph{Block-level chain rule.}
Differentiating Eq.~\eqref{eq:f-blk} yields the chain rule that
both routes reuse.  For a block $b$ inside chunk $c$, let
\[
  \mathbf{dh}_b
  \;\equiv\;
  \partial\mathcal{L}/\partial\mathbf{S}_{bB}
\]
denote the clean state gradient at the end of block $b$, accumulated
so far from later blocks in the same chunk, and let
$\mathbf{d}\tilde{\mathbf{S}}_{b,\text{init}}$ denote the gradient
produced by reverse-sweeping the noisy recurrence of
Eq.~\eqref{eq:noisy-block-fwd} on block $b$ alone.  The block-level
recursion is then
\begin{equation}
\label{eq:dh-block-recursion}
  \mathbf{dh}_{b-1}
  \;=\;
  \mathbf{d}\tilde{\mathbf{S}}_{b,\text{init}}
  \;+\;
  \exp\,\bigl(G_{bB}-G_{(b-1)B}\bigr)\,
  \mathbf{dh}_b.
\end{equation}
The first term is block $b$'s direct contribution to the gradient
at its seed (the clean state at the end of block $b{-}1$).  The
second term is the gate-decayed gradient from blocks
$b{+}1, b{+}2, \ldots$ propagated through $f^{\text{blk}}_b$'s
scalar decay.  Isolating the \emph{within-chunk} part of this scan
by seeding the recursion with $\mathbf{dh}_{cM{+}M} = \mathbf{0}$
(dropping external contributions to chunk $c$) and iterating $M$
times yields $\mathbf{dh}_{cM}$, the \emph{within-chunk
contribution} $\mathbf{dh}^{\text{inject}}_{[c]}$.  The chunk-level
scan of Eq.~\eqref{eq:chunk-bwd-scan} subsequently adds the two
external contributions (from later chunks and from chunk $c$'s
clean-output gradient) on top of this injection.

\subsubsection{Backward primitives}
\label{app:kernel-bwd-primitives}

Three primitives appear repeatedly in both routes: a per-token
backward inside one block (\S\ref{app:kernel-per-step-bwd}); a
chunk-level backward across chunks on the clean side
(\S\ref{app:kernel-chunk-bwd}); and a clean-transition backward
that differentiates $f^{\text{blk}}_b$
(\S\ref{app:kernel-clean-trans-bwd}).  We derive each once and
invoke them by name in Routes~I and~II.

\paragraph{Per-step backward.}
\label{app:kernel-per-step-bwd}

The per-step backward reverses a single token of
Eq.~\eqref{eq:per-step-fwd}; each reverse sweep over a noisy block
consists of $B$ consecutive invocations of this primitive.  Let
$\mathbf{dS}_t = \partial\mathcal{L}/\partial\mathbf{S}_t$,
$\mathbf{do}_t = \partial\mathcal{L}/\partial\mathbf{o}_t$,
$\mathbf{dS}^{\text{pre}}_t = \partial\mathcal{L}/\partial\mathbf{S}^{\text{pre}}_t$,
and $\mathbf{d}u_t = \partial\mathcal{L}/\partial u_t$.  Because
$\mathbf{S}_t$ appears both in $\mathbf{o}_t$ and, through the
subsequent token, in $\mathbf{S}_{t+1}$, its gradient is an
\emph{accumulator}: by the time $\mathbf{dS}_{t-1}$ is computed,
$\mathbf{dS}_t$ has already received the contribution from
$\mathbf{S}_{t+1}$ handed back by the reverse step on token
$t{+}1$.  The equations below therefore use `$\mathrel{+}=$' (and
in Eqs.~\eqref{eq:dalpha}--\eqref{eq:dDelta} also
`$\mathrel{-}=$') in the programming sense: $x\mathrel{+}=y$
denotes $x \leftarrow x + y$.  Differentiating the five lines of
Eq.~\eqref{eq:per-step-fwd} in reverse yields
\begin{equation}
\label{eq:per-step-bwd}
\begin{aligned}
  & \mathbf{dq}_t = s\,\mathbf{S}_t\mathbf{do}_t,
    & \text{(from $\mathbf{o}_t$)}\\
  & \mathbf{dS}_t \mathrel{+}= s\,\mathbf{q}_t\,\mathbf{do}_t^{\!\top},
    & \text{(from $\mathbf{o}_t$; add into $\mathbf{dS}_t$)}\\
  & \mathbf{d}\boldsymbol\delta_t = \mathbf{dS}_t^{\!\top}\mathbf{k}_t,
    \ \ \mathbf{dk}_t^{\text{write}} = \mathbf{dS}_t\,\boldsymbol\delta_t,
    & \text{(from the outer-product write)}\\
  & \mathbf{d}u_t = \beta_t\,\mathbf{d}\boldsymbol\delta_t,
    \ \ \mathbf{dv}_t = \mathbf{d}u_t,
    \ \ d\beta_t = \mathbf{d}\boldsymbol\delta_t^{\!\top}u_t,
    & \text{(from the delta update)}\\
  & \mathbf{dk}_t^{\text{pred}} = -\mathbf{S}^{\text{pre}}_t\,\mathbf{d}u_t,
    \ \ \mathbf{dk}_t = \mathbf{dk}_t^{\text{write}} + \mathbf{dk}_t^{\text{pred}},
    & \text{(from the prediction error)}\\
  & \mathbf{dS}^{\text{pre}}_t = \mathbf{dS}_t - \mathbf{k}_t\,\mathbf{d}u_t^{\!\top},
    & \text{(splits $\mathbf{S}_t = \mathbf{S}^{\text{pre}}_t + \mathbf{k}_t\boldsymbol\delta_t^{\!\top}$)}\\
  & dg_t = \bigl\langle \mathbf{dS}^{\text{pre}}_t,\;\mathbf{S}^{\text{pre}}_t\bigr\rangle_{\mathrm{F}},
    & \text{(from $\mathbf{S}^{\text{pre}}_t = \exp\,(g_t)\,\mathbf{S}_{t-1}$)}\\
  & \mathbf{dS}_{t-1} = \exp\,(g_t)\,\mathbf{dS}^{\text{pre}}_t,
    & \text{(gating reverse recurrence)}
\end{aligned}
\end{equation}
where $\mathbf{d}\boldsymbol\delta_t = \partial\mathcal{L}/\partial\boldsymbol\delta_t \in \mathbb{R}^{d_v}$
is the gradient with respect to the delta update,
$\mathbf{dk}_t^{\text{write}}, \mathbf{dk}_t^{\text{pred}} \in \mathbb{R}^{d_k}$
are the two contributions to $\mathbf{dk}_t$ from the
outer-product write and the prediction error, respectively, and
$\langle\cdot,\cdot\rangle_{\mathrm{F}}$ denotes the Frobenius
inner product.  The key gradient sums the two contributions
because $\mathbf{k}_t$ enters both the outer-product write and the
prediction error.  Denote this primitive by \textsc{PerStepBwd}:
it accepts
$(\mathbf{dS}_t, \mathbf{do}_t, \mathbf{q}_t, \mathbf{k}_t, \mathbf{v}_t, g_t, \beta_t, \mathbf{S}^{\text{pre}}_t, \mathbf{S}_t)$
and returns
$(\mathbf{dq}_t, \mathbf{dk}_t, \mathbf{dv}_t, dg_t, d\beta_t, \mathbf{dS}_{t-1})$.
Eq.~\eqref{eq:per-step-bwd} assumes the token-causal readout
$\mathbf{o}_t = s\,\mathbf{S}_t^{\!\top}\mathbf{q}_t$: each per-step
state feeds exactly one output, so the output-side gradient enters
the state accumulator at the same step.  This is correct for the
clean stream; the noisy stream uses a different output rule, which
we cover next.

\paragraph{Noisy-stream backward under block-end readout.}
The noisy-block forward of Eq.~\eqref{eq:noisy-block-fwd} reads
\emph{every} noisy output in block $b$ from the same end-of-block
state $\tilde{\mathbf{S}}_{bB}$, so the mapping from outputs to
states is many-to-one instead of one-to-one.  Differentiating this
readout gives a block-level output-side contribution that lives
entirely on $\tilde{\mathbf{S}}_{bB}$'s gradient, plus a per-step
query gradient that uses the same single state:
\begin{equation}
\label{eq:noisy-readout-bwd}
  \mathbf{d}\tilde{\mathbf{S}}_{bB}
  \;=\;
  s\!\!\sum_{\ell=(b-1)B+1}^{bB}\!\!\tilde{\mathbf{q}}_\ell\,\mathbf{d}\tilde{\mathbf{o}}_\ell^{\!\top},
  \qquad
  \mathbf{d}\tilde{\mathbf{q}}_\ell
  \;=\;
  s\,\tilde{\mathbf{S}}_{bB}\,\mathbf{d}\tilde{\mathbf{o}}_\ell
  \quad\text{for all } \ell\in\{(b-1)B+1,\ldots,bB\}.
\end{equation}
A reverse sweep over the noisy block then runs $B$ consecutive
\textsc{PerStepBwd} calls with $\mathbf{d}\tilde{\mathbf{o}}_\ell$ set to
$\mathbf{0}$: the per-step output terms (the $\mathbf{dq}_t$ and
$\mathbf{dS}_t \mathrel{+}{=} s\,\mathbf{q}_t\,\mathbf{do}_t^{\!\top}$
lines of Eq.~\eqref{eq:per-step-bwd}) are already supplied by
Eq.~\eqref{eq:noisy-readout-bwd}, and the remaining lines (gating,
outer-product write, prediction error, and gating reverse
recurrence) are unchanged.  Throughout this appendix, ``reverse
sweep over a noisy block of $B$ tokens'' refers to this procedure:
initialize $\mathbf{d}\tilde{\mathbf{S}}_{bB}$ with
Eq.~\eqref{eq:noisy-readout-bwd}, emit $\mathbf{d}\tilde{\mathbf{q}}_\ell$
directly from Eq.~\eqref{eq:noisy-readout-bwd}, then invoke
\textsc{PerStepBwd} with $\mathbf{do} = \mathbf{0}$ for each of the
$B$ steps in reverse.

\paragraph{Chunk-level clean backward.}
\label{app:kernel-chunk-bwd}

The chunk-level backward differentiates Eq.~\eqref{eq:chunk-state}
together with the clean-output kernel that emits
$\mathbf{O}_{[c]}$.  Its only cross-chunk serial step is a reverse
matmul scan over chunk-boundary state gradients
$\mathbf{dh}_{[c]} \equiv \partial\mathcal{L}/\partial\mathbf{S}_{[c]}$:
\begin{equation}
\label{eq:chunk-bwd-scan}
  \mathbf{dh}_{[c]}
  \;=\;
  \underbrace{\mathbf{dh}^{\text{inject}}_{[c]}}_{\text{external hand-off}}
  \;+\;
  \underbrace{\mathcal{C}_{[c]}}_{\text{in-chunk clean-output contribution}}
  \;+\;
  \exp\,\bigl(G_{(c+1)C}-G_{cC}\bigr)\,\mathbf{dh}_{[c+1]},
\end{equation}
where $\mathbf{dh}^{\text{inject}}_{[c]} \in \mathbb{R}^{d_k \times d_v}$
is a per-chunk hand-off slot that accepts external
state-gradients and adds them into the scan (written by Route~I's
fill-in backward, \S\ref{app:route-refine}, or by Route~II's fused
kernel, \S\ref{app:route-twostream}), and
$\mathcal{C}_{[c]} \in \mathbb{R}^{d_k \times d_v}$ is the
in-chunk clean-output contribution obtained by differentiating
$\mathbf{O}_{[c]}$ through $\mathbf{S}_{[c]}$ via a dense matmul
over the $C$ tokens of chunk $c$.  $\mathcal{C}_{[c]}$ is
identical to the AR baseline's clean-side computation and is not
modified by either route; the two-stream-specific content is
routed entirely through $\mathbf{dh}^{\text{inject}}_{[c]}$.
Denote this primitive by \textsc{ChunkBwd}.

\paragraph{Clean-transition backward.}
\label{app:kernel-clean-trans-bwd}

The clean-transition backward links the noisy-side reverse sweep
to the clean-side cross-chunk scan: it converts the state
gradient $\mathbf{dh}_b$ at the end of block $b$ into per-token
clean gradients and a propagated block-initial state gradient.
It is the block-level analog of \S\ref{app:kernel-chunk-bwd}.
We differentiate $f^{\text{blk}}_b$ (Eq.~\eqref{eq:f-blk}) with
respect to its clean-side inputs
$\bigl(\mathbf{k}_t, \hat{\mathbf{v}}_t, G_t\bigr)_{t\in\text{block}(b)}$
and the incoming state $\mathbf{S}$.  Let
$\rho_t \equiv \exp\,(G_{bB}-G_t)$ denote the per-token decay
factor inside block $b$, and let
$\mathbf{dh}_b \equiv \partial\mathcal{L}/\partial\mathbf{S}_{bB}$
denote the incoming state gradient.

\paragraph{Gradient through the scalar decay.}
$f^{\text{blk}}_b$'s leading term $\exp\,(G_{bB}-G_{(b-1)B})\,\mathbf{S}$
propagates $\mathbf{dh}_b$ to $\mathbf{S}$ with factor
$\exp\,(G_{bB}-G_{(b-1)B})$; this factor is exactly the multiplicative
factor in the block-level recursion of Eq.~\eqref{eq:dh-block-recursion},
and is how the recursion is derived.  The same decay also yields
gradients into $G_{bB}$ and $G_{(b-1)B}$:
\begin{equation}
\label{eq:dalpha}
\begin{aligned}
  & dG_{bB} \mathrel{+}= \exp\,\bigl(G_{bB}-G_{(b-1)B}\bigr)\,\bigl\langle \mathbf{dh}_b,\;\mathbf{S}\bigr\rangle_{\mathrm{F}},
    & \text{(contribution to $dG_{bB}$)}\\
  & dG_{(b-1)B} \mathrel{-}= \exp\,\bigl(G_{bB}-G_{(b-1)B}\bigr)\,\bigl\langle \mathbf{dh}_b,\;\mathbf{S}\bigr\rangle_{\mathrm{F}}.
    & \text{(contribution to $dG_{(b-1)B}$)}
\end{aligned}
\end{equation}
The two writes have equal magnitude and opposite signs because
$\partial(G_j - G_i)/\partial G_j = 1$ and
$\partial(G_j - G_i)/\partial G_i = -1$.  Since each $dG$
accumulator receives contributions from several sources (this
block's scalar-decay term of Eq.~\eqref{eq:dalpha}, this block's $\boldsymbol\Delta$ term
below, and any later calls that hit the same $G$), we keep the
`$\mathrel{+}=$' / `$\mathrel{-}=$' form throughout
\S\ref{app:kernel-clean-trans-bwd} to make the accumulator
semantics explicit.

\paragraph{Gradient through the outer-product write.}
The write term $\boldsymbol\Delta_b = \sum_t \rho_t\,\mathbf{k}_t\,\hat{\mathbf{v}}_t^{\!\top}$
contributes, for each $t$ in block $b$:
\begin{equation}
\label{eq:dDelta}
\begin{aligned}
  & \mathbf{dk}_t = \mathbf{dh}_b\bigl(\hat{\mathbf{v}}_t\,\rho_t\bigr),
    & \text{(clean key)}\\
  & \mathbf{d}\hat{\mathbf{v}}_t = \rho_t\,\mathbf{dh}_b^{\!\top}\mathbf{k}_t,
    & \text{(chunk-corrected value)}\\
  & \sigma_t := \rho_t\,\mathbf{k}_t^{\!\top}\mathbf{dh}_b\,\hat{\mathbf{v}}_t,
    \ \ dG_{bB}\mathrel{+}= \sigma_t,
    \ \ dG_t\mathrel{-}= \sigma_t,
    & \text{(cumulative gates)}
\end{aligned}
\end{equation}
where $\mathbf{dk}_t \in \mathbb{R}^{d_k}$ and
$\mathbf{d}\hat{\mathbf{v}}_t \in \mathbb{R}^{d_v}$ are the
per-token gradients of the block's outer-product write term with
respect to the clean key and chunk-corrected clean value, and
$\sigma_t \in \mathbb{R}$ is the scalar obtained by contracting
$\mathbf{dh}_b$ with the rank-one update
$\mathbf{k}_t\hat{\mathbf{v}}_t^{\!\top}$ and multiplying by
$\rho_t$; it is the common magnitude of the two $dG$ writes.  The
two $G$-writes have opposite signs because $G_{bB}$ and $G_t$
enter $\rho_t$ with opposite signs.  Summed over
$t \in \text{block}(b)$ (and over all blocks and all contribution
sources), these give the final per-token clean-side gradients
$(\mathbf{dk}_t, \mathbf{d}\hat{\mathbf{v}}_t, dG_t)$ of block $b$.

\paragraph{Summary.}
Denote the combined sub-procedure by \textsc{CleanTransBwd}: it
accepts
$(\mathbf{dh}_b, \mathbf{S}, \{(\mathbf{k}_t, \hat{\mathbf{v}}_t, G_t)\}_{t\in\text{block}(b)})$
and returns $(\mathbf{dk}_t, \mathbf{d}\hat{\mathbf{v}}_t, dG_t)$
for every token in block $b$.  Its outputs are partial-gradient
slots: the downstream chunk-level clean backward (\textsc{ChunkBwd},
together with the WY backward that converts
$\mathbf{d}\hat{\mathbf{v}}$ into $(\mathbf{dv}, \mathbf{dk}, d\beta)$)
completes backpropagation into the final clean parameter
gradients.

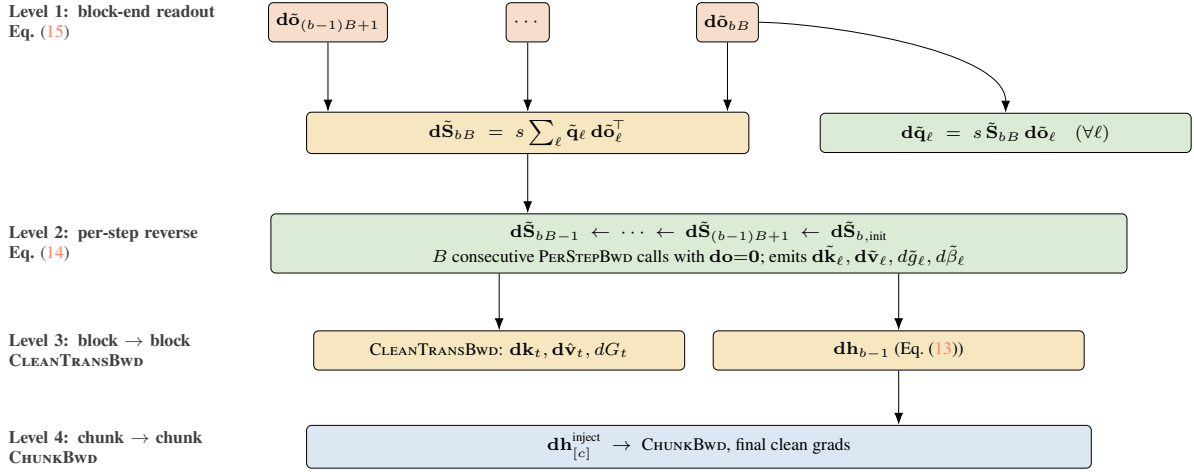
\begin{figure}[h]
\centering
\resizebox{0.95\linewidth}{!}{%
\begin{tikzpicture}[font=\scriptsize, >=Latex]
  \tikzstyle{noded}=[draw, rounded corners=2pt, align=center, inner sep=3pt,
                     minimum height=0.55cm, font=\scriptsize]
  \definecolor{outCol}{HTML}{F7DDCB}
  \definecolor{gradCol}{HTML}{DCEBD6}
  \definecolor{blockCol}{HTML}{F6E7C1}
  \definecolor{xcnkCol}{HTML}{DCE7F3}

  \def\labelx{-8.0}
  \tikzstyle{stageL}=[font=\scriptsize\bfseries, anchor=west, text=black!75,
                      align=left, text width=2.9cm]

  \node[noded, fill=outCol] (do1) at (-3.4, 0) {$\mathbf{d}\tilde{\mathbf{o}}_{(b-1)B+1}$};
  \node[noded, fill=outCol] (do2) at (-0.6, 0) {$\cdots$};
  \node[noded, fill=outCol] (doB) at ( 2.2, 0) {$\mathbf{d}\tilde{\mathbf{o}}_{bB}$};
  \node[stageL] at (\labelx, 0)
       {Level 1: block-end readout\\Eq.~\eqref{eq:noisy-readout-bwd}};
  \coordinate (labcol) at (\labelx, 0);

  \node[noded, fill=blockCol, minimum width=6.2cm] (dSend)
       at (-0.6, -1.55)
       {$\mathbf{d}\tilde{\mathbf{S}}_{bB} \;=\; s\sum_\ell \tilde{\mathbf{q}}_\ell\,\mathbf{d}\tilde{\mathbf{o}}_\ell^{\!\top}$};
  \node[noded, fill=gradCol, minimum width=5.2cm] (dq)
       at ( 6.1, -1.55)
       {$\mathbf{d}\tilde{\mathbf{q}}_\ell \;=\; s\,\tilde{\mathbf{S}}_{bB}\,\mathbf{d}\tilde{\mathbf{o}}_\ell\quad(\forall\ell)$};

  \draw[->] (do1.south) -- (dSend.north -| do1.south);
  \draw[->] (do2.south) -- (dSend.north);
  \draw[->] (doB.south) -- (dSend.north -| doB.south);
  \draw[->] (doB.east) .. controls +(0.8,0) and +(0,0.9) .. (dq.north);

  \node[noded, fill=gradCol, minimum width=12.0cm] (chain)
       at ( 1.8, -3.1)
       {$\mathbf{d}\tilde{\mathbf{S}}_{bB-1}\ \leftarrow\ \cdots\ \leftarrow\ \mathbf{d}\tilde{\mathbf{S}}_{(b-1)B+1}\ \leftarrow\ \mathbf{d}\tilde{\mathbf{S}}_{b,\text{init}}$\\
        $B$ consecutive \textsc{PerStepBwd} calls with $\mathbf{do}{=}\mathbf{0}$; emits $\mathbf{d}\tilde{\mathbf{k}}_\ell,\mathbf{d}\tilde{\mathbf{v}}_\ell,d\tilde g_\ell,d\tilde\beta_\ell$};
  \node[stageL] at (labcol |- chain)
       {Level 2: per-step reverse\\Eq.~\eqref{eq:per-step-bwd}};
  \draw[->] (dSend.south) -- (dSend.south |- chain.north);

  \node[noded, fill=blockCol, minimum width=5.2cm] (dcT)
       at (-1.0, -4.6)
       {\textsc{CleanTransBwd}: $\mathbf{dk}_t,\mathbf{d}\hat{\mathbf{v}}_t,dG_t$};
  \node[noded, fill=blockCol, minimum width=5.2cm] (dhnext)
       at ( 4.6, -4.6)
       {$\mathbf{dh}_{b-1}$ (Eq.~\eqref{eq:dh-block-recursion})};
  \node[stageL] at (labcol |- dcT)
       {Level 3: block $\to$ block\\\textsc{CleanTransBwd}};
  \draw[->] (dcT.north |- chain.south) -- (dcT.north);
  \draw[->] (dhnext.north |- chain.south) -- (dhnext.north);

  \node[noded, fill=xcnkCol, minimum width=11.0cm] (dhinject)
       at ( 1.8, -5.95)
       {$\mathbf{dh}^{\text{inject}}_{[c]}\ \to\ \textsc{ChunkBwd}$, final clean grads};
  \node[stageL] at (labcol |- dhinject)
       {Level 4: chunk $\to$ chunk\\\textsc{ChunkBwd}};
  \draw[->] (dhnext.south) -- (dhnext.south |- dhinject.north);
\end{tikzpicture}%
}
\caption{Backward dataflow for one noisy block $b$ of chunk $c$
under the block-end readout of
Eq.~\eqref{eq:flare-gdn-schedule}.
\textbf{Level~1} (output $\to$ state) folds the $B$ per-token
output gradients into the single block-end state accumulator
$\mathbf{d}\tilde{\mathbf{S}}_{bB}$ and produces $\mathbf{d}\tilde{\mathbf{q}}_\ell$
for every $\ell$ directly
(Eq.~\eqref{eq:noisy-readout-bwd}).
\textbf{Level~2} (per-step reverse) runs $B$ \textsc{PerStepBwd} calls
with $\mathbf{do}{=}\mathbf{0}$, propagating the block-end gradient
to the block-initial state $\mathbf{d}\tilde{\mathbf{S}}_{b,\text{init}}$
while emitting the remaining per-token gradients.
\textbf{Level~3} (block $\to$ block) runs \textsc{CleanTransBwd} on
$\mathbf{dh}_b$ and combines $\mathbf{d}\tilde{\mathbf{S}}_{b,\text{init}}$
with the gate-decayed $\mathbf{dh}_b$ (Eq.~\eqref{eq:dh-block-recursion})
to produce $\mathbf{dh}_{b-1}$.
\textbf{Level~4} (chunk $\to$ chunk) closes the block recursion into
$\mathbf{dh}^{\text{inject}}_{[c]}$, which is consumed by
\textsc{ChunkBwd}.  Both routes produce this dataflow; they differ
only in which kernel hosts which level.}
\label{fig:kernel-grad-flow}
\end{figure}

\paragraph{Gradient map.}
Figure~\ref{fig:kernel-grad-flow} traces the composition of the
three primitives on a single noisy token.  Both routes produce
the same tree and differ only in which kernel computes which
subtree.

\subsubsection{Overview of the two implementation routes}
\label{app:route-overview}

\begin{figure}[h]
\centering
\definecolor{cleanBox}{HTML}{DCE7F3}
\definecolor{stateBox}{HTML}{F6E7C1}
\definecolor{noisyBox}{HTML}{F7DDCB}
\definecolor{fusedBox}{HTML}{DCEBD6}

\begin{subfigure}[b]{\linewidth}\centering
\begin{tikzpicture}[font=\scriptsize, node distance=0.35cm and 0.3cm, >=Latex]
  \tikzstyle{stage}=[draw, rounded corners=2pt, align=center,
                     minimum height=1.0cm, minimum width=3.1cm, inner sep=2.5pt]
  \node[stage, fill=cleanBox] (wy1)
    {Clean chunkwise fwd\\\S\ref{app:kernel-single-stream}\\$\to\mathbf{o},\{\mathbf{S}_{[c]}\},\hat{\mathbf{v}},G$};
  \node[stage, fill=stateBox, right=of wy1] (refine1)
    {Fill-in fwd\\$M-1$ applications of $f^{\text{blk}}$ per chunk\\writes $\{\mathbf{S}_{(b-1)B}\}_{b=1}^{L/B}$ to HBM};
  \node[stage, fill=noisyBox, right=of refine1] (noisy1)
    {Block-local noisy fwd\\$L/B$ independent blocks\\$\to\tilde{\mathbf{o}}$};
  \node[left=0.1cm of wy1, font=\footnotesize\bfseries, text=black!70, align=right]
       {Route~I};
  \draw[->,thick] (wy1) -- (refine1);
  \draw[->,thick] (refine1) -- (noisy1);

  \node[stage, fill=cleanBox, below=0.7 of wy1] (wy2)
    {Clean chunkwise fwd\\same outputs as Route~I\\plus strided clean-state ckpts};
  \node[stage, fill=fusedBox, right=of wy2, minimum width=6.5cm] (fused2)
    {Two-stream fwd (fused, Alg.~\ref{alg:route-twostream-fwd})\\one program per chunk; replays block-boundary clean states in registers,\\runs noisy blocks in place $\to\tilde{\mathbf{o}}$};
  \node[left=0.1cm of wy2, font=\footnotesize\bfseries, text=black!70, align=right]
       {Route~II};
  \draw[->,thick] (wy2) -- (fused2);
\end{tikzpicture}
\subcaption{Forward dataflow.  Route~I materializes every
$\mathbf{S}_{(b-1)B}$ in HBM before the noisy blocks run;
Route~II replays each in registers from $\mathbf{S}_{[c]}$, writing only strided checkpoints for the backward.}
\label{fig:route-fwd}
\end{subfigure}

\vspace{10pt}

\begin{subfigure}[b]{\linewidth}\centering
\begin{tikzpicture}[font=\scriptsize, node distance=0.35cm and 0.3cm, >=Latex]
  \tikzstyle{stage}=[draw, rounded corners=2pt, align=center,
                     minimum height=1.1cm, minimum width=3.1cm, inner sep=2.5pt]
  \node[stage, fill=noisyBox] (nbwd)
    {Block-local noisy bwd\\Alg.~\ref{alg:route-i-noisybwd}\\
     $\to$ noisy grads $+$ $\{\mathbf{d}\tilde{\mathbf{S}}_{b,\text{init}}\}$ to HBM};
  \node[stage, fill=stateBox, right=of nbwd] (rbwd)
    {Fill-in bwd\\Alg.~\ref{alg:route-i-fillinbwd}\\
     $\to$ in-chunk clean grads $+$ $\mathbf{dh}^{\text{inject}}_{[c]}$};
  \node[stage, fill=cleanBox, right=of rbwd] (cbwd)
    {\textsc{ChunkBwd} (\S\ref{app:kernel-chunk-bwd})\\consumes $\mathbf{dh}^{\text{inject}}_{[c]}$\\$\to$ final clean grads};
  \node[left=0.1cm of nbwd, font=\footnotesize\bfseries, text=black!70, align=right]
       {Route~I};
  \draw[->,thick] (nbwd) -- (rbwd);
  \draw[->,thick] (rbwd) -- (cbwd);

  \node[stage, fill=fusedBox, minimum width=6.5cm, anchor=north west] (fbwd)
    at ($(nbwd.south west)+(0,-0.7cm)$)
    {Two-stream bwd (fused; Alg.~\ref{alg:route-twostream-bwd})\\one program per chunk; emits all noisy $+$ in-chunk clean grads\\plus $\mathbf{dh}^{\text{inject}}_{[c]}$ to HBM};
  \node[stage, fill=cleanBox, right=of fbwd] (cbwd2)
    {\textsc{ChunkBwd} (\S\ref{app:kernel-chunk-bwd})\\consumes $\mathbf{dh}^{\text{inject}}_{[c]}$\\$\to$ final clean grads};
  \node[left=0.1cm of fbwd, font=\footnotesize\bfseries, text=black!70, align=right]
       {Route~II};
  \draw[->,thick] (fbwd) -- (cbwd2);
\end{tikzpicture}
\subcaption{Backward dataflow.  Route~I chains three kernels
through two HBM hand-off tensors; Route~II collapses the first
two into a single fused kernel and emits only
$\mathbf{dh}^{\text{inject}}_{[c]}$ before \textsc{ChunkBwd}.}
\label{fig:route-bwd}
\end{subfigure}

\caption{Dataflow of the two implementation routes (forward, a;
backward, b).  Each box is a single Triton kernel launch.  Both
routes leave the clean-side kernels of
\S\ref{app:kernel-single-stream} and \S\ref{app:kernel-chunk-bwd}
untouched.}
\label{fig:route-overview}
\end{figure}
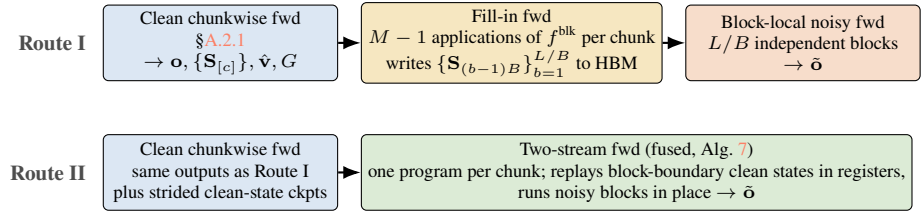
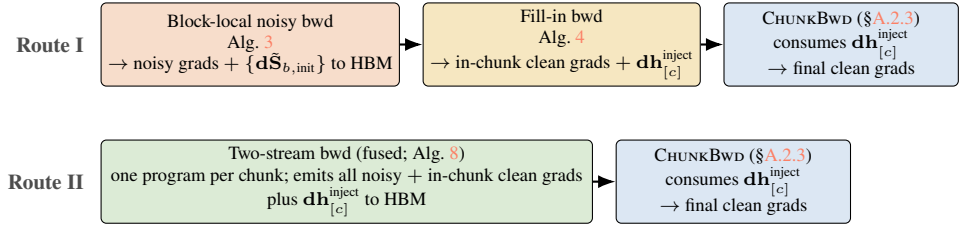

Both routes compute identical gradients and differ only in (i) how
the block-boundary clean states $\mathbf{S}_{(b-1)B}$ are delivered
on the forward and (ii) how block-level gradients are routed into
the cross-chunk scan.  Figure~\ref{fig:route-overview} contrasts
their dataflow; \S\ref{app:route-refine} and
\S\ref{app:route-twostream} specify the algorithms and
\S\ref{app:route-comparison} quantifies the trade-off.

The block-boundary clean states $\mathbf{S}_{(b-1)B}$ follow
\S\ref{app:kernel-seeded-block}; both routes ultimately route their
gradients into the chunk-boundary slot consumed by
\textsc{ChunkBwd}.

\subsubsection{Route~I: Chunk-then-Refine}
\label{app:route-refine}

Route~I materializes every $\mathbf{S}_{(b-1)B}$ in HBM prior to
the noisy forward.  The noisy stream then reduces to $L/B$
independent block-local recurrences that run in parallel.

\begin{enumerate}
\item \textbf{Clean chunkwise forward.}  Run
\S\ref{app:kernel-single-stream} on the clean stream; emit
$\mathbf{o}$, $\{\mathbf{S}_{[c]}\}$, $\{\hat{\mathbf{v}}_t\}$,
and $\{G_t\}$.
\item \textbf{Block-state fill-in (forward).}  For each chunk $c$,
apply $f^{\text{blk}}_{cM+1}, \ldots, f^{\text{blk}}_{cM+M-1}$
(Eq.~\eqref{eq:f-blk}) starting from $\mathbf{S}_{[c]}$, and write
every $\mathbf{S}_{(b-1)B}$ for $b$ inside chunk $c$ to HBM.  The
pass has no tensor-core work; it is a per-block recurrent scan whose
only purpose is to populate the tensor
$\{\mathbf{S}_{(b-1)B}\}_{b=1}^{L/B}$.
\item \textbf{Block-local noisy forward.}  Treat the noisy stream as
$L/B$ independent length-$B$ sequences; each starts from its
fill-in state $\mathbf{S}_{(b-1)B}$, runs Eq.~\eqref{eq:noisy-block-fwd}
for $B$ steps, and emits $\tilde{\mathbf{o}}$ for those
positions.
\end{enumerate}

\paragraph{Backward.}
The backward mirrors the forward as three kernels connected by
autograd through two HBM hand-off tensors
(Figure~\ref{fig:route-bwd}, Route~I row).  Using the primitives
of \S\ref{app:kernel-bwd-primitives}:

\begin{algorithm}[h]
\caption{Route~I noisy backward (one program per block $b$).}
\label{alg:route-i-noisybwd}
\small
\begin{algorithmic}[1]
\Require block $b$'s noisy tokens $(\tilde{\mathbf{q}}, \tilde{\mathbf{k}}, \tilde{\mathbf{v}}, \tilde g, \tilde\beta)$, seed $\mathbf{S}_{(b-1)B}$, noisy-output grads $\{\mathbf{d}\tilde{\mathbf{o}}_\ell\}_{\ell\in\text{block}(b)}$
\State run Eq.~\eqref{eq:noisy-block-fwd} forward for $B$ steps, caching $(\tilde{\mathbf{S}}^{\text{pre}}_\ell, \tilde{\mathbf{S}}_\ell)$ per noisy token
\State $\mathbf{d}\tilde{\mathbf{S}}_{bB} \gets s\sum_{\ell=(b-1)B+1}^{bB} \tilde{\mathbf{q}}_\ell\,\mathbf{d}\tilde{\mathbf{o}}_\ell^{\!\top}$ \Comment{block-end readout init, Eq.~\eqref{eq:noisy-readout-bwd}}
\State $\mathbf{d}\tilde{\mathbf{q}}_\ell \gets s\,\tilde{\mathbf{S}}_{bB}\,\mathbf{d}\tilde{\mathbf{o}}_\ell$ for each $\ell$ in block $b$
\For{$\ell = bB, bB-1, \ldots, (b-1)B+1$}
  \State $(\_, \mathbf{d}\tilde{\mathbf{k}}_\ell, \mathbf{d}\tilde{\mathbf{v}}_\ell, d\tilde g_\ell, d\tilde\beta_\ell, \mathbf{d}\tilde{\mathbf{S}}_{\ell-1}) \gets \textsc{PerStepBwd}(\mathbf{d}\tilde{\mathbf{S}}_\ell, \mathbf{0}, \ldots)$ \Comment{pass $\mathbf{do}{=}\mathbf{0}$}
\EndFor
\State \textbf{write} $\mathbf{d}\tilde{\mathbf{S}}_{b,\text{init}} \gets \mathbf{d}\tilde{\mathbf{S}}_{(b-1)B}$ to HBM
\State \Return per-noisy-token gradients
\end{algorithmic}
\end{algorithm}

\begin{algorithm}[h]
\caption{Route~I fill-in backward (one program per chunk $c$).}
\label{alg:route-i-fillinbwd}
\small
\begin{algorithmic}[1]
\Require chunk $c$'s clean tokens, HBM-resident
$\{\mathbf{S}_{(b-1)B}\}_{b\in\text{chunk}(c)}$ and
$\{\mathbf{d}\tilde{\mathbf{S}}_{b,\text{init}}\}_{b\in\text{chunk}(c)}$
\State $\mathbf{dh}\gets\mathbf{0}$
  \Comment{accumulator for Eq.~\eqref{eq:dh-block-recursion}}
\For{$j = M-1, \ldots, 0$}
  \State $b\gets cM + j + 1$;\ \ \ $\mathbf{S}\gets\mathbf{S}_{(b-1)B}$
  \State \textsc{CleanTransBwd}$\bigl(\mathbf{dh}, \mathbf{S}, \{(\mathbf{k}_t, \hat{\mathbf{v}}_t, G_t)\}_{t\in\text{block}(b)}\bigr)$
       \Comment{writes $(\mathbf{dk}_t, \mathbf{d}\hat{\mathbf{v}}_t, dG_t)$}
  \State $\mathbf{dh}\gets \mathbf{d}\tilde{\mathbf{S}}_{b,\text{init}} + \exp\,\bigl(G_{bB}-G_{(b-1)B}\bigr)\,\mathbf{dh}$
       \Comment{Eq.~\eqref{eq:dh-block-recursion}}
\EndFor
\State \textbf{write} $\mathbf{dh}^{\text{inject}}_{[c]}\gets \mathbf{dh}$ to HBM
\end{algorithmic}
\end{algorithm}

\begin{algorithm}[h]
\caption{\textsc{Chunk-then-Refine} (pipeline).}
\label{alg:route-refine}
\small
\begin{algorithmic}[1]
\State \textbf{fwd:} $(\mathbf{o}, \{\mathbf{S}_{[c]}\}, \hat{\mathbf{v}}, G) \gets$ Clean-fwd (\S\ref{app:kernel-single-stream})
\State \textbf{fwd:} $\{\mathbf{S}_{(b-1)B}\} \gets$ Fill-in-fwd by repeated application of $f^{\text{blk}}$ (Eq.~\eqref{eq:f-blk})
       \Comment{$L/B$ states to HBM}
\State \textbf{fwd:} \textbf{for each} block $b$ \textbf{in parallel}: run Eq.~\eqref{eq:noisy-block-fwd} for $B$ steps from $\mathbf{S}_{(b-1)B}$
\State \textbf{bwd:} noisy-bwd (Alg.~\ref{alg:route-i-noisybwd}, for all $b$ in parallel)
\State \textbf{bwd:} fill-in-bwd (Alg.~\ref{alg:route-i-fillinbwd}, for all $c$ in parallel)
\State \textbf{bwd:} final clean grads $\gets$ \textsc{ChunkBwd}($\mathbf{do}$, $\mathbf{dh}^{\text{inject}}_{[c]}$)
\end{algorithmic}
\end{algorithm}

Route~I's two costs are an
$L/B$-sized tensor of block-boundary clean states in HBM and a
three-kernel serial backward chain; Route~II
(\S\ref{app:route-twostream}) eliminates both, and
\S\ref{app:route-comparison} quantifies the difference.

\subsubsection{Route~II: Fused Two-Stream}
\label{app:route-twostream}

Route~II collapses forward stages~2--3 (and the corresponding
backward stages) into a single fused kernel per direction.  The
forward kernel launches one program per chunk, loads
$\mathbf{S}_{[c]}$ into registers, and interleaves noisy-block
forward passes with applications of $f^{\text{blk}}$ that advance
the in-register clean state.  No block-boundary clean state is
written to HBM.  The backward launches one program per
$(c,\, i_V)$ pair, where $i_V \in \{1, \ldots, \lceil d_v/b_v \rceil\}$
indexes the value-dimension tile, and sweeps the same structure in reverse, invoking the
primitives of \S\ref{app:kernel-bwd-primitives} as named
sub-procedures.

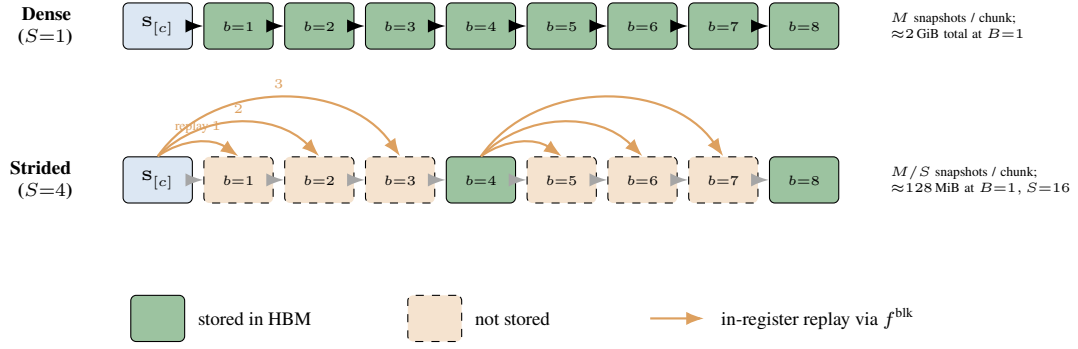
\begin{figure}[h]
\centering
\resizebox{0.95\linewidth}{!}{%
\begin{tikzpicture}[font=\scriptsize, >=Latex]
  \definecolor{storedCol}{HTML}{9CC3A0}
  \definecolor{recompCol}{HTML}{DDA05F}
  \definecolor{chunkCol}{HTML}{DCE7F3}
  \tikzstyle{blkstored}=[draw, rounded corners=2pt, minimum height=0.6cm,
                         minimum width=0.9cm, fill=storedCol, align=center,
                         inner sep=1pt, font=\tiny]
  \tikzstyle{blkrecomp}=[draw, rounded corners=2pt, minimum height=0.6cm,
                         minimum width=0.9cm, dashed, fill=recompCol!30,
                         align=center, inner sep=1pt, font=\tiny]
  \tikzstyle{blkchunk}=[draw, rounded corners=2pt, minimum height=0.6cm,
                        minimum width=0.9cm, fill=chunkCol, align=center,
                        inner sep=1pt, font=\tiny]

  \node[anchor=east, align=right, text width=2.0cm, font=\scriptsize\bfseries]
       at (-1.0, 0.0) {Dense\\($S{=}1$)};
  \node[blkchunk] (a0) at (0.0, 0) {$\mathbf{S}_{[c]}$};
  \foreach \i in {1,...,8} {
    \pgfmathsetmacro{\xpos}{\i * 1.05}
    \node[blkstored] (a\i) at (\xpos, 0) {$b{=}\i$};
  }
  \foreach \i [evaluate=\i as \prev using int(\i-1)] in {1,...,8}
    \draw[->, thick] (a\prev.east) -- (a\i.west);
  \node[anchor=west, align=left, text width=2.6cm, font=\tiny]
        at (9.4, 0) {$M$ snapshots / chunk;\\ ${\approx}2$\,GiB total at $B{=}1$};

  \node[anchor=east, align=right, text width=2.0cm, font=\scriptsize\bfseries]
       at (-1.0, -2.0) {Strided\\($S{=}4$)};
  \node[blkchunk] (b0) at (0.0, -2.0) {$\mathbf{S}_{[c]}$};
  \foreach \i in {1,2,3,5,6,7} {
    \pgfmathsetmacro{\xpos}{\i * 1.05}
    \node[blkrecomp] (b\i) at (\xpos, -2.0) {$b{=}\i$};
  }
  \foreach \i in {4,8} {
    \pgfmathsetmacro{\xpos}{\i * 1.05}
    \node[blkstored] (b\i) at (\xpos, -2.0) {$b{=}\i$};
  }
  \foreach \i [evaluate=\i as \prev using int(\i-1)] in {1,...,8}
    \draw[->, thick, gray!70] (b\prev.east) -- (b\i.west);
  \draw[->, thick, recompCol, bend left=35]
       (b0.north) to node[midway, above, font=\tiny, text=recompCol]
       {replay $1$} (b1.north);
  \draw[->, thick, recompCol, bend left=45]
       (b0.north) to node[midway, above, font=\tiny, text=recompCol] {$2$} (b2.north);
  \draw[->, thick, recompCol, bend left=55]
       (b0.north) to node[midway, above, font=\tiny, text=recompCol] {$3$} (b3.north);
  \draw[->, thick, recompCol, bend left=35]
       (b4.north) to (b5.north);
  \draw[->, thick, recompCol, bend left=45]
       (b4.north) to (b6.north);
  \draw[->, thick, recompCol, bend left=55]
       (b4.north) to (b7.north);
  \node[anchor=west, align=left, text width=2.6cm, font=\tiny]
        at (9.4, -2.0) {$M/S$ snapshots / chunk;\\ ${\approx}128$\,MiB at $B{=}1, S{=}16$};

  \node[blkstored, minimum width=0.7cm] at (0.0, -3.8) {};
  \node[anchor=west, font=\scriptsize] at (0.4, -3.8) {stored in HBM};
  \node[blkrecomp, minimum width=0.7cm] at (3.6, -3.8) {};
  \node[anchor=west, font=\scriptsize] at (4.0, -3.8) {not stored};
  \draw[->, thick, recompCol] (6.4, -3.8) -- (7.1, -3.8);
  \node[anchor=west, font=\scriptsize] at (7.2, -3.8)
       {in-register replay via $f^{\text{blk}}$};
\end{tikzpicture}%
}
\caption{Dense ($S{=}1$, top) versus strided ($S{=}4$, bottom)
checkpointing of the block-boundary clean states inside one chunk
($M{=}8$).  Solid boxes are stored in HBM; dashed boxes are
reconstructed in registers at backward time by applying $\le S{-}1$
copies of $f^{\text{blk}}$ from the nearest stored snapshot.
Route~II's HBM footprint scales as $L/(BS)$ versus Route~I's
$L/B$.}
\label{fig:ckpt-strided}
\end{figure}

\paragraph{Strided clean-state checkpoints.}
The backward requires $\mathbf{S}_{(b-1)B}$ for every block $b$ in
the chunk, but the forward keeps only $\mathbf{S}_{[c]}$ resident.
A direct remedy is to store every block-end clean state to HBM
($M$ snapshots per chunk, $L/B$ snapshots per layer per sample),
which reproduces Route~I's fill-in tensor and its $L/B$ footprint.
We instead adopt a \emph{strided-checkpoint} tensor parameterised
by the \emph{checkpoint stride}
$S \in \{1,2,\ldots,M\}$: within each chunk only every $S$-th
block-boundary clean state is stored, giving
$N_{\text{ckpt}} = M/S$ snapshots per chunk and
$N_C\cdot M/S = L/(BS)$ snapshots per layer per sample.  The
remaining $(S{-}1)/S$ fraction of block-end states are not
persisted; the backward reconstructs each by loading the nearest
prior snapshot and applying at most $S{-}1$ copies of
$f^{\text{blk}}$ in registers (Fig.~\ref{fig:ckpt-strided}).
$S$ parameterises a time/memory trade-off: the HBM cost scales as
$1/S$ and the per-block replay work scales as $S{-}1$.  Defaults
are chosen by block size (\S\ref{app:kernel-stride-defaults}); on
the Qwen3.5-2B shape at $B{=}1, S{=}16$ the checkpoint tensor is
approximately $128$~MiB, roughly $16\times$ smaller than
Route~I's approximately $2$~GiB state tensor.

\begin{algorithm}[h]
\caption{Strided clean-state checkpoint build (one program per chunk $c$).}
\label{alg:ckpt-build}
\small
\begin{algorithmic}[1]
\Require chunk-start state $\mathbf{S}_{[c]}$; clean tokens of chunk $c$; stride $S$
\State $\mathbf{S} \gets \mathbf{S}_{[c]}$
\For{$s = 0, 1, \ldots, M/S - 1$}
  \For{$r = 0, 1, \ldots, S - 1$}
    \State $b \gets cM + sS + r + 1$;\ \ \ $\mathbf{S} \gets f^{\text{blk}}_b(\mathbf{S})$
       \Comment{Eq.~\eqref{eq:f-blk}, in registers}
  \EndFor
  \State \textbf{write} snapshot $\mathbf{S}$ to the $s$-th checkpoint slot of chunk $c$
\EndFor
\end{algorithmic}
\end{algorithm}

\paragraph{Choosing the stride.}
\label{app:kernel-stride-defaults}
$S$ is chosen so that Route~II's extra HBM stays roughly constant
independent of $B$: since the per-chunk snapshot count is
$N_{\text{ckpt}} = M/S = C/(BS)$, we scale $S$ with $M$ so that
$N_{\text{ckpt}}$ stays small ($\mathcal{O}(1)$) per chunk.
With $C{=}64$ fixed, the defaults are $S{=}16$ at $B{=}1$
($N_{\text{ckpt}}{=}4$), $S{=}8$ at $B{=}2$
($N_{\text{ckpt}}{=}4$), $S{=}2$ at $B{=}4$
($N_{\text{ckpt}}{=}8$), and $S{=}\min(8,M)$ otherwise; across
these configurations $N_{\text{ckpt}}\in\{1,4,8\}$ and the checkpoint
tensor occupies between $32$~MiB and $256$~MiB on the Qwen3.5-2B shape.  The per-block replay cost is at
most $S{-}1$ evaluations of $f^{\text{blk}}$ in registers and
remains register-bound because each $f^{\text{blk}}$ is a single
$\mathbb{R}^{d_k\times d_v}$ update that fits in one tile.

\paragraph{Fused forward kernel.}
The forward traverses each chunk block-by-block in order: at block
$j$ the in-register clean state serves as the noisy-recurrence
initial state; after the $B$ noisy-token forward steps, the clean
state is advanced by $f^{\text{blk}}$ and the iteration proceeds.
Only noisy outputs are written to HBM; the clean output is
produced by the separate chunkwise kernel of
\S\ref{app:kernel-single-stream}.

\begin{algorithm}[h]
\caption{Route~II forward kernel (one program per chunk $c$).}
\label{alg:route-twostream-fwd}
\small
\begin{algorithmic}[1]
\Require $\mathbf{S}_{[c]}$, clean tokens of chunk $c$, noisy tokens of chunk $c$
\State $\mathbf{S} \gets \mathbf{S}_{[c]}$ \Comment{in registers}
\For{$j = 0, 1, \ldots, M - 1$}
  \State $b \gets cM + j + 1$;\ \ \ $\tilde{\mathbf{S}}_{b,\text{init}} \gets \mathbf{S}$
       \Comment{in registers; no HBM write}
  \State run Eq.~\eqref{eq:noisy-block-fwd} forward on block $b$; emit $\tilde{\mathbf{o}}_\ell$ for $\ell \in \text{block}(b)$
  \State $\mathbf{S} \gets f^{\text{blk}}_b(\mathbf{S})$
       \Comment{advance clean state, Eq.~\eqref{eq:f-blk}}
\EndFor
\end{algorithmic}
\end{algorithm}

\begin{figure}[h]
\centering
\resizebox{0.85\linewidth}{!}{%
\begin{tikzpicture}[font=\scriptsize, >=Latex, node distance=0.35cm and 0.15cm]
  \definecolor{hbmCol}{HTML}{F0DEC8}
  \definecolor{kernelCol}{HTML}{DCEBD6}
  \definecolor{stepCol}{HTML}{C9DFC3}
  \tikzstyle{band}=[draw, rounded corners=3pt, minimum height=0.9cm,
                    inner sep=4pt, align=center, font=\scriptsize]
  \tikzstyle{tile}=[draw, rounded corners=2pt, minimum height=0.55cm,
                    minimum width=1.9cm, align=center, inner sep=2pt, font=\tiny]
  \tikzstyle{stepbox}=[draw, rounded corners=2pt, fill=stepCol, minimum height=0.7cm,
                       minimum width=2.8cm, align=center, inner sep=2pt, font=\scriptsize]
  \tikzstyle{lab}=[font=\footnotesize\bfseries, text=black!70, anchor=east]

  \node[tile, fill=hbmCol]                  (in1) at (0, 0)    {$\mathbf{S}_{[c]}$};
  \node[tile, fill=hbmCol, right=of in1]    (in2) {strided ckpts};
  \node[tile, fill=hbmCol, right=of in2]    (in3) {$\mathbf{k},\hat{\mathbf{v}},G$ (clean)};
  \node[tile, fill=hbmCol, right=of in3]    (in4) {$\tilde{\mathbf{q}},\tilde{\mathbf{k}},\tilde{\mathbf{v}},\tilde g,\tilde\beta$};
  \node[tile, fill=hbmCol, right=of in4]    (in5) {$\mathbf{d}\tilde{\mathbf{o}}$};
  \node[lab] at ($(in1.west)+(-0.2,0)$) {HBM in};

  \node[band, fill=kernelCol,
        minimum width=11.0cm, minimum height=2.1cm,
        below=1.0 of in3] (kernel)
    {\ };
  \node[anchor=north west, font=\footnotesize\bfseries, text=black!75]
       at ($(kernel.north west)+(0.15,-0.1)$)
       {In-register kernel body: reverse sweep over blocks
        $j=M{-}1,\ldots,0$};
  \node[stepbox, anchor=west] (s1) at ($(kernel.west)+(0.25,-0.35)$)
       {(i) Replay:\\recover $\mathbf{S}_{(b{-}1)B}$};
  \node[stepbox, right=of s1] (s2) {(ii) Noisy bwd (\textsc{PerStepBwd}):\\$\partial / \partial (\tilde{\mathbf{q}},\tilde{\mathbf{k}},\tilde{\mathbf{v}},\tilde g,\tilde\beta), \mathbf{d}\tilde{\mathbf{S}}_{b,\text{init}}$};
  \node[stepbox, right=of s2] (s3) {(iii) \textsc{CleanTransBwd}:\\$\partial / \partial (\mathbf{k},\hat{\mathbf{v}},G), \mathbf{dh}_{b-1}$};
  \draw[->, thick, black!60] (s1.east) -- (s2.west);
  \draw[->, thick, black!60] (s2.east) -- (s3.west);
  \node[lab] at ($(kernel.west)+(-0.2,0)$) {kernel};

  \node[tile, fill=hbmCol, below=1.0 of kernel] (o2)
     {$\partial/\partial(\mathbf{k},\hat{\mathbf{v}},G)$};
  \node[tile, fill=hbmCol, left=of o2] (o1)
     {$\partial/\partial(\tilde{\mathbf{q}},\tilde{\mathbf{k}},\tilde{\mathbf{v}},\tilde g,\tilde\beta)$};
  \node[tile, fill=hbmCol, right=of o2] (o3)
     {$\mathbf{dh}^{\text{inject}}_{[c]}$};
  \node[lab] at ($(o1.west)+(-0.2,0)$) {HBM out};

  \draw[->, line width=1.1pt, black!75] (in3.south) -- (kernel.north);
  \draw[->, line width=1.1pt, black!75] (kernel.south) -- (o2.north);

  \node[anchor=west, align=left, text width=3.2cm, font=\scriptsize\itshape,
        text=black!70]
       at ($(kernel.east)+(0.25,0)$)
       {all intermediates in registers;\\no mid-chunk HBM writes};
\end{tikzpicture}%
}
\caption{Route~II backward kernel dataflow for a single program
(one chunk $c$, one value-dimension tile).  Five inputs are read
from HBM (top); the reverse sweep of
Alg.~\ref{alg:route-twostream-bwd} runs entirely in registers
(middle); three outputs are written back to HBM (bottom).
\textsc{ChunkBwd} (\S\ref{app:kernel-chunk-bwd}) consumes
$\mathbf{dh}^{\text{inject}}_{[c]}$ afterwards.}
\label{fig:route2-kernel}
\end{figure}
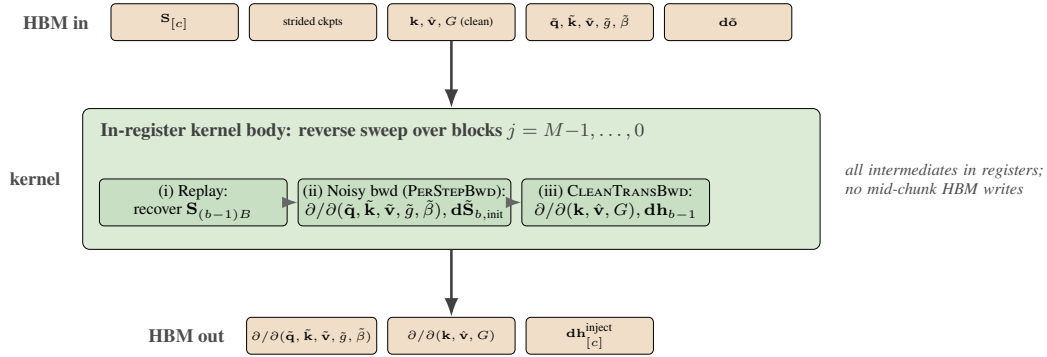

\paragraph{Fused backward kernel.}
The backward traverses each chunk block-by-block in reverse; the
body is given in Alg.~\ref{alg:route-twostream-bwd} and visualized
in Fig.~\ref{fig:route2-kernel}.  All per-block replay, reverse
sweep, and clean-transition work is performed in registers; the
kernel writes only the single chunk-boundary tensor
$\mathbf{dh}^{\text{inject}}_{[c]}$ to HBM, which
\textsc{ChunkBwd} consumes exactly as in Route~I.

\begin{algorithm}[h]
\caption{Route~II backward kernel (one program per chunk $c$, value tile $i_V$).}
\label{alg:route-twostream-bwd}
\small
\begin{algorithmic}[1]
\Require $\mathbf{S}_{[c]}$, strided checkpoints, clean and noisy tokens of chunk $c$, $\mathbf{d}\tilde{\mathbf{o}}$
\State $\mathbf{dh} \gets \mathbf{0}$ \Comment{in registers}
\For{$j = M-1, \ldots, 0$}
  \State $b \gets cM + j + 1$
  \State \textsc{Replay:} $\mathbf{S}_{(b-1)B} \gets$ load nearest checkpoint, apply $\leq S{-}1$ copies of $f^{\text{blk}}$
  \State run Eq.~\eqref{eq:noisy-block-fwd} forward for $B$ steps (re-compute caches in registers)
  \State $\bigl(\mathbf{d}\tilde{\mathbf{q}}, \mathbf{d}\tilde{\mathbf{k}}, \mathbf{d}\tilde{\mathbf{v}}, d\tilde g, d\tilde\beta, \mathbf{d}\tilde{\mathbf{S}}_{b,\text{init}}\bigr) \gets$ block-end-readout reverse sweep (Eq.~\eqref{eq:noisy-readout-bwd} + \textsc{PerStepBwd} with $\mathbf{do}{=}\mathbf{0}$)
  \State $(\mathbf{dk}_t, \mathbf{d}\hat{\mathbf{v}}_t, dG_t)_{t\in\text{block}(b)} \gets$ \textsc{CleanTransBwd}$(\mathbf{dh}, \mathbf{S}_{(b-1)B}, \ldots)$
  \State $\mathbf{dh} \gets \mathbf{d}\tilde{\mathbf{S}}_{b,\text{init}} + \exp\,\bigl(G_{bB}-G_{(b-1)B}\bigr)\,\mathbf{dh}$
       \Comment{Eq.~\eqref{eq:dh-block-recursion}}
\EndFor
\State \textbf{write} $\mathbf{dh}^{\text{inject}}_{[c]} \gets \mathbf{dh}$ \Comment{consumed by \textsc{ChunkBwd}}
\end{algorithmic}
\end{algorithm}

\paragraph{Parallelism structure.}
The backward of Fig.~\ref{fig:kernel-grad-flow} groups into three
parallelism tiers once Levels~1 and~2 (within a block) are folded
together: \emph{tier~A} performs one block-end output contraction
plus $B$ per-step reverse steps inside a block; \emph{tier~B}
sweeps $M$ such blocks inside a chunk via \textsc{CleanTransBwd}
and the block-level recursion; and \emph{tier~C} sweeps $N_C$
chunks across the sequence via \textsc{ChunkBwd}.  Route~II
performs tiers~A and~B entirely in registers inside the fused
kernel and delegates tier~C to \textsc{ChunkBwd}, the same
chunk-level matmul scan used by the AR baseline.  Since tier~B is
strictly within-chunk, chunks are independent at the fused-kernel
level, which enables Route~II's grid of
$N_C\cdot B_{\text{batch}}\cdot H\cdot\lceil d_v/b_v\rceil$
programs; Route~I combines tier~B with tier~C in the same kernel
and is therefore restricted to a grid of
$B_{\text{batch}}\cdot H\cdot\lceil d_v/b_v\rceil$ programs.

\subsubsection{Implementation comparison}
\label{app:route-comparison}

The two routes produce identical gradients, so the selection
between them is governed by resource trade-offs.  Route~I has
fewer moving parts: three small kernels per direction and no
strided-checkpoint bookkeeping.  Route~II reduces peak memory and
increases backward parallelism.  In the deployed \model
configuration ($B < 16$) Route~II dominates on every axis
considered below; at $B \ge 16$ Route~I is faster on latency
while Route~II retains the peak-memory advantage.
The three axes are listed in order of importance at small $B$.

\paragraph{Peak memory.}
Route~I materializes an $L/B$-scaling tensor of block-boundary
clean states: $\{\mathbf{S}_{(b-1)B}\}_{b=1}^{L/B}$ is written in
forward stage~2 and kept alive until consumed by the block-local
noisy forward (and, in general, by the fill-in backward).  Its
size is
\begin{equation*}
  \text{HBM}_{\text{Route~I, states}}
  \;=\;
  \underbrace{L/B}_{\text{grows as $B$ shrinks}}
  \cdot\,H\cdot d_k\cdot d_v\cdot\text{(bytes per element)} ,
\end{equation*}
which at $B{=}1$ on the Qwen3.5-2B shape is approximately $2$~GiB
per layer per sample in bf16; across the backbone's $18$ GDN
layers this exhausts the available HBM at the small block
sizes used in \model training.  Route~II does not
form this tensor: $\mathbf{S}_{(b-1)B}$ is recomputed in registers
from the nearest strided checkpoint and used immediately.  The
only extra HBM Route~II allocates is the strided-checkpoint tensor
of Alg.~\ref{alg:ckpt-build}, which scales as $L/(BS)$ and at the
$B{=}1$ default $S{=}16$ on the same shape is approximately
$128$~MiB, roughly $16\times$ smaller.

\paragraph{Per-chunk parallelism.}
Route~I's backward combines cross-chunk propagation with the
per-block reverse sweep in a single kernel, so its noisy-backward
grid is
$B_{\text{batch}} \cdot H \cdot \lceil d_v / b_v \rceil$
(where $b_v$ is the value-dimension tile size).  On the Qwen3.5-2B
shape with $B_{\text{batch}}{=}1$, $H{=}16$, $b_v{=}32$ this
yields $64$ programs, well below the $108$ SMs of an A100.
Route~II delegates the cross-chunk propagation to \textsc{ChunkBwd}, making each
chunk's fused kernel independent; the grid becomes
$N_C \cdot B_{\text{batch}} \cdot H \cdot \lceil d_v / b_v \rceil$
and at $N_C{=}64$ reaches $4096$ programs, a $64\times$ increase
that saturates SM occupancy.  The shared cross-chunk
\textsc{ChunkBwd} kernel itself runs the same $64$-program
$(B_{\text{batch}}\!\cdot\!H\!\cdot\!\lceil d_v/b_v \rceil)$ grid
in both routes and in the AR baseline.

\paragraph{HBM hand-off tensors.}
Route~I's backward runs three kernels in strict sequence; each
waits for the previous kernel's output tensor to be fully written
to HBM before launching.  The two internal hand-off tensors have
sizes $L/B \cdot H \cdot d_k \cdot d_v$ (noisy-init gradients) and
$N_C \cdot H \cdot d_k \cdot d_v$ (chunk-boundary gradients); the
former again carries the $L/B$ factor.  Route~II collapses the
first two stages and emits only the chunk-boundary tensor
$\mathbf{dh}^{\text{inject}}_{[c]}$ ($N_C$-sized) directly into
\textsc{ChunkBwd}, halving the hand-off count and removing the
$L/B$-sized intermediate.

\paragraph{Kernel latency and peak memory benchmark.}
Tables~\ref{tab:route-comparison-gda} and~\ref{tab:route-comparison-shortconv} report
an empirical wall-clock and peak-memory comparison of Route~I
versus Route~II for the GDR and the
ShortConv kernels, respectively, across block sizes
$B \in \{1, 2, 4, 8, 16, 32\}$ on the Qwen3.5-2B training shape.
On the GDR, Route~II cuts total wall-clock from
$135.1$~ms to $37.7$~ms and peak memory from $18.14$~GiB to
$456$~MiB at $B{=}1$; Route~I overtakes Route~II at $B{\ge}16$
once the chunk-level matmul saturates tensor cores.  For the
ShortConv, Route~II is faster than Route~I at every
$B$ in the sweep and holds peak memory at
${\sim}294$~MiB throughout.  We therefore auto-dispatch Route~II
for the GDR at $B{<}16$ and Route~I at $B{\ge}16$,
and use Route~II for the ShortConv at every block size.

\begin{table}[h]
\centering
\footnotesize
\setlength{\tabcolsep}{4pt}
\caption{\textbf{GDR: Route I (Chunk-then-Refine)
vs.\ Route II (Fused Two-Stream)} wall-clock and peak memory at
varying diffusion-block size $B$.  One A100-80GB, bf16, $L=8192$
(half clean, half noisy) on the Qwen3.5-2B GDN shape
($H=16$, $d_k=d_v=128$).  Wall-clock is mean $\pm$ std (ms) over
three outer runs of ten iterations each (trimmed mean inside each
outer run); peak is the maximum live-memory delta (MiB) across
the combined forward and backward window.  Bold marks the faster
/ smaller of the two routes at each block size.}
\label{tab:route-comparison-gda}
\begin{tabular*}{\linewidth}{@{\extracolsep{\fill}} r l r r r r @{}}
\toprule
Block size $B$ & Route & fwd (ms) & bwd (ms) & total (ms) & peak (MiB) \\
\midrule
 1 & Route I  & $26.36 \pm 0.72$ & $108.74 \pm 15.27$ & $135.10$ & $18\,576$ \\
 1 & Route II & $\mathbf{15.99 \pm 1.10}$ & $\mathbf{21.70 \pm 0.19}$ & $\mathbf{37.69}$ & $\mathbf{456}$ \\
\midrule
 2 & Route I  & $\mathbf{16.70 \pm 1.10}$ & $61.45 \pm 0.37$ & $78.14$ & $9\,360$ \\
 2 & Route II & $19.19 \pm 1.63$ & $\mathbf{21.22 \pm 0.29}$ & $\mathbf{40.42}$ & $\mathbf{456}$ \\
\midrule
 4 & Route I  & $\mathbf{13.80 \pm 0.66}$ & $48.35 \pm 3.35$ & $62.14$ & $4\,752$ \\
 4 & Route II & $17.64 \pm 1.31$ & $\mathbf{20.61 \pm 0.02}$ & $\mathbf{38.26}$ & $\mathbf{712}$ \\
\midrule
 8 & Route I  & $\mathbf{12.00 \pm 0.89}$ & $39.07 \pm 1.29$ & $51.07$ & $2\,448$ \\
 8 & Route II & $15.65 \pm 1.46$ & $\mathbf{23.19 \pm 0.36}$ & $\mathbf{38.84}$ & $\mathbf{378}$ \\
\midrule
16 & Route I  & $\mathbf{6.61 \pm 0.71}$ & $\mathbf{12.07 \pm 0.50}$ & $\mathbf{18.68}$ & $1\,296$ \\
16 & Route II & $20.00 \pm 4.59$ & $28.61 \pm 2.16$ & $48.61$ & $\mathbf{378}$ \\
\midrule
32 & Route I  & $\mathbf{9.27 \pm 1.27}$ & $\mathbf{11.76 \pm 2.29}$ & $\mathbf{21.03}$ & $720$ \\
32 & Route II & $16.08 \pm 0.78$ & $36.96 \pm 0.34$ & $53.04$ & $\mathbf{378}$ \\
\bottomrule
\end{tabular*}
\end{table}

\subsection{1D Causal ShortConv: implementation routes}
\label{app:short-conv}

The 1D Causal ShortConv of width $W$ (typically $W{=}4$)
is the second sub-component of a GDN layer.  It applies
a depthwise causal 1D convolution to each channel of the input
stream before the linear projections that feed the GDR
of \S\ref{app:gdn-routes}, conditioning each token's
$(\mathbf{q}_t, \mathbf{k}_t, \mathbf{v}_t)$ on its local temporal
context.  Under \model's two-stream training, ShortConv is subject
to the same class of constraints analyzed
for the GDR in
\S\ref{app:kernel-problem}--\S\ref{app:route-comparison}: each
noisy output position must read from the clean stream whenever its
$W$-wide receptive field extends across the block boundary, while
the clean stream is unaffected by the noisy stream.  This
subsection applies the GDR construction to ShortConv,
identifies the two structural simplifications that
reduce its cost, and states the two implementation routes in the
same Route~I / Route~II terminology.

\subsubsection{Two-stream contract for a width-$W$ causal 1D convolution}
\label{app:short-conv-contract}

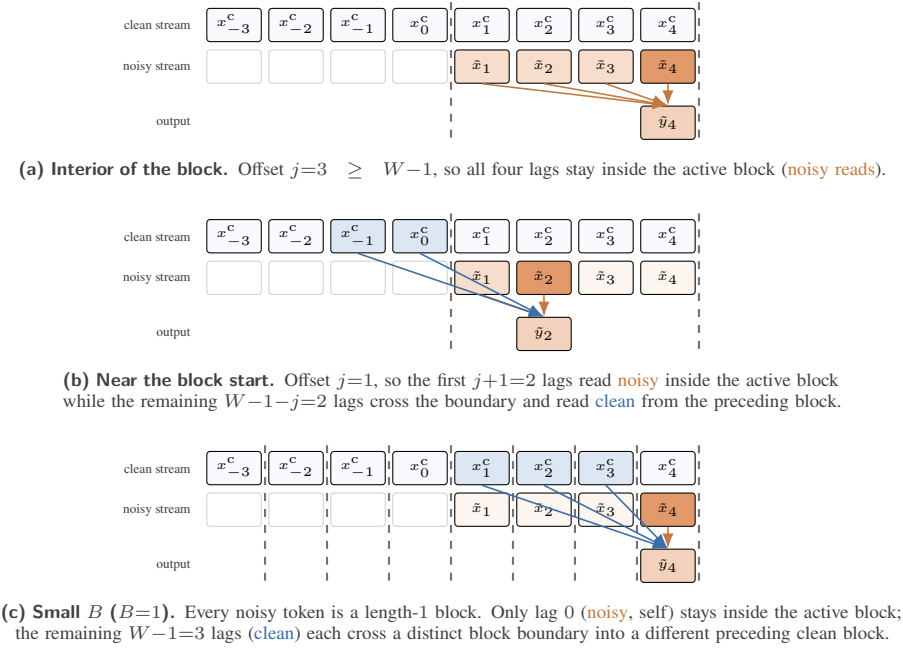
\begin{figure}[t]
\centering
\definecolor{cleanBox}{HTML}{DCE7F3}
\definecolor{noisyBox}{HTML}{F7DDCB}
\definecolor{activeBox}{HTML}{E29A6B}
\definecolor{lagNoisy}{HTML}{C17A40}
\definecolor{lagClean}{HTML}{3B6EA8}

\newcommand{\shortconvRowLabel}[2]{%
  \node[font=\tiny, anchor=east, text=black!70] at (#1, #2) {}}

\begin{subfigure}[t]{\linewidth}\centering
\begin{tikzpicture}[font=\scriptsize, >=Latex, x=0.82cm, y=0.55cm]
  \tikzstyle{tok}=[draw, rounded corners=1.2pt, minimum height=0.44cm,
                   minimum width=0.72cm, align=center, inner sep=0pt,
                   font=\tiny]
  \tikzstyle{dim}=[draw=black!18, fill=white]
  \tikzstyle{bnd}=[dashed, thick, black!55]

  \node[font=\tiny, anchor=east, text=black!75] at (-3.55, 1.0)  {clean stream};
  \node[font=\tiny, anchor=east, text=black!75] at (-3.55, 0)    {noisy stream};
  \node[font=\tiny, anchor=east, text=black!75] at (-3.55, -1.35) {output};

  \draw[bnd] (0.5, 1.55) -- (0.5, -1.75);
  \draw[bnd] (4.5, 1.55) -- (4.5, -1.75);

  \foreach \p in {-3,-2,-1,0} \node[tok, fill=cleanBox!20] (c\p) at (\p, 1.0) {$x^{\mathrm{c}}_{\p}$};
  \foreach \p in {1,2,3,4} \node[tok, fill=cleanBox!20] at (\p, 1.0) {$x^{\mathrm{c}}_{\p}$};

  \foreach \p in {-3,-2,-1,0} \node[tok, dim] at (\p, 0) {};
  \node[tok, fill=noisyBox]  (n1) at (1, 0) {$\tilde{x}_{1}$};
  \node[tok, fill=noisyBox]  (n2) at (2, 0) {$\tilde{x}_{2}$};
  \node[tok, fill=noisyBox]  (n3) at (3, 0) {$\tilde{x}_{3}$};
  \node[tok, fill=activeBox] (n4) at (4, 0) {$\tilde{x}_{4}$};

  \node[tok, fill=activeBox!45] (ya) at (4, -1.35) {$\tilde{y}_{4}$};

  \draw[->, semithick, lagNoisy] (n4.south) -- (ya.north);
  \draw[->, semithick, lagNoisy] (n3.south) -- (ya.north);
  \draw[->, semithick, lagNoisy] (n2.south) -- (ya.north);
  \draw[->, semithick, lagNoisy] (n1.south) -- (ya.north);

  \node[font=\scriptsize, anchor=north, align=center, text width=0.82\linewidth,
        text=black!75]
       at (0.5, -2.05)
       {\textbf{(a) Interior of the block.}  Offset $j{=}3 \ge W{-}1$, so all four
        lags stay inside the active block (\textcolor{lagNoisy}{noisy reads}).};
\end{tikzpicture}
\end{subfigure}

\vspace{8pt}

\begin{subfigure}[t]{\linewidth}\centering
\begin{tikzpicture}[font=\scriptsize, >=Latex, x=0.82cm, y=0.55cm]
  \tikzstyle{tok}=[draw, rounded corners=1.2pt, minimum height=0.44cm,
                   minimum width=0.72cm, align=center, inner sep=0pt,
                   font=\tiny]
  \tikzstyle{dim}=[draw=black!18, fill=white]
  \tikzstyle{bnd}=[dashed, thick, black!55]

  \node[font=\tiny, anchor=east, text=black!75] at (-3.55, 1.0)  {clean stream};
  \node[font=\tiny, anchor=east, text=black!75] at (-3.55, 0)    {noisy stream};
  \node[font=\tiny, anchor=east, text=black!75] at (-3.55, -1.35) {output};

  \draw[bnd] (0.5, 1.55) -- (0.5, -1.75);
  \draw[bnd] (4.5, 1.55) -- (4.5, -1.75);

  \foreach \p in {-3,-2} \node[tok, fill=cleanBox!20] (c\p) at (\p, 1.0) {$x^{\mathrm{c}}_{\p}$};
  \foreach \p in {-1,0} \node[tok, fill=cleanBox] (c\p) at (\p, 1.0) {$x^{\mathrm{c}}_{\p}$};
  \foreach \p in {1,2,3,4} \node[tok, fill=cleanBox!20] at (\p, 1.0) {$x^{\mathrm{c}}_{\p}$};

  \foreach \p in {-3,-2,-1,0} \node[tok, dim] at (\p, 0) {};
  \node[tok, fill=noisyBox]  (n1) at (1, 0) {$\tilde{x}_{1}$};
  \node[tok, fill=activeBox] (n2) at (2, 0) {$\tilde{x}_{2}$};
  \node[tok, fill=noisyBox!30]  (n3) at (3, 0) {$\tilde{x}_{3}$};
  \node[tok, fill=noisyBox!30]  (n4) at (4, 0) {$\tilde{x}_{4}$};

  \node[tok, fill=activeBox!45] (yb) at (2, -1.35) {$\tilde{y}_{2}$};

  \draw[->, semithick, lagNoisy] (n2.south) -- (yb.north);
  \draw[->, semithick, lagNoisy] (n1.south) -- (yb.north);

  \draw[->, semithick, lagClean] (c0.south)  -- (yb.north);
  \draw[->, semithick, lagClean] (c-1.south) -- (yb.north);

  \node[font=\scriptsize, anchor=north, align=center, text width=0.82\linewidth,
        text=black!75]
       at (0.5, -2.05)
       {\textbf{(b) Near the block start.}  Offset $j{=}1$, so the first
        $j{+}1{=}2$ lags read \textcolor{lagNoisy}{noisy} inside the active
        block while the remaining $W{-}1{-}j{=}2$ lags cross the boundary
        and read \textcolor{lagClean}{clean} from the preceding block.};
\end{tikzpicture}
\end{subfigure}

\vspace{8pt}

\begin{subfigure}[t]{\linewidth}\centering
\begin{tikzpicture}[font=\scriptsize, >=Latex, x=0.82cm, y=0.55cm]
  \tikzstyle{tok}=[draw, rounded corners=1.2pt, minimum height=0.44cm,
                   minimum width=0.72cm, align=center, inner sep=0pt,
                   font=\tiny]
  \tikzstyle{dim}=[draw=black!18, fill=white]
  \tikzstyle{bnd}=[dashed, thick, black!55]

  \node[font=\tiny, anchor=east, text=black!75] at (-3.55, 1.0)  {clean stream};
  \node[font=\tiny, anchor=east, text=black!75] at (-3.55, 0)    {noisy stream};
  \node[font=\tiny, anchor=east, text=black!75] at (-3.55, -1.35) {output};

  \foreach \b in {-2.5,-1.5,-0.5,0.5,1.5,2.5,3.5,4.5}
    \draw[bnd] (\b, 1.55) -- (\b, -1.75);

  \node[tok, fill=cleanBox!20] at (-3, 1.0) {$x^{\mathrm{c}}_{-3}$};
  \node[tok, fill=cleanBox!20] at (-2, 1.0) {$x^{\mathrm{c}}_{-2}$};
  \node[tok, fill=cleanBox!20] at (-1, 1.0) {$x^{\mathrm{c}}_{-1}$};
  \node[tok, fill=cleanBox!20] at (0, 1.0) {$x^{\mathrm{c}}_{0}$};
  \node[tok, fill=cleanBox] (c1r) at (1, 1.0) {$x^{\mathrm{c}}_{1}$};
  \node[tok, fill=cleanBox] (c2r) at (2, 1.0) {$x^{\mathrm{c}}_{2}$};
  \node[tok, fill=cleanBox] (c3r) at (3, 1.0) {$x^{\mathrm{c}}_{3}$};
  \node[tok, fill=cleanBox!20] at (4, 1.0) {$x^{\mathrm{c}}_{4}$};

  \node[tok, dim]              at (-3, 0) {};
  \node[tok, dim]              at (-2, 0) {};
  \node[tok, dim]              at (-1, 0) {};
  \node[tok, dim]              at (0, 0) {};
  \node[tok, fill=noisyBox!30] at (1, 0) {$\tilde{x}_{1}$};
  \node[tok, fill=noisyBox!30] at (2, 0) {$\tilde{x}_{2}$};
  \node[tok, fill=noisyBox!30] at (3, 0) {$\tilde{x}_{3}$};
  \node[tok, fill=activeBox]  (n4) at (4, 0) {$\tilde{x}_{4}$};

  \node[tok, fill=activeBox!45] (yc) at (4, -1.35) {$\tilde{y}_{4}$};
  \draw[->, semithick, lagNoisy] (n4.south) -- (yc.north);
  \draw[->, semithick, lagClean] (c3r.south) -- (yc.north);
  \draw[->, semithick, lagClean] (c2r.south) -- (yc.north);
  \draw[->, semithick, lagClean] (c1r.south) -- (yc.north);

  \node[font=\scriptsize, anchor=north, align=center, text width=0.82\linewidth,
        text=black!75]
       at (0.5, -2.05)
       {\textbf{(c) Small $B$ ($B{=}1$).}  Every noisy
        token is a length-$1$ block.  Only lag $0$ (\textcolor{lagNoisy}{noisy},
        self) stays inside the active block; the remaining $W{-}1{=}3$ lags
        (\textcolor{lagClean}{clean}) each cross a distinct block boundary into
        a different preceding clean block.};
\end{tikzpicture}
\end{subfigure}

\caption{Two-stream read pattern of the width-$W{=}4$ causal 1D
ShortConv for a noisy output $\tilde{y}_\ell$, shown
under three regimes: (a) interior ($j \ge W{-}1$); (b) straddle
($j < W{-}1$, clean lags in the single preceding block); (c)
small-$B$ ($B{=}1$, each non-self
clean lag lands in a different preceding block).  Per-panel
streams: clean (top row), noisy (middle row), active output
(bottom row).  \textcolor{lagNoisy}{Orange arrows}: noisy reads;
\textcolor{lagClean}{blue arrows}: clean reads; dashed lines:
block boundaries; faded tokens: irrelevant to $\tilde{y}_\ell$.}
\label{fig:short-conv-lags}
\end{figure}

Let $\mathbf{w} \in \mathbb{R}^{D \times W}$ be the depthwise
filter applied by a standard single-stream causal convolution to
a length-$L$ input of $D$-dimensional tokens; we write $w_i$ for
its $i$-th lag slice (acting channel-wise).  Under the two-stream
contract, the clean-side output $y^{\mathrm{c}}_t$ at position $t$
is computed from the clean stream in the standard form:
\begin{equation}
\label{eq:short-conv-clean}
  y^{\mathrm{c}}_t \;=\; \sum_{i=0}^{W-1} w_i\,x^{\mathrm{c}}_{t-i},
\end{equation}
where $x^{\mathrm{c}}_{t-i} \in \mathbb{R}^D$ is the clean-stream
input at lag $i$ (with out-of-range entries treated as zero) and
$w_i\,x^{\mathrm{c}}_{t-i}$ denotes the depthwise (channel-wise)
product.  The noisy-side output $\tilde{y}_\ell$ at position
$\ell$ inside block $b$ (positions $(b{-}1)B+1, \ldots, bB$) reads
from the \emph{noisy} stream for lags that remain within block
$b$ and from the \emph{clean} stream for lags that fall before
block $b$'s start:
\begin{equation}
\label{eq:short-conv-noisy}
  \tilde{y}_\ell
  \;=\;\sum_{i=0}^{W-1} w_i\,z_{\ell,i},
  \qquad
  z_{\ell,i} \;=\;
  \begin{cases}
    \tilde{x}_{\ell-i} & \text{if } \ell-i \ge (b{-}1)B+1, \\
    x^{\mathrm{c}}_{\ell-i} & \text{otherwise},
  \end{cases}
\end{equation}
where $z_{\ell,i} \in \mathbb{R}^D$ is the lag-$i$ input read by
the noisy output at position $\ell$, selected from the noisy
stream when the lag position falls inside block $b$ and from the
clean stream otherwise.
The condition $\ell - i \ge (b{-}1)B+1$ is the ShortConv
analog of the clean-to-noisy visibility rule stated in
Challenge~1 (\S\ref{app:kernel-problem}): noisy tokens read from
the clean stream only across the block boundary.
Figure~\ref{fig:short-conv-lags} illustrates the rule for
$W{=}4$.

\subsubsection{Analysis of implementation difficulty compared to Gated Delta Rule}
\label{app:short-conv-easier}

Two structural simplifications reduce the cost of the
ShortConv case relative to the GDR case.

\paragraph{The noisy initial state is a raw tensor slice.}
The GDR's block-boundary clean state is a
$d_k \times d_v$ matrix obtained by running $B$ per-step
recurrence updates (Eq.~\eqref{eq:per-step-fwd}) starting from
$\mathbf{S}_{[c]}$, incurring a serial replay or checkpoint lookup
(\S\ref{app:route-refine}, \S\ref{app:route-twostream}).  The ShortConv seed required
by Eq.~\eqref{eq:short-conv-noisy} is the last $W{-}1$ clean tokens
preceding block $b$'s start,
$\bigl(x^{\mathrm{c}}_{(b{-}1)B-(W-1)+1}, \ldots, x^{\mathrm{c}}_{(b{-}1)B}\bigr)$,
which is a slice of the clean-stream input tensor.  No recurrence
is replayed and no checkpoint tensor is constructed; the seed is
read directly from the clean activations already consumed by the
clean-side convolution.

\paragraph{The backward has no block-level state recursion.}
The GDR's backward (Fig.~\ref{fig:kernel-grad-flow})
composes four levels: a block-end readout (Level~1,
Eq.~\eqref{eq:noisy-readout-bwd}), a per-step reverse sweep within a
block (Level~2, Eq.~\eqref{eq:per-step-bwd}), a block-level reverse
recursion within a chunk through $f^{\text{blk}}_b$ (Level~3,
Eq.~\eqref{eq:dh-block-recursion}), and a cross-chunk matmul scan
on chunk-boundary states (Level~4,
Eq.~\eqref{eq:chunk-bwd-scan}).  The ShortConv backward omits the
block-level state recursion (Level~3): the gradient deposited by
each noisy output into its clean-side lag positions is scattered
directly into the corresponding positions of $\mathbf{dx}$,
without a block-level state-gradient recursion, because the
noisy-side seed is not a recurrence terminus.  The cross-chunk
Level~4 reduces to the standard clean-side causal-conv backward and
requires no additional hand-off.

\subsubsection{Route~I: batched initial-state}
\label{app:short-conv-route-i}

The ShortConv analog of Route~I materializes the noisy
initial states in HBM and then invokes the stock single-stream
causal-conv kernel in batch.  Forward:

\begin{algorithm}[h]
\caption{ShortConv Route~I forward.}
\label{alg:short-conv-route-i-fwd}
\small
\begin{algorithmic}[1]
\Require clean tokens $\{x^{\mathrm{c}}_t\}$, noisy tokens $\{\tilde{x}_\ell\}$, filter $\mathbf{w}$, block size $B$
\State run the stock causal-conv kernel on $\{x^{\mathrm{c}}_t\}$ to produce $\{y^{\mathrm{c}}_t\}$
\State build $\mathbf{H}^{\mathrm{init}} \in \mathbb{R}^{L/B \times D \times (W-1)}$ by slicing the last $W{-}1$ clean tokens before each noisy block (zero-pad at document starts)
\State reshape $\{\tilde{x}_\ell\}$ into $L/B$ independent length-$B$ sub-sequences
\State run the stock causal-conv kernel in batch over the $L/B$ sub-sequences, each with $\mathbf{H}^{\mathrm{init}}[b{-}1]$ as its prepended context
\State return $\{y^{\mathrm{c}}_t\}$ and $\{\tilde{y}_\ell\}$
\end{algorithmic}
\end{algorithm}

Backward mirrors this structure.  The batched noisy backward emits
$\mathbf{d}\tilde{\mathbf{x}}$, the filter/bias gradients, and a
per-block gradient $\mathbf{dH}^{\mathrm{init}}[b{-}1]$ with respect
to each prepended context.  A final scatter-add sends each
$\mathbf{dH}^{\mathrm{init}}[b{-}1]$ back to the last $W{-}1$
positions of $\mathbf{dx}$ that it was read from, and
the clean-side causal-conv backward is run once on the full clean
stream.  The state tensor $\mathbf{H}^{\mathrm{init}}$ lives in HBM
throughout; its size is $L/B \cdot D \cdot (W-1)$ bytes (times
precision), the ShortConv analog of Route~I's
$L/B$-scaling term for the GDR.

\subsubsection{Route~II: fused two-stream}
\label{app:short-conv-route-ii}

Route~II replaces both the init-state construction and the batched
noisy conv with a single fused Triton kernel that implements
Eq.~\eqref{eq:short-conv-noisy} in place.

\begin{algorithm}[h]
\caption{ShortConv Route~II forward (one program tile per $(\ell\text{-chunk}, D\text{-tile})$ pair).}
\label{alg:short-conv-route-ii-fwd}
\small
\begin{algorithmic}[1]
\Require clean tokens $\{x^{\mathrm{c}}_t\}$, noisy tokens $\{\tilde{x}_\ell\}$, filter $\mathbf{w}$, block size $B$
\For{each noisy output position $\ell$ in the tile}
  \State compute $\ell$'s block start $\ell_{\mathrm{blk}} = \lfloor (\ell-1)/B \rfloor \cdot B + 1$
  \For{lag $i = 0, 1, \ldots, W-1$}
    \State $z_{\ell,i} \gets \tilde{x}_{\ell-i}$ if $\ell-i \ge \ell_{\mathrm{blk}}$, else $x^{\mathrm{c}}_{\ell-i}$
  \EndFor
  \State $\tilde{y}_\ell \gets \sum_i w_i\,z_{\ell,i}$
\EndFor
\end{algorithmic}
\end{algorithm}

A companion fused kernel implements the backward: for each noisy
output position $\ell$ and lag $i$, it computes the contribution
$w_i\,\partial\mathcal{L}/\partial \tilde{y}_\ell$ and atomically
adds it to $\mathbf{d}\tilde{\mathbf{x}}_{\ell-i}$ if
$\ell-i \ge \ell_{\mathrm{blk}}$, or to
$\mathbf{dx}_{\ell-i}$ otherwise; filter and bias
gradients are accumulated into per-tile partial tensors and
reduced.  The clean-side output and its gradient are produced by
the stock single-stream causal-conv forward and backward unchanged
from the AR baseline.  No $L/B$-scaling HBM tensor is materialized
and no batched noisy-conv kernel is launched.

\subsubsection{Implementation comparison}
\label{app:short-conv-wins}

The three axes of \S\ref{app:route-comparison} reapply in a
reordered ranking, since the ShortConv state is small
relative to the GDR state.  In the deployed \model
configuration ($W{=}4$ with silu on CUDA, small $B$), Route~II is
preferable on all three axes; outside this configuration Route~I
remains a fallback.

\paragraph{Per-chunk parallelism.}
Route~I's batched noisy conv has grid $(L/B) \cdot \lceil D/b_d \rceil$,
where $b_d$ is the channel-dimension tile size; Route~II's fused
kernel instead tiles along the $t$-dimension, with grid
$N_t \cdot \lceil D/b_d \rceil$ for $N_t$ the $t$-dimension chunk
count, decoupling the launchable-program count from the block size $B$.

\paragraph{Kernel-fusion overhead.}
Route~I issues three kernel calls per direction (forward: clean
conv, init-state build, batched noisy conv; backward: batched
noisy backward, init-state scatter, clean conv backward); Route~II
issues two (clean conv and fused two-stream conv, per direction).
The launch-overhead savings are material at the short sequence
lengths typical of ShortConv layers.

\paragraph{Peak memory.}
Route~I's $\mathbf{H}^{\mathrm{init}}$ has size
$L/B \cdot D \cdot (W-1)$, a few hundred MB per layer per sample at
the smallest block sizes and shrinking with $B$.  This is much
smaller than the analogous $L/B$-scaling tensor in the GDR
case (gigabytes), so memory is not the dominant factor in
selecting Route~II for the ShortConv.

\paragraph{Kernel latency and peak memory benchmark.}
Table~\ref{tab:route-comparison-shortconv} reports the empirical
wall-clock and peak-memory comparison of ShortConv
Route~I versus Route~II at $B \in \{1, 2, 4, 8, 16, 32\}$ on the
Qwen3.5-2B ShortConv shape.  Route~II is faster than
Route~I at every block size in the sweep and keeps peak memory
within ${\sim}294$~MiB throughout, so \model uses Route~II for
the ShortConv at every block size.

\begin{table}[h]
\centering
\footnotesize
\setlength{\tabcolsep}{4pt}
\caption{\textbf{1D Causal ShortConv: Route I (batched
initial-state) vs.\ Route II (fused two-stream)} wall-clock and
peak memory at varying $B$.  Same hardware and setup as
Table~\ref{tab:route-comparison-gda}, on the Qwen3.5-2B
ShortConv shape ($D=6144$, $W=4$, $t$-dimension tile size
$BT=64$).  Bold marks
the faster / smaller of the two routes at each block size.}
\label{tab:route-comparison-shortconv}
\begin{tabular*}{\linewidth}{@{\extracolsep{\fill}} r l r r r r @{}}
\toprule
Block size $B$ & Route & fwd (ms) & bwd (ms) & total (ms) & peak (MiB) \\
\midrule
 1 & Route I  & $13.41 \pm 0.86$ & $42.65 \pm 0.66$ & $56.06$ & $1\,452$ \\
 1 & Route II & $\mathbf{6.26 \pm 0.60}$ & $\mathbf{3.40 \pm 0.11}$ & $\mathbf{9.65}$ & $\mathbf{294}$ \\
\midrule
 2 & Route I  & $\mathbf{11.99 \pm 1.12}$ & $21.07 \pm 0.22$ & $33.06$ & $816$ \\
 2 & Route II & $13.22 \pm 10.73$ & $\mathbf{4.25 \pm 1.09}$ & $\mathbf{17.47}$ & $\mathbf{294}$ \\
\midrule
 4 & Route I  & $12.38 \pm 7.43$ & $11.06 \pm 0.08$ & $23.44$ & $372$ \\
 4 & Route II & $\mathbf{7.00 \pm 0.98}$ & $\mathbf{3.56 \pm 0.34}$ & $\mathbf{10.55}$ & $\mathbf{294}$ \\
\midrule
 8 & Route I  & $11.44 \pm 0.54$ & $6.30 \pm 0.08$ & $17.74$ & $330$ \\
 8 & Route II & $\mathbf{6.86 \pm 0.51}$ & $\mathbf{3.09 \pm 0.15}$ & $\mathbf{9.95}$ & $\mathbf{294}$ \\
\midrule
16 & Route I  & $16.72 \pm 10.98$ & $3.64 \pm 0.01$ & $20.36$ & $309$ \\
16 & Route II & $\mathbf{6.71 \pm 0.14}$ & $\mathbf{2.93 \pm 0.27}$ & $\mathbf{9.64}$ & $\mathbf{294}$ \\
\midrule
32 & Route I  & $9.11 \pm 1.39$ & $\mathbf{2.43 \pm 0.01}$ & $11.54$ & $299$ \\
32 & Route II & $\mathbf{6.86 \pm 0.20}$ & $2.86 \pm 0.12$ & $\mathbf{9.72}$ & $\mathbf{294}$ \\
\bottomrule
\end{tabular*}
\end{table}

\paragraph{Summary. }
We auto-select Route~II for every benchmarked ShortConv
configuration supported by the fused kernel ($W \le 4$ with
silu/swish or no activation on CUDA), and retain Route~I as a
fallback for configurations outside it (e.g.\ alternative
activations or larger filter widths).

\subsection{Document-packed training}
\label{app:doc-packed}

This subsection specifies the document-packing guarantees
required by Challenge~2 of \S\ref{app:kernel-problem}.  The
kernels in \S\ref{app:gdn-routes}--\S\ref{app:short-conv} are
described as operating on a single $L$-token document; for
training throughput, multiple shorter documents are packed into
each training sequence with per-document start offsets
$\mathbf{cu} = (c_0, c_1, \ldots, c_{N_{\text{doc}}})$,
$c_0 = 0$, $c_{N_{\text{doc}}} = L$ (the standard
\texttt{cu\_seqlens} layout).  The two-stream mask is defined
per-document; under packing, the block-diffusion visibility rule
is composed with a document-level mask that forbids cross-document
attention, so each document is trained as if in isolation.  This
subsection states how the GDR and 1D Causal ShortConv
kernels of
\S\ref{app:gdn-routes}--\S\ref{app:short-conv} preserve
single-document semantics under packing.

\paragraph{Constraint: document starts are block-aligned.}
Every entry of $\mathbf{cu}[:-1]$ must be a multiple of the
diffusion-block size $B$, i.e.\ every document's first token must
also be a block start.  This is a hard precondition: the
two-stream block mask of \S\ref{sec:method-training} is defined
on block boundaries, so a document starting in the interior of a
block would cause its clean-to-noisy visibility to straddle a
document boundary, violating the doc-level visibility contract.
The final packed tail ($c_{N_{\text{doc}}}$) is permitted to be a
partial block; the implementation zero-pads the trailing tokens
up to a block boundary.  A single helper enforces the check at
every two-stream entry point; violations abort the training step
with an explicit error.

\paragraph{Three kernel-level guards for cross-document correctness.}
Under packing the three two-stream kernels of \S\ref{app:gdn-routes}
and \S\ref{app:short-conv} each carry one document-boundary guard
that the single-document versions do not need.

\begin{itemize}\itemsep2pt
\item \textbf{Noisy recurrence's initial state at document starts}
(Route~II forward kernel of \S\ref{app:route-twostream}; same
boundary also handled by Route~I's fill-in kernel).
The fused two-stream kernel normally seeds the noisy recurrence of
block $b$ from the clean state $\mathbf{S}_{(b-1)B}$
(Eq.~\eqref{eq:noisy-block-fwd}).  When block $b$ is the first
block of a document, $\mathbf{S}_{(b-1)B}$ would be read from the
previous document's tail chunk and carry its clean-prefix
information across the document boundary.  The kernel detects this
case (via the per-program \texttt{is\_doc\_first\_chunk} flag built
from $\mathbf{cu}$) and zeroes the noisy initial state instead.
The forward's noisy block $b{=}1$ of every document therefore
starts from $\mathbf{0}$, matching the single-document semantics
where each document is trained independently.

\item \textbf{Cross-chunk gradient shift at document boundaries}
(Route~I fill-in backward of \S\ref{app:route-refine}; also
analogous code path inside the clean-side cross-chunk scan of
\S\ref{app:kernel-chunk-bwd}).  The normal cross-chunk shift turns
$\partial\mathcal{L}/\partial\mathbf{S}_{[c]}$ (a chunk-boundary
gradient) into the previous chunk's boundary gradient
$\mathbf{dh}_{[c-1]}$ for the scan.  At a document boundary this
shift must not cross: the boundary gradient of a new document's
first chunk has no corresponding end of a preceding chunk \emph{in
the same document} to flow into.  The document-aware shift
implements this by masking the cross-document assignment to zero;
concretely, the contribution $\mathbf{dh}_{[c-1]} \mathrel{+{=}} \mathbf{dh}_{[c]}$
is taken if chunks $c-1$ and $c$ are in the same document, and
zeroed otherwise.  Per-document
chunk counts and offsets are computed once from $\mathbf{cu}$ at
the start of the backward step.

\item \textbf{ShortConv boundary read of the first noisy
block of each document} (Route~II fused ShortConv kernel of
\S\ref{app:short-conv-route-ii}).  The lag-read rule of
Eq.~\eqref{eq:short-conv-noisy} lets a noisy output position read
the preceding clean block when its receptive field straddles the
block boundary.  At a document boundary this read would cross into
the previous document's clean tokens.  The kernel avoids this by
bounds-checking every read with \texttt{src >= 0} and
\texttt{src < doc\_T} after offsetting by the document's \texttt{bos};
reads outside the current document contribute zero.  For Route~I
the analogous behavior is enforced by zeroing the
\texttt{initial\_states} tensor at every document start, and by
masking the corresponding $\mathbf{dh}_0$ gradient in the backward
so that no gradient flows back into the preceding document's
trailing clean tokens.
\end{itemize}

\paragraph{Variable-length kernel support.}
In addition to the three guards above, every kernel in
\S\ref{app:gdn-routes}--\S\ref{app:short-conv} accepts a
\texttt{cu\_seqlens} argument with a compile-time
\texttt{IS\_VARLEN} switch.  When \texttt{cu\_seqlens} is
provided, each Triton program loads its per-document
\texttt{bos}/\texttt{eos} offsets from $\mathbf{cu}$ and uses them
to translate an in-program position into the correct global token
index, replacing the default $i_n \cdot T$ layout.  This is
standard variable-length kernel infrastructure inherited from the
FLA reference implementation and is not \model-specific; packed
and unpacked inputs share the same kernel code path.

\paragraph{Consequence for the cost model.}
The quantitative claims of \S\ref{app:route-comparison} and
\S\ref{app:short-conv-wins} (HBM scaling, backward-grid size,
hand-off count) are stated in terms of a packed sequence length
$L = c_{N_{\text{doc}}}$; document count does not appear because
the document-level guards above add $\mathcal{O}(1)$ work per
document and no additional per-layer tensors.  In particular,
Route~II's strided-checkpoint tensor has size
$N_C \cdot (M/S) \cdot H \cdot d_k \cdot d_v$ (bytes $\times$
precision) regardless of how many documents are packed into the
$N_C$ chunks.

\subsection{End-to-end MFU ablation}
\label{app:mfu-ablation}

\S\ref{sec:flare-kernel} reports the aggregate MFU lift produced
by our kernel stack on \model-2B.  This subsection breaks that
number into its four incremental contributors and reports peak
HBM alongside MFU at four block sizes, so that each MFU gain is
attributable to a specific kernel change.

\paragraph{Setup.}
All runs train \model-2B on a single node of $8{\times}$A100 80~GB
in bf16, using fully-sharded data parallelism (FSDP) and no
tensor-, pipeline-, context-, or expert-parallelism.  Attention
runs under FlexAttention; to fit A100 shared memory at head
dimension $256$, we instruct it to reduce the kernel's pipeline
depth to two stages.  All runs use per-operator activation
checkpointing (the PyTorch selective-AC policy with granularity
``op''), a fused cross-entropy kernel, and the two training-side
variance reduction options of \S\ref{sec:method-training}:
complementary-mask block pairs and antithetic mask sampling.  The
sequence length is $L{=}4096$, the global batch is $8{\times}$ the
local batch ($8$ for the local-batch-$1$ rows, $64$ for the
local-batch-$8$ rows), the
training corpus is the Nemotron post-training v2 SFT mix (see
Appendix~\ref{app:data}), and the GDR chunk size
(\S\ref{app:kernel-single-stream}) is $C{=}64$.  Each row reports
MFU and peak per-GPU HBM as the mean over measurement steps
$11$--$30$ of a $30$-step warm-up-then-measure run.  \emph{Local
batch} is the per-GPU batch $B_{\text{batch}}$; \emph{block size}
is the diffusion-block size $B$ of \S\ref{sec:method-training}.
The AR reference is a plain Qwen3.5-2B run on the same tokens
under a pure-AR single-stream model spec; it uses local batch $1$
because that spec computes cross-entropy over the full logits
tensor (the fused cross-entropy kernel above is only wired into
the \model two-stream specs).

\begin{table}[h]
\centering
\scriptsize
\setlength{\tabcolsep}{3pt}
\caption{\textbf{Incremental MFU ablation for \model-2B on
$8{\times}$A100-80GB (bf16).} Each row adds one kernel-path change
on top of the row above.  \textbf{GDR} is the GDR
kernel; \textbf{ShortConv} is the $1$D causal ShortConv
kernel (\S\ref{app:short-conv}); \textbf{local batch} is the
per-GPU batch; $B$ is the diffusion-block size.  \textbf{Bold}
marks the best cell at each block size.  \textbf{---} in the
$B{=}1$ baseline row marks an out-of-memory run: Route~I's
$L/B$-scaling block-boundary state tensor, multiplied across the
$18$ GDN layers of the backbone, exceeds the
$80$~GiB HBM budget.  The AR reference is a plain Qwen3.5-2B run
on the same tokens.}
\label{tab:app-mfu-ablation}
\begin{tabular*}{\linewidth}{@{\extracolsep{\fill}} l l l c c c c c c c c c @{}}
\toprule
\multirow{2}{*}{Setting} & \multirow{2}{*}{GDR} & \multirow{2}{*}{ShortConv}
& \multirow{2}{*}{\texttt{local batch}}
& \multicolumn{4}{c}{MFU\,(\%) at block size $B$}
& \multicolumn{4}{c}{Peak HBM (GiB) at block size $B$} \\
\cmidrule(lr){5-8} \cmidrule(l){9-12}
& & & & $1$ & $4$ & $8$ & $16$ & $1$ & $4$ & $8$ & $16$ \\
\midrule
AR reference (Qwen3.5-2B)
 & --- & --- & 1
 & \multicolumn{4}{c}{$24.04 \pm 0.06$ (single-stream)}
 & \multicolumn{4}{c}{$20.5$} \\
\midrule
(1) baseline
 & Route~I  & Route~I & 1
 & ---     & $13.80$ & $16.73$ & $22.06$
 & ---     & $18.3$  & $13.8$  & $12.7$ \\
(2) + GDR Route~II
 & Route~II & Route~I & 1
 & $12.56$ & $17.93$ & $18.51$ & $18.36$
 & $12.7$  & $12.7$  & $12.7$  & $12.7$ \\
(3) + local-batch $1\to 8$
 & Route~II & Route~I & 8
 & $14.20$ & $21.40$ & $22.26$ & $22.14$
 & $49.0$  & $45.8$  & $45.9$  & $45.6$ \\
\textbf{(4) + ShortConv Route~II (\model)}
 & Route~II & Route~II & 8
 & $\mathbf{24.20}$ & $\mathbf{24.81}$ & $\mathbf{23.69}$ & $\mathbf{22.55}$
 & $45.0$  & $45.0$  & $45.0$  & $45.0$ \\
\bottomrule
\end{tabular*}
\end{table}

\paragraph{Per-row interpretation.}
\emph{(1) Baseline.}  Both linear-attention sub-components run
the Chunk-then-Refine path of \S\ref{app:route-refine}.  At
$B{=}1$ the row is out of memory; at $B{=}16$ the baseline
reaches $22.06\%$ because the $L/B$-scaling state tensor is
$16{\times}$ smaller and the chunkwise matmul achieves high
tensor-core utilization.
\emph{(2) GDR Route~II.}  Switching the GDR
to the fused two-stream kernel of
\S\ref{app:route-twostream} unblocks $B{=}1$ ($12.56\%$) and
caps peak memory at every block size to $12.7$~GiB, because the
$L/B$ state tensor is no longer materialized in HBM.  At $B{=}4$
the fused path is faster than Route~I and lifts MFU by roughly
$+4$ points (and by $\sim{+}1.8$ at $B{=}8$); at $B{=}16$ it is slower, matching the crossover
of \S\ref{app:route-comparison} where the dense chunk-level
matmul overtakes the fused kernel.
\emph{(3) Raising the per-GPU batch from $1$ to $8$.}  The
$\sim 6$~GiB saved at $B{=}4$ by setting~(2) is spent on a larger
per-GPU batch.  No kernel code changes here; the memory headroom is
converted into tensor-core utilization.  The step is worth
$\sim{+}3.5$--$3.8$ MFU points at $B{\in}\{4,8,16\}$ and pushes peak
HBM to $\sim 46$~GiB, still well within the $80$~GiB budget.
\emph{(4) ShortConv Route~II (\model).}  The
ShortConv fused kernel replaces $L/B$ small per-block
launches by a single fused kernel per sequence.  The gain is
largest at small $B$ ($+10$ points at $B{=}1$), since launch
overhead scales as $L/B$; it tapers off as $B$ grows.  The final
row reaches $24.81\%$ MFU at $B{=}4$, matching and slightly
exceeding the pure-AR reference of $24.04\%$ on identical tokens.  This is the ceiling
one expects for a hybrid-backbone dLLM: \model processes both
the clean and the noisy streams, and the complementary-mask
training of \S\ref{sec:method-training} trains each block under
both the masked and the unmasked partition in a single forward,
so \model performs roughly $4\times$ the per-input-token
attention work of the AR baseline; matching AR MFU therefore
means the kernel stack has absorbed that extra work without
dropping hardware efficiency.

\paragraph{The $B{=}16$ crossover is consistent with the
microbenchmark.}  Reading Tab.~\ref{tab:app-mfu-ablation} at
fixed $B{=}16$, Route~II for the GDR (row~2)
underperforms Route~I (row~1), for the same reason quantified in
Tab.~\ref{tab:route-comparison-gda}: at larger block sizes
Route~I's dense chunk-level matmul overtakes Route~II's smaller
launch footprint.  Setting~(4) still matches or beats setting~(1)
at $B{=}16$ because the ShortConv Route~II gain is
additive and independent of the GDR choice.

\paragraph{Caveats.}
Two properties of the measurement window should be kept in mind
when comparing to single-stream MFU numbers reported elsewhere.
\emph{(i)} The complementary-mask training of
\S\ref{sec:method-training} trains each block under both the
masked and the unmasked partition in a single forward, doubling
the effective batch seen by attention; turning this off would
approximately double all \model MFU numbers in the table but
would no longer reflect the production training regime of the
released \model checkpoints.  \emph{(ii)} The FlexAttention
pipeline-depth reduction above costs roughly $1$--$2\%$ of
attention throughput uniformly across all four settings and the
AR reference, so it is already folded into every row.  Absolute
numbers on H100 or in FP8 will differ; the relative gains
between settings should nonetheless be representative of the
kernel stack's contribution.

\section{Efficient Inference and Implementation}
\label{app:inference}

\subsection{Overview}
\label{app:inference-overview}

\begin{figure}[h]
\centering
\includegraphics[width=\linewidth]{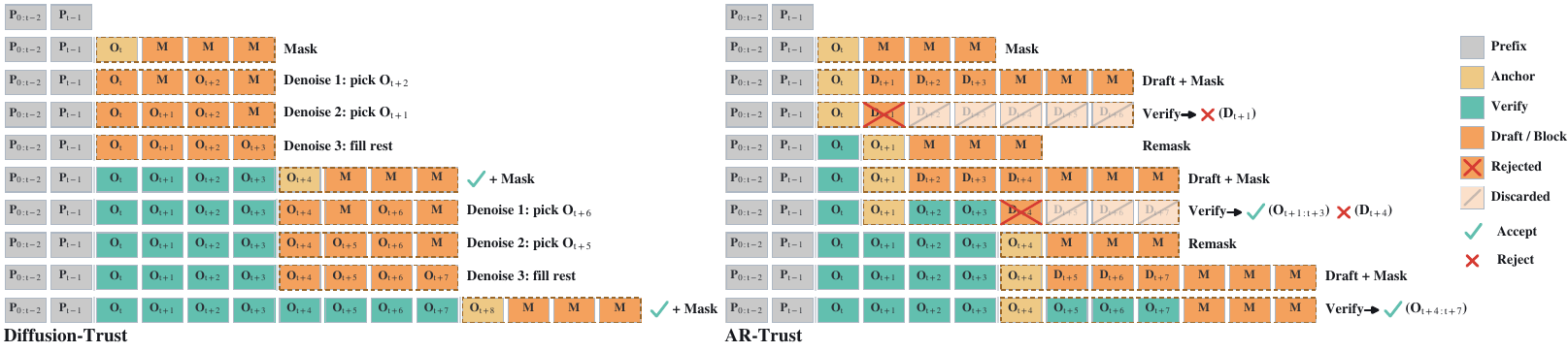}
\caption{Two decoding procedures supported by a single trained \model checkpoint.
\textbf{Left: Diffusion-Trust.} The noisy stream is trusted: a masked block is denoised in parallel over one or more refinement steps and then committed once finalized.
\textbf{Right: AR-Trust.} The clean stream is trusted: noisy-stream logits propose multiple draft tokens, and clean-stream logits verify them left-to-right within the same checkpoint.}
\label{fig:trust-paths}
\end{figure}

Section~\ref{sec:method-inference} defines the two decoding
interfaces used by \model, summarized in Figure~\ref{fig:trust-paths}. This appendix does not redefine those
interfaces; instead, it records the serving-side details needed to
implement them on a hybrid-attention backbone. We first show the
path-specific masks and block layouts, then describe the proposal
policies, fused verification kernels, denoising-loop controls, and
recurrent-state machinery used by our SGLang-based serving
stack~\citep{zheng2024sglang}.

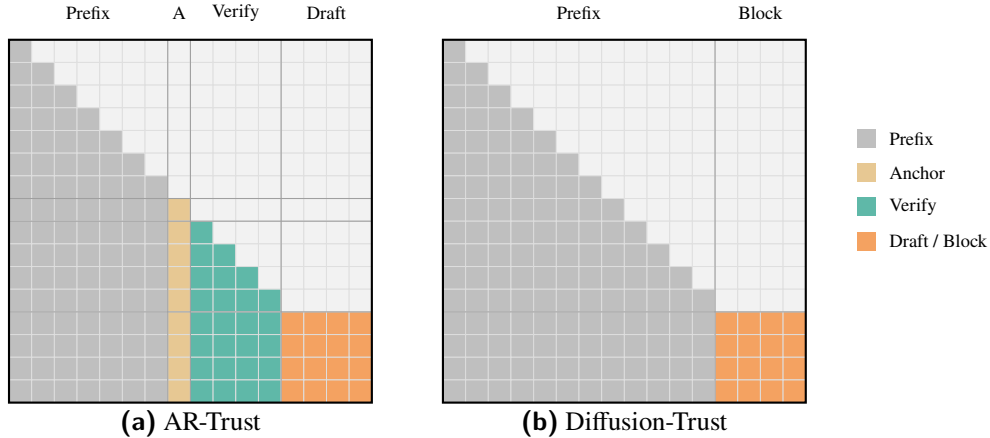
\begin{figure}[h]
\centering
\definecolor{activeCol}{HTML}{F4A261}
\definecolor{ctxCol}{HTML}{BFBFBF}
\definecolor{verifyCol}{HTML}{5FB8A8}
\definecolor{anchorCol}{HTML}{E7C893}
\definecolor{bgCol}{HTML}{F2F2F2}

\begin{tabular}{@{}c@{\hspace{0.055\linewidth}}c@{\hspace{0.035\linewidth}}l@{}}
\begin{tikzpicture}[font=\scriptsize]
  \def\cs{0.30}
  \pgfmathtruncatemacro{\full}{16}
  \foreach \r in {0,...,15} {
    \foreach \c in {0,...,15} {
      \def\fc{bgCol}
      \ifnum\r<7 \ifnum\c<7
        \ifnum\c>\r \else \def\fc{ctxCol} \fi
      \fi\fi
      \ifnum\r>6 \ifnum\r<12
        \ifnum\c<7 \def\fc{ctxCol} \fi
        \ifnum\c>6 \ifnum\c<12
          \ifnum\c>\r \else
            \ifnum\c=7 \def\fc{anchorCol} \else \def\fc{verifyCol} \fi
          \fi
        \fi\fi
      \fi\fi
      \ifnum\r>11
        \ifnum\c<7  \def\fc{ctxCol}    \fi
        \ifnum\c=7  \def\fc{anchorCol} \fi
        \ifnum\c>7  \ifnum\c<12 \def\fc{verifyCol} \fi \fi
        \ifnum\c>11 \def\fc{activeCol} \fi
      \fi
      \fill[\fc] (\c*\cs,-\r*\cs) rectangle ++(\cs,-\cs);
      \draw[gray!25,line width=0.08pt] (\c*\cs,-\r*\cs) rectangle ++(\cs,-\cs);
    }
  }
  \foreach \b in {7,8,12} {
    \draw[gray!75,thin] (\b*\cs,0) -- (\b*\cs,-\full*\cs);
    \draw[gray!75,thin] (0,-\b*\cs) -- (\full*\cs,-\b*\cs);
  }
  \draw[thick] (0,0) rectangle (\full*\cs,-\full*\cs);
  \def\labely{0.36}
  \node at (3.5*\cs, \labely) {Prefix};
  \node at (7.5*\cs, \labely) {A};
  \node at (10.0*\cs, \labely) {Verify};
  \node at (14.0*\cs, \labely) {Draft};
\end{tikzpicture}
&
\begin{tikzpicture}[font=\scriptsize]
  \def\cs{0.30}
  \pgfmathtruncatemacro{\full}{16}
  \foreach \r in {0,...,15} {
    \foreach \c in {0,...,15} {
      \def\fc{bgCol}
      \ifnum\r<12 \ifnum\c<12
        \ifnum\c>\r \else \def\fc{ctxCol} \fi
      \fi\fi
      \ifnum\r>11
        \ifnum\c<12 \def\fc{ctxCol}    \fi
        \ifnum\c>11 \def\fc{activeCol} \fi
      \fi
      \fill[\fc] (\c*\cs,-\r*\cs) rectangle ++(\cs,-\cs);
      \draw[gray!25,line width=0.08pt] (\c*\cs,-\r*\cs) rectangle ++(\cs,-\cs);
    }
  }
  \foreach \b in {12} {
    \draw[gray!75,thin] (\b*\cs,0) -- (\b*\cs,-\full*\cs);
    \draw[gray!75,thin] (0,-\b*\cs) -- (\full*\cs,-\b*\cs);
  }
  \draw[thick] (0,0) rectangle (\full*\cs,-\full*\cs);
  \def\labely{0.36}
  \node at (6.0*\cs, \labely) {Prefix};
  \node at (14.0*\cs, \labely) {Block};
\end{tikzpicture}
&
\raisebox{1.95cm}{
\begin{tikzpicture}[font=\scriptsize]
  \def\sw{0.24}
  \fill[ctxCol] (0,0) rectangle ++(\sw,-\sw);
  \node[right] at (\sw+0.06,-\sw/2) {Prefix};
  \fill[anchorCol] (0,-0.45) rectangle ++(\sw,-\sw);
  \node[right] at (\sw+0.06,-0.45-\sw/2) {Anchor};
  \fill[verifyCol] (0,-0.90) rectangle ++(\sw,-\sw);
  \node[right] at (\sw+0.06,-0.90-\sw/2) {Verify};
  \fill[activeCol] (0,-1.35) rectangle ++(\sw,-\sw);
  \node[right] at (\sw+0.06,-1.35-\sw/2) {Draft / Block};
\end{tikzpicture}}
\\[-0.2em]
\textbf{(a)} AR-Trust
&
\textbf{(b)} Diffusion-Trust
&
\end{tabular}

\caption{Attention patterns presented to the hybrid backbone during decoding.
\textbf{AR-Trust} uses an anchor token and token-causal verify rows for clean-stream verification, followed by bidirectional draft rows whose noisy-stream logits propose the next speculative group.
\textbf{Diffusion-Trust} uses a bidirectional active block behind a causal prefix; active positions attend to the prefix and to one another.
Gray denotes prefix/context, tan denotes the anchor, teal denotes verify rows, and orange denotes bidirectional draft or active-block rows.}
\label{fig:flare-inference-masks}
\end{figure}

\paragraph{Notation.}
We write the prompt as $x_{1:n}$, the generation buffer as
$y \in \mathcal{V}^{L}$ over vocabulary $\mathcal{V}$ (initially
$y_i = \texttt{MASK}$ for $i > n$), the diffusion block size as
$B$, the per-block denoise step budget as $S_{\mathrm{step}}$, and the top-$1$
confidence
$c_i = \max_{v\in\mathcal{V}} p_\theta(y_i = v \mid \cdot)$ under
the checkpoint's output distribution $p_\theta$.  A confidence
threshold $\gamma \in (0,1)$, temperature $T$, and truncation
parameters $(k, p)$ parametrize sampling.  On the AR-Trust side
$N$ is the speculation horizon, $\mathbf{d} = (d_0, \ldots,
d_{N-2})$ are the held draft tokens, and $z$,
$\tilde z$ denote the clean-stream and noisy-stream logits emitted
by one forward over the block (with $z_i$, $\tilde z_i$ their
rows at position $i$); $p_i$, $q_i$ are the target (clean-stream)
and draft (noisy-stream) distributions at the $i$-th verify
position, formed from $z_i$, $\tilde z_i$ under the policy in
use.  We write $\mathcal{B}_b = \{n + bB + 1, \ldots, n + (b{+}1)B\}$
for the position set of the $b$-th generation block.  On the recurrent
side, $\mathbf{S}$ is the GDN state (the same
$\mathbf{S}$ as in Appendix~\ref{app:kernel-routes}),
$L_{\mathrm{gdn}}$ the number of GDN layers in the
backbone, $d_{\mathrm{state}}$ the per-layer state dimension,
$B_{\text{batch}}$ the scheduler batch size, and
$\mathbf{S}^{(r)}$ the state after $r$ accepted updates inside a
verify round.  $\mathcal{K}$ denotes a request's KV cache slot
set; writing a token $y_i$ to $\mathcal{K}$ means appending its
key and value tensors as new entries.  The set of GDN
states indexed by $\mathcal{K}$ is denoted $\mathbf{S}[\mathcal{K}]$.

\subsection{AR-Trust Sampling}
\label{app:inference-ar}

Algorithm~\ref{alg:ar-trust-interface} gives the method-level
interface. Here we specify the serving block layout and proposal
policies used to realize that interface. Figure~\ref{fig:flare-inference-masks}(a)
shows the attention pattern: the active block contains one anchor
position (the last accepted token $y_{\mathrm{anc}}$), $K \le N{-}1$
verify positions that hold the draft tokens to be accepted or
rejected, and $N{-}1$ MASK positions that produce the next round's
drafts. The anchor and verify positions are token-causal to the prefix
and to each other; the draft positions attend bidirectionally over the
whole $2N{-}1$-wide active block and causally to the prefix.

\paragraph{Block layout.}
The active block has length $2N{-}1$ (the maximum-width, steady-state
shape); shorter rounds left-pad with \texttt{MASK}.  In a
\emph{cold-start} round the block is $[t_0,
\texttt{MASK}, \ldots, \texttt{MASK}]$ with $1 + 2(N{-}1) = 2N{-}1$
positions: position $0$ is a fresh clean token and the $2(N{-}1)$
masked positions produce the next round's $N{-}1$ draft tokens
(the first $N{-}1$ MASK positions are unused padding on the cold
start).  In a \emph{verify} round the block is
$[y_{\mathrm{anc}}, d_0, d_1, \ldots, d_{K-1}, \texttt{MASK}, \ldots,
\texttt{MASK}]$ of length $1 + K + (N{-}1)$ where $y_{\mathrm{anc}}$ is the last
accepted token and $d_0, \ldots, d_{K-1}$ are the $K \le N{-}1$
draft tokens held from the previous round; at steady state
$K = N{-}1$ and the block reaches its full width $2N{-}1$.  The
same forward returns clean logits at positions $1, \ldots, K$,
used to accept or reject each draft token, and fresh
clean-plus-draft proposals from the $N{-}1$ masked tail positions.
Algorithm~\ref{alg:flare-ar-trust} expands the compact interface
of Algorithm~\ref{alg:ar-trust-interface} with serving state, KV-cache
bookkeeping, and recurrent-state updates.

\begin{algorithm}[t]
\caption{AR-Trust serving round (implementation details).}
\label{alg:flare-ar-trust}
\small
\begin{algorithmic}[1]
\Require draft policy $\pi \in \{\text{Exact-Truncated}, \text{Softmax-Argmax}, \text{Truncated-Argmax}\}$, $y_{\mathrm{anc}}$, $d_0, \ldots, d_{K-1}$, $q_0, \ldots, q_{K-1}$ (or
$\varnothing$ on cold start); horizon $N$; 

\If{cold start} $y_{\mathrm{blk}} \gets [t_0,
\texttt{MASK}^{2(N-1)}]$ \Comment{length $2N{-}1$; last $N{-}1$ MASKs hold the next-round drafts}
\Else{} $y_{\mathrm{blk}} \gets [y_{\mathrm{anc}}, d_0, \ldots, d_{K-1},
\texttt{MASK}^{N-1}]$ \Comment{length $1 + K + (N{-}1) \le 2N{-}1$}
\EndIf
\State run one forward over $y_{\mathrm{blk}}$ under the hybrid
block mask; obtain clean logits $z$ and noisy logits
$\tilde z$

\State $r \gets 0$; $\mathit{accepted} \gets \varnothing$;
$\mathcal{D}_{\mathrm{tail}} \gets \varnothing$
\For{$i = 0, 1, \ldots, K{-}1$} \Comment{verify pass}
\State form $p_i$ from the clean logit $z_i$; retrieve or
approximate $q_i$ according to policy $\pi$; draw
$u \sim \mathcal{U}(0,1)$
  \State evaluate the modified rejection rule of policy $\pi$
  (Eqs.~\eqref{eq:accept-exact}, \eqref{eq:accept-softargmax},
  \eqref{eq:accept-truncargmax})
  \If{rejected}
    \State set $\mathcal{D}_{\mathrm{tail}}$ to the policy's
    correction distribution at position $i$; \textbf{break}
  \EndIf
  \State $\mathit{accepted} \gets \mathit{accepted} \cup
  \{d_i\}$; $r \gets r + 1$
\EndFor
\If{$r=K$}
  \State set $\mathcal{D}_{\mathrm{tail}}$ to the clean-stream target
  distribution at the first masked position
\EndIf
\State retain the first $r$ new KV entries in $\mathcal{K}$ and
release the remainder; update the recurrent state pool to
$\mathbf{S}^{(r)}$ via the mechanism of
\S\ref{app:inference-machinery}
\State sample $y^\star \sim \mathcal{D}_{\mathrm{tail}}$; draft
$\mathbf{d}' = (d'_0, \ldots, d'_{N-2})$ from
$\tilde z$ at the $N{-}1$ masked positions under policy $\pi$
\State emit $\mathit{accepted} \cup \{y^\star\}$; hold
$\mathbf{d}'$ (and $\mathbf{q}'$ for Exact-Truncated) for the
next round
\end{algorithmic}
\end{algorithm}

\paragraph{Draft policies.}
We consider three draft policies: \textit{Exact-Truncated}, \textit{Softmax-Argmax}, and \textit{Truncated-Argmax}. They differ in how they form the proposal distribution $q_i$ and, consequently, in the modified rejection rule and correction distribution they induce. The key subtlety is proposal consistency. Classical speculative decoding is distribution preserving only when $q_i$ is the conditional law that actually produced $d_i$ under the same proposal factorization assumed by verification. AR-Trust violates this condition if one simply recomputes $q_i$ from current noisy-stream logits, because the noisy stream drafts multiple positions in parallel under a masked block context rather than an autoregressive proposal context. Exact-Truncated therefore stores the proposal probabilities used at draft time, while the two argmax policies intentionally approximate this condition to reduce serving overhead. Empirically, we do not observe significant quality degradation, but we separate these policies to make the exactness--efficiency trade-off explicit.

\noindent\textit{Exact-Truncated.}
Draft tokens are sampled from the same truncated distribution
$q_i$ (top-$k$ and top-$p$ applied to $\tilde z_i / T$) that
verification will consult.  The compact support
$(\mathrm{top}\text{-}k\text{ ids}, \text{probs})$ of $q_i$ is
stored alongside each $d_i$ until the next round.
The corresponding modified rejection-sampling rule is given in Eq.~\eqref{eq:accept-exact}; the stored $q_i$ is the proposal distribution used when $d_i$ was sampled, which is the closest match to the classical speculative-decoding condition.

\noindent\textit{Softmax-Argmax.}
Draft tokens are the argmax of $\tilde z_i$, so $q_i$ is a point
mass at $\arg\max \tilde z_i$.  The ratio $p_i(d_i) / q_i(d_i)$
degenerates and the accept rule reduces to a target-only check,
with $p_i^{\mathrm{full}}$ the full-softmax target over
$z_i / T$:
\begin{equation}
\label{eq:accept-softargmax}
\text{accept } d_i \iff u \leq \min\!\Big(1,\,
p_i^{\mathrm{full}}(d_i)\Big),
\quad y^\star \sim p_i^{\mathrm{full}} \setminus \{d_i\}
\text{ (via Gumbel-max on } z_i/T \text{)}.
\end{equation}
No draft distribution is stored.

\noindent\textit{Truncated-Argmax.}
Draft tokens are the argmax of $\tilde z_i$, but the accept rule
uses the truncated $p_i^{\mathrm{trunc}}$ and $q_i^{\mathrm{trunc}}$
obtained by applying $(k, p)$ truncation to both $z_i$ and
$\tilde z_i$:
\begin{equation}
\label{eq:accept-truncargmax}
\text{accept } d_i \iff u \leq \min\!\Big(1,\,
p_i^{\mathrm{trunc}}(d_i) / q_i^{\mathrm{trunc}}(d_i)\Big),
\quad y^\star \sim \mathrm{normalize}\!\big(\max(p_i^{\mathrm{trunc}}
- q_i^{\mathrm{trunc}},\, 0)\big).
\end{equation}
Because $d_i$ is obtained by argmax rather than by sampling from
$q_i^{\mathrm{trunc}}$, the induced distribution over accepted
tokens does not match the target distribution exactly.  This
policy does not require storing $q_i$ across rounds, and the
verify kernel operates on a $k$-wide support.

\paragraph{Fused full-softmax verify kernel.}
For the Softmax-Argmax policy, verification requires
$p_i^{\mathrm{full}}(d_i)$ on every verify row.  A baseline
implementation launches four or more kernels per row (softmax of
$z_i$, gather of $p_i(d_i)$, the accept check, and, on
rejection, construction and sampling of the correction
distribution), and it materializes an $\mathcal{O}(K V)$
probability tensor for the correction step at vocabularies of
$V \approx 1.5\!\times\!10^5$.  This tensor dominates the
AR-Trust decode step's memory traffic: without fusion, verify
becomes I/O-bound.  We implement the verify step as a single
Triton kernel that does not materialize this tensor.  The kernel
streams over $z_i$ in tiles of $\mathcal{O}(10^3)$ tokens,
maintaining a numerically-stable running $(\max,
\mathrm{sum\text{-}exp})$ pair in the style of
FlashAttention~\citep{dao2023flashattention}, and computes
$p_i^{\mathrm{full}}(d_i)$ directly from the final pair and the
logit at $d_i$.  Accepted rows exit the kernel after the accept
check without performing correction-side work.  Rejected rows
take a second streaming pass and select the correction token by
Gumbel-max over $\log\max(p_i - q_i, 0) + \mathrm{Gumbel}(0, 1)$,
tracking a running argmax across tiles.  Gumbel-max samples the
correction token without normalizing or CDF-inverting the
correction distribution, removing two kernel launches that a
separate CDF-inverse sampler would require.

\paragraph{Fused sparse verify kernel.}
For the Exact-Truncated and Truncated-Argmax policies, $p_i$ and
$q_i$ both have compact supports of size $k$ (the top-$k$ of each
side's truncation).  The sparse verify kernel operates on
$[k, k]$ tiles: loading the $k$ token-id / probability pairs for
target and draft takes $4k$ floats plus $2k$ integer indices per
row.  The correction $\max(p_i - q_i, 0)$ is evaluated on the
$k$ token ids where $p_i$ has mass via a $k\!\times\!k$ lookup
that reads $q_i(v)$ for each $v \in \mathrm{supp}(p_i)$, and inverse-CDF
sampling over $k$ entries selects the correction token.  The
per-row memory traffic is $\mathcal{O}(k)$ instead of
$\mathcal{O}(V)$, which reduces the verify-kernel memory
footprint by roughly three orders of magnitude at $k \sim 50$
and $V \sim 1.5 \times 10^5$; the kernel contains no runtime
branches on the compile-time-known $k$.

\paragraph{Tiled top-$k$ logits over the vocabulary projection.}
The draft policies require, at each masked position, the top-$k$
ids and probabilities of the noisy-stream distribution.  Let
$\mathbf{H} \in \mathbb{R}^{M \times d_{\mathrm{model}}}$ denote
the final-layer hidden states at the $M$ positions for which
logits are needed, and $W_{\mathrm{lm}} \in \mathbb{R}^{V \times
d_{\mathrm{model}}}$ the LM-head projection.  A baseline
implementation materializes the full $\mathbf{H}
W_{\mathrm{lm}}^{\top}$ logits tensor of shape $[M, V]$ and
then applies a top-$k$.  We tile the vocabulary-projection matrix
multiplication: for vocabulary chunks of
$\mathcal{O}(3\!\times\!10^4)$ rows, we compute the partial
$\mathbf{H} W_{\mathrm{lm}}^{\top}\vert_{\text{chunk}}$ matrix,
take its top-$k$ per row, and merge the per-chunk top-$k$ sets
into a running top-$k$ across chunks by concatenation and
re-selection.  Peak auxiliary memory is bounded by the size of
the per-chunk partial matrix multiplication plus the running
top-$k$, and the full $[M, V]$ logits tensor is never
instantiated.  This is a chunked matrix multiplication with
top-$k$ merging rather than a fully fused Triton LM-head kernel;
it avoids the dominant memory term in diffusion decoding and
produces the compact logits that the sparse verify kernel
consumes.

\subsection{Diffusion-Trust Sampling}
\label{app:inference-diffusion}

Algorithm~\ref{alg:diffusion-trust-interface} gives the
method-level denoising interface. Here we describe the serving
controls that make the same interface compatible with the hybrid
state cache. Figure~\ref{fig:flare-inference-masks}(b) shows the
attention pattern used during each denoising forward.

\paragraph{Per-block denoise loop.}
A block begins with a clean seed at its first position followed
by $B{-}1$ masked positions.  The seed is sampled from the final
logit row of the previous block (or from the prompt prefill for
block $0$).  Each denoise iteration forms the unresolved set
$\mathcal{U}^{(s)}$, runs one bidirectional forward over the block,
samples a candidate token at every unresolved position from the
shifted noisy-stream logits under $(T, k, p)$, and commits the
selected set $\mathcal{C}^{(s)}$ according to the threshold schedule
$\gamma_s$.  When all positions are set or the step budget is reached,
a token-causal forward writes $\mathbf{S}$ back to the state pool,
appends the block to $\mathcal{K}$, and produces the final logit row
from which the next block's seed is sampled.
Algorithm~\ref{alg:flare-decode} expands the compact interface
of Algorithm~\ref{alg:diffusion-trust-interface} with seed sampling,
state-pool updates, and per-forward control flags.

\paragraph{Differences between the denoise pass and the state-update pass.}
Each denoise iteration operates on speculative (masked) rows
whose committed values are not yet known, so writing the
resulting recurrent state back to the live pool would contaminate
$\mathbf{S}$ with intermediate speculative updates.  We therefore
process every block in two forwards with different control flow.
(i) The denoise forward uses the bidirectional block mask of
Figure~\ref{fig:flare-inference-masks}(b); the state-update forward
uses a token-causal mask over the block.  (ii) The denoise
forward reads $\mathbf{S}$ from the recurrent pool but does not
write it back; the state-update forward writes the final
per-layer state back to the pool.  (iii) The denoise forward
returns the full vocabulary distribution at every block position
because multiple positions are sampled per iteration, while the
state-update forward returns only the final logit row for seed
sampling.  Our implementation exposes these three axes as flags
on the forward-batch descriptor that the hybrid-attention kernels
read to select mask type, state-update rule, and logits
granularity.

\begin{algorithm}[H]
\caption{Diffusion-Trust serving loop (implementation details).}
\label{alg:flare-decode}
\small
\begin{algorithmic}[1]
\Require prompt $x_{1:n}$, max generation length $L$, block size
$B$, step budget $S_{\mathrm{step}}$, thresholds
$\gamma_1,\ldots,\gamma_{S_{\mathrm{step}}}$, sampling parameters $(T, k, p)$
\State $y_{n+1:n+L} \gets \texttt{MASK}$; prefill $x_{1:n}$ under
a causal mask, write $\mathbf{S}$ back to the state pool, and
sample seed $s_0$ from the final prompt logit
\For{$b = 0, 1, \ldots, \lceil L/B \rceil - 1$}
  \State $\mathcal{B}_b \gets \{\, n + bB + 1, \ldots, n + (b{+}1)B \,\}$
  \State set $y_{n + bB + 1} \gets s_b$
  \For{$s = 1, \ldots, S_{\mathrm{step}}$}
    \State $\mathcal{U}^{(s)} \gets \{i\in\mathcal{B}_b: y_i=\texttt{MASK}\}$
    \If{$\mathcal{U}^{(s)}=\varnothing$} \textbf{break} \EndIf
    \State run a bidirectional forward over $\mathcal{B}_b$
    without writing $\mathbf{S}$ back to the state pool
    \For{each position $i \in \mathcal{U}^{(s)}$}
      \State sample $\hat y_i$ from the shifted noisy-stream logit
      under $(T, k, p)$; record $c_i = p_\theta(\hat y_i \mid \cdot)$
    \EndFor
    \State $\mathcal{C}^{(s)} \gets \{i\in\mathcal{U}^{(s)}:
    c_i\ge\gamma_s\}$ \Comment{$\gamma_{S_{\mathrm{step}}}$ commits all remaining positions}
    \State set $y_i \gets \hat y_i$ for all $i\in\mathcal{C}^{(s)}$;
    $\mathcal{U}^{(s+1)}\gets\mathcal{U}^{(s)}\setminus\mathcal{C}^{(s)}$
  \EndFor
  \State run a token-causal forward over $\mathcal{B}_b$; write
  $\mathbf{S}$ back to the state pool; append the keys and values
  of $\mathcal{B}_b$ to $\mathcal{K}$
  \State sample seed $s_{b+1}$ from the final logit row
  \If{$\texttt{EOS} \in y_{\mathcal{B}_b}$} \textbf{break} \EndIf
\EndFor
\State \Return $y$ up to the first \texttt{EOS} (or the length budget)
\end{algorithmic}
\end{algorithm}

\FloatBarrier

\subsection{Serving-system machinery for the hybrid backbone}
\label{app:inference-machinery}

\begin{figure}[t]
\centering
\definecolor{acceptCol}{HTML}{7FB77E}
\definecolor{rejectCol}{HTML}{E88B8B}
\definecolor{stateCol}{HTML}{D7E7F5}
\definecolor{bufCol}{HTML}{FCD3D6}
\definecolor{liveCol}{HTML}{FFEAA7}
\definecolor{replayCol}{HTML}{F4A261}
\resizebox{0.95\linewidth}{!}{%
\begin{tikzpicture}[font=\scriptsize, >=Latex, x=1.0cm, y=0.7cm]
  \tikzstyle{stepbox}=[draw, rounded corners=1.5pt, minimum width=0.85cm,
                    minimum height=0.55cm, align=center, inner sep=1pt,
                    font=\tiny]
  \tikzstyle{live}=[draw, rounded corners=1.5pt, minimum width=1.1cm,
                    minimum height=0.55cm, align=center, inner sep=1pt,
                    font=\tiny, fill=liveCol]
  \tikzstyle{lab}=[anchor=east, font=\tiny, text=black!75,
                   align=right, text width=1.9cm]

  \node[lab] at (-0.9, 3.2) {verify\\forward};
  \foreach \i/\l in {0/$\mathbf{S}^{(0)}$,1/$\mathbf{S}^{(1)}$,
                     2/$\mathbf{S}^{(2)}$,3/$\mathbf{S}^{(3)}$} {
    \node[stepbox, fill=stateCol] (a\i) at (\i, 3.2) {\l};
  }
  \foreach \i [evaluate=\i as \prev using int(\i-1)] in {1,2,3}
    \draw[->, semithick] (a\prev) -- (a\i);
  \node[font=\tiny, anchor=west, text=black!65] at (3.6, 3.2)
       {$K{=}3$ speculative steps};

  \node[font=\tiny, anchor=south, text=acceptCol!50!black]
       at (1, 3.55) {accept};
  \node[font=\tiny, anchor=south, text=acceptCol!50!black]
       at (2, 3.55) {accept};
  \node[font=\tiny, anchor=south, text=rejectCol!60!black]
       at (3, 3.55) {reject};

  \node[lab] at (-0.9, 1.55) {(a)\\replay};
  \foreach \i/\l in {0/$\mathbf{S}^{(0)}$,1/$\mathbf{S}^{(1)}$,
                     2/$\mathbf{S}^{(2)}$} {
    \node[stepbox, fill=stateCol] (r\i) at (\i, 1.55) {\l};
  }
  \foreach \i [evaluate=\i as \prev using int(\i-1)] in {1,2}
    \draw[->, semithick, replayCol, line width=0.9pt] (r\prev) -- (r\i);
  \node[live] (rlive) at (4.3, 1.55) {$\mathbf{S}[\mathcal{K}]$};
  \draw[->, semithick] (r2.east) -- (rlive.west);
  \node[font=\tiny, anchor=west, text=replayCol!50!black, align=left, text width=3.6cm]
       at (5.2, 1.55)
       {extra per-layer matrix multiply over $r{=}2$ accepted tokens
        ($L_{\text{gdn}} \cdot r$ recurrent updates)};

  \node[lab] at (-0.9, -0.25) {(b)\\cache +\\scatter};
  \node[stepbox, fill=bufCol] (b0) at (0, -0.25) {$\mathbf{S}^{(0)}$};
  \node[stepbox, fill=bufCol] (b1) at (1, -0.25) {$\mathbf{S}^{(1)}$};
  \node[stepbox, fill=bufCol] (b2) at (2, -0.25) {$\mathbf{S}^{(2)}$};
  \node[font=\tiny, anchor=west, text=black!65] at (3.6, -0.25)
       {$K$ verify-position states $\mathbf{S}^{(0)}\!\ldots\mathbf{S}^{(K-1)}$ written as kernel epilogue};
  \node[font=\tiny, anchor=west, text=black!65] at (3.6, -0.75)
       {during the verify forward (draft positions skipped)};

  \node[live] (blive) at (4.3, -2.0) {$\mathbf{S}[\mathcal{K}]$};
  \draw[->, semithick] (b2.south) -- ++(0, -0.5) -| ($(blive.north) + (-0.1, 0)$);

  \node[font=\tiny, anchor=west, text=black!65, align=left, text width=3.6cm]
       at (5.2, -2.0)
       {single fused gather-scatter at offset $r{=}2$ across all
        $L_{\text{gdn}}$ layers; one kernel launch};

  \node[lab] at (-0.9, -2.0) {live\\recurrent\\pool};
\end{tikzpicture}%
}
\caption{Two candidate mechanisms for producing the recurrent
state $\mathbf{S}[\mathcal{K}]$ after a verify round that
accepts $r{=}2$ of $K{=}3$ speculative steps; we adopt
path~\textbf{(b)}.  \textbf{(a) Replay.}
Re-execute the recurrence on the $r$ accepted tokens after the
verify result is known, producing $\mathbf{S}^{(r)}$ at the cost
of an extra per-layer matrix-multiply pass.  \textbf{(b)
Cache-and-scatter (ours).}  The verify forward writes the $K$
verify-position intermediate states into an $[L_{\text{gdn}},
B_{\text{batch}}, K, d_{\mathrm{state}}]$ buffer as the recurrent
kernel's epilogue (one store per verify step, no extra launch),
and a single fused Triton kernel selects the slice at offset $r$
and writes it to the live pool for every GDN layer in
one launch.  Draft-position states are skipped because the accept
offset $r$ can never land on a draft position.}
\label{fig:accepted-state-commit}
\end{figure}
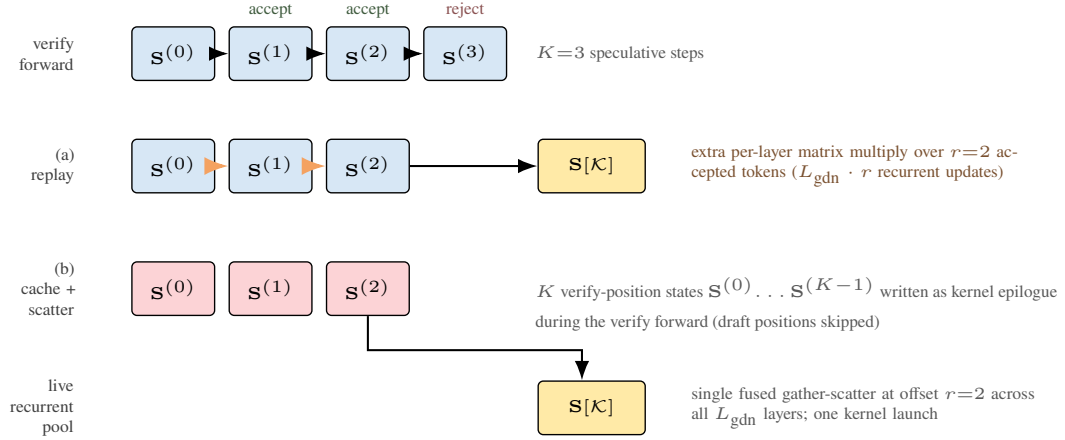

\paragraph{Recovering the recurrent state at the accept boundary.}
On a pure-softmax Transformer backbone, accepting $r$ of $K$
speculative tokens is a KV tail-trim: retain the first $r$ new
entries of $\mathcal{K}$ and release the rest.  The Gated
DeltaNet recurrent state $\mathbf{S}$ has no such per-position
structure: the verify forward advances $\mathbf{S}$ through $K$
rank-one updates, and only the state after the $K$-th update is
retained in the live pool.  Two mechanisms can recover the
required post-accept state $\mathbf{S}^{(r)}$
(Figure~\ref{fig:accepted-state-commit}).  Path~(a) --
\emph{replay} -- re-executes the first $r$ accepted updates
through every GDN layer after the verify decision, at
a cost of $L_{\text{gdn}} \cdot r$ additional recurrent updates
(matrix multiplications).  Path~(b) -- \emph{cache-and-scatter}
-- records $\mathbf{S}^{(t)}$ for every verify position $t \in
\{0, \ldots, K{-}1\}$ during the verify forward itself, at a cost
of $L_{\text{gdn}} \cdot K$ additional stores (memory writes, no
floating-point work).  Draft-position intermediates are not
stored, since the accept offset $r \le K$ can never land on a
draft position.  \textbf{We adopt path~(b)}, because the store
cost is dominated by the recurrent kernel's existing matrix
multiplication and is substantially cheaper than path~(a)'s
extra matrix multiplication.  Concretely, the cache-and-scatter
writes are produced inside the same per-step loop that already
holds $\mathbf{S}$ in registers: at the start of step $t < K$, the
kernel writes the current state $\mathbf{S} = \mathbf{S}^{(t)}$ to a
buffer of shape
$[L_{\text{gdn}}, B_{\text{batch}}, K, d_{\mathrm{state}}]$ and then
performs the step-$t$ update, with no additional kernel launch.  Once
the verify decision is known, a single fused Triton kernel reads
the accepted offset $r$ per request, gathers the corresponding
state slice from the buffer, and writes it to $\mathbf{S}[\mathcal{K}]$
for every GDN layer.  The kernel is launched with one
program per $(\text{request},\ \text{layer},\ \text{state-tile})$
triple, so the $B_{\text{batch}} \cdot L_{\text{gdn}}$ copies
execute in one launch with per-tile parallelism over the state
dimension, and requests with $r = 0$ early-exit without memory
traffic.  A baseline PyTorch implementation
of the same scatter step requires five operations (validity
mask, $\mathrm{nonzero}$ on valid request indices, two
advanced-index selects, and a per-layer scatter), each a separate
launch serialized over the layer dimension; our single-launch
design removes this overhead.

\paragraph{Native dLLM mask modes and the prefix-tile fast path.}
The mask required by both sampling paths is causal over the
prefix, causal for the clean rows of the active block, and
bidirectional for the masked rows.  FlashInfer's generic custom
mask interface expresses this mask as a dense boolean tensor of
size $\mathcal{O}(B(B + n))$ per request.  Our implementation
replaces this encoding and the per-score predicate it triggers
with three mechanisms in FlashInfer's prefill kernel.  First,
two new mask modes accept compact metadata
$(B, \text{row\_type}_{1..B})$ in place of the dense boolean
tensor: one mode assumes a uniform block layout across the batch,
the other admits per-request row-type variation.  Second, the
kernel exploits the following \emph{prefix-tile visibility
invariant}: for any KV tile whose indices lie entirely in the
prefix, every score is unmasked regardless of query row type,
because a causal clean row sees every prefix token and a
bidirectional masked row also sees every prefix token.  Third,
when the kernel detects that a KV tile ends before the prefix
boundary (a single integer comparison per tile), it skips the
per-score mask predicate and takes the causal fast path used by
non-dLLM decode.  The compact encoding plus the prefix-tile
fast path reduce the dLLM mask's per-score overhead to zero on
prefix-only tiles, which constitute the majority of attention
compute at long context.

\paragraph{Graph-replay eligibility under diffusion-LLM invariants.}
Fixed-shape decode blocks are captured into CUDA graphs to
amortize launch overhead.  A standard shape-based eligibility
check is, however, either too strict for the two sampling paths
(disabling graph capture whenever any dLLM flag differs from the
captured state) or too lax (silently replaying the wrong graph
under a mask-mode or logits-mode mismatch); the latter case
corrupts AR-Trust verification.  Our replay-eligibility
predicate matches four invariants at replay time: (i) block
size, (ii) mask-mode metadata, (iii) recurrent-state update
mode, and (iv) logits-output mode.  The draft policies of
\S\ref{app:inference-ar} differ on invariant (iv):
Softmax-Argmax consumes dense logits whereas Exact-Truncated and
Truncated-Argmax consume top-$k$ logits.  We therefore capture
one graph per draft policy and dispatch through the
four-invariant predicate.  This preserves graph-capture speedups
on the steady-state decode loop and eliminates the silent
correctness failure mode.

\section{Transfer-data mixes: sources and curation pipeline}
\label{app:data}

This appendix documents every public dataset used to build the four
transfer-data mixes of \S\ref{sec:study}
(\texttt{Long-CoT}, \texttt{Short-CoT+Math}, \texttt{Long-CoT+Math},
\texttt{Long-CoT+Math+IF}), the common curation pipeline we apply to
those sources, and the per-mix filtering and weighting decisions.
\S\ref{app:data-sources} catalogs the source datasets with their
size, sample format, prompt origin, and generator.
\S\ref{app:data-pipeline} describes the shared preprocessing steps
(message-schema unification, reasoning-trace handling, token
estimation, filtering).  \S\ref{app:data-mixes} walks through each
of the four paper-facing mixes and shows which sources it is built
from and how its components are weighted at training time.

\subsection{Source datasets}
\label{app:data-sources}

All sources are public SFT corpora with a shared
\texttt{[\{role, content\}]} chat schema (some carry an auxiliary
\texttt{reasoning\_content} field that we merge into the assistant
message, see \S\ref{app:data-pipeline}).  Each paragraph below gives
a one-sentence summary of what the dataset contains, the sample
count, the prompt source, and the generator that produced the
assistant responses; the Hugging Face URL is in a footnote.  Sample
counts, storage sizes, and token statistics in this subsection are
as reported on the dataset cards at the time of download.

\paragraph{Llama-Nemotron-Post-Training-Dataset (SFT splits)\protect\footnote{\url{https://huggingface.co/datasets/nvidia/Llama-Nemotron-Post-Training-Dataset}}.}
A post-training SFT mix of long-chain-of-thought traces across four
splits, approximately $4.2$M samples total (math, $2.22$M; code,
$953$K; science, $708$K; chat, $349$K).  Each sample carries a
\texttt{reasoning} field set to \texttt{on} or \texttt{off};
\texttt{on} samples contain an assistant response with an explicit
\texttt{<think>\ldots</think>} trace baked into the content.
Prompts are drawn from NVIDIA's curated post-training prompt pool
spanning math competitions, coding benchmarks, science
question-answering, and chat.  Assistant responses are generated by
the Llama-Nemotron synthesis pipeline (primarily DeepSeek-R1 and
variants).  Storage: $114$~GB.

\paragraph{Nemotron-Post-Training-Dataset-v2\protect\footnote{\url{https://huggingface.co/datasets/nvidia/Nemotron-Post-Training-Dataset-v2}}.}
A general-purpose post-training SFT corpus spanning four English
categories (stem, chat, math, code) and five multilingual categories
(Japanese, German, French, Spanish, Italian); only the English
splits ($1{,}397{,}187$ samples) are used in this paper.  Average
CoT length is shorter than Llama-Nemotron's long-CoT traces (roughly
$800$--$1{,}350$ tokens per assistant message, depending on
category).  Assistant responses have \texttt{<think>} reasoning
baked into \texttt{content} directly rather than carried in a
separate field.  Synthetic data is generated with Qwen3 and
DeepSeek-R1-0528.  Storage: $92$~GB.

\paragraph{Nemotron-Instruction-Following-Chat-v1\protect\footnote{\url{https://huggingface.co/datasets/nvidia/Nemotron-Instruction-Following-Chat-v1}}.}
Instruction-following chat conversations ($320$K samples) with an
explicit \texttt{capability\_target} label that distinguishes
instruction-following from structured-output samples.  Average
assistant message is around $3.9$K tokens.  Reasoning is stored in
a separate \texttt{reasoning\_content} field.  Prompts come from
NVIDIA's instruction-following prompt pool; responses are generated
with DeepSeek-R1-family models.  Storage: $6.4$~GB.

\paragraph{Nemotron-SFT-Instruction-Following-Chat-v2\protect\footnote{\url{https://huggingface.co/datasets/nvidia/Nemotron-SFT-Instruction-Following-Chat-v2}}.}
The v2 refresh of the instruction-following chat set, with broader
prompt coverage and $1.99$M samples.  Average assistant message is
around $1.2$K tokens (shorter than v1).  Same
\texttt{reasoning\_content} schema as v1.  Storage: $14.4$~GB.

\paragraph{Nemotron-Science-v1\protect\footnote{\url{https://huggingface.co/datasets/nvidia/Nemotron-Science-v1}}.}
Science / STEM reasoning conversations ($226$K samples, average
assistant message around $2.5$K tokens).  Topics include physics,
chemistry, biology, and STEM-style problem solving.  Reasoning is
stored in \texttt{reasoning\_content}.  Generators include
DeepSeek-R1 and Qwen3-family models.  Storage: $2.3$~GB.

\paragraph{Nemotron-Math-Proofs-v1\protect\footnote{\url{https://huggingface.co/datasets/nvidia/Nemotron-Math-Proofs-v1}}.}
Math-proof conversations ($925$K samples, average assistant message
around $4.3$K tokens) emphasising symbolic derivations and
proof-style chain-of-thought.  Reasoning is stored in
\texttt{reasoning\_content}.  Storage: $27$~GB.

\paragraph{Nemotron-SFT-Competitive-Programming-v2\protect\footnote{\url{https://huggingface.co/datasets/nvidia/Nemotron-SFT-Competitive-Programming-v2}}.}
Competitive programming problem/solution pairs.  The schema varies
across its four splits: an \texttt{exercism} split ($79$K samples)
with no reasoning traces; a \texttt{text\_to\_sql} split ($97$K);
and two competitive-coding splits in C++ ($333$K) and Python
($337$K), both with \texttt{reasoning\_content} traces averaging
$30$K--$56$K tokens.  Weighted total: $845$K samples.  Prompt
sources span Exercism exercises, text-to-SQL benchmarks, and
CodeForces-style competitive problems.  Responses are generated by
DeepSeek-R1 variants.  Storage: $90$~GB.

\paragraph{Nemotron-Cascade-1-SFT-Data (Stage-2 instruction-following only)\protect\footnote{\url{https://huggingface.co/datasets/nvidia/Nemotron-Cascade-1-SFT-Data}}.}
NVIDIA's first-generation ``cascade'' SFT dataset; we only use its
instruction-following subset ($146$K samples).  Prompts are taken
from the Tulu-3 SFT
mixture\footnote{\url{https://huggingface.co/datasets/allenai/tulu-3-sft-mixture}};
assistant responses are generated by DeepSeek-R1-0528.

\paragraph{Nemotron-Cascade-2-SFT-Data (instruction-following subset)\protect\footnote{\url{https://huggingface.co/datasets/nvidia/Nemotron-Cascade-2-SFT-Data}}.}
NVIDIA's second-generation cascade SFT dataset.  We use its
instruction-following split after applying NVIDIA's own
constraint-satisfaction filter, which selects $362$K samples from a
raw pool of about $820$K.  Assistant responses are generated by the
Cascade-2 pipeline (GPT-OSS-120B).

\subsection{Curation pipeline}
\label{app:data-pipeline}

Every source goes through the same preprocessing pipeline before it
is combined into any of the four transfer mixes.  The pipeline has
four stages, implemented in our data-preparation scripts and applied
uniformly across sources.

\paragraph{(1) Message-schema unification.}
All samples are rewritten into a single \texttt{[\{role, content\}]}
chat schema.  Three conversion variants handle the three schema
families we encounter in the sources above:
\begin{itemize}\itemsep1pt
\item Sources with a separate \texttt{reasoning\_content} field
(Chat-v1, Chat-v2, Science-v1, Math-Proofs-v1,
Competitive-Programming-v2) have it merged into the assistant
\texttt{content} wrapped in a \texttt{<think>\{reasoning\}</think>}
block.
\item Llama-Nemotron's \texttt{input}/\texttt{output} schema is
flattened into a messages list; the \texttt{<think>} trace is
already baked into the output string, so no reasoning merge is
needed.
\item Post-Training-v2 already ships with \texttt{messages} and
\texttt{<think>} tags inline; we only filter to the English
categories and pass the rest through unchanged.
\end{itemize}
Samples without any reasoning trace are padded with an empty
\texttt{<think></think>} block so that downstream prompt-template
handling is uniform across sources.  The goal of this unification is
that a plain string-concatenation of the resulting
\texttt{[\{role, content\}]} messages under Qwen3's ChatML template
yields the same token sequence that a full Qwen3/Qwen3.5
\texttt{apply\_chat\_template} call would produce on the upstream
raw sample.  This lets our data pipeline avoid instantiating a
tokenizer: every source is rewritten once into the canonical
messages form, and the trainer handles tokenisation.

\paragraph{(2) Category and source tagging.}
Each sample is assigned a \texttt{category} label (one of \emph{math},
\emph{code}, \emph{stem}, \emph{chat}) and a \texttt{source} tag
identifying the upstream dataset.  These labels are preserved through
all subsequent stages and let us compute per-category statistics and
apply category-specific filtering.

\paragraph{(3) Token-count estimation.}
Because the combined source pool has tens of millions of samples,
tokenizing the full corpus with the Qwen3-1.7B tokenizer is
prohibitively slow.  We instead estimate per-sample token counts from
character counts plus a per-source char-to-token ratio: we tokenize
$3{,}000$ randomly-sampled messages per source to compute that
source's median character-to-token ratio, and divide the per-sample
character count by this ratio (with a fixed $15$-token template
overhead).  Empirical spot checks against full tokenisation put the
estimation error under $5\%$.  We record both a total-token and an
assistant-token estimate for each sample.

\paragraph{(4) Filters applied uniformly.}
Two filters run on every mix regardless of its target length or
domain.  First, samples whose assistant message contains an
\texttt{<think>} tag without a matching \texttt{</think>} (indicating
a truncated reasoning trace in the upstream source) are dropped;
this removes approximately $45$K samples from the \texttt{Long-CoT}
source pool and $6$K from the \texttt{Math} source pool.  Second,
when multiple sources are concatenated, we deduplicate across the
combined pool using a hash of the serialized message list; this
removes approximately $11$K duplicates when the \texttt{IF} sources
are merged.

\paragraph{Output format.}
After preprocessing each source yields a curated table where every
row carries: the serialized messages list, the category and source
tags, the estimated total and assistant token counts, and the
sample's origin metadata.  The four transfer mixes of
\S\ref{app:data-mixes} are built by selecting, filtering, and
concatenating rows from these curated source tables.

\subsection{The four transfer-data mixes}
\label{app:data-mixes}

The four mixes combine the curated sources of
\S\ref{app:data-sources} under per-mix filtering and weighting
rules.  Each mix is trained at the same $4{,}096$-token packed
context with global batch size $256$ for $9{,}000$ optimiser steps
on the Qwen3-1.7B seed (\S\ref{sec:study}).

\subsubsection{Long-CoT}
\label{app:data-longcot}

\texttt{Long-CoT} targets the high-reasoning-depth regime by
keeping only samples whose estimated total token count is at least
$3{,}000$.  It draws from seven sources of
\S\ref{app:data-sources}: Llama-Nemotron-Post-Training (all four
SFT splits), Post-Training-v2 (English splits only),
Instruction-Following-Chat-v1, Instruction-Following-Chat-v2,
Science-v1, Math-Proofs-v1, and Competitive-Programming-v2.  The
construction proceeds as follows:
\begin{enumerate}\itemsep1pt
\item Apply the common preprocessing pipeline
(\S\ref{app:data-pipeline}) to each source.
\item For efficiency, apply a cheap character-count prefilter at
twice the target token threshold (since the character/token ratio
is bounded below by approximately $2$).
\item Apply the final filter using the calibrated total-token
estimate, keeping samples with at least $3{,}000$ tokens.
\item Concatenate the filtered per-source tables and shuffle with
a fixed seed.
\end{enumerate}
Final stats: $4{,}588{,}156$ samples total, broken down as math
$2.55$M, code $1.23$M, chat $635$K, stem $179$K.  Storage: roughly
$186$~GB.  At the $4{,}096$-token packing cap used during training,
this corresponds to approximately $18.3$B total tokens and
$16.8$B assistant (supervised) tokens.

\subsubsection{Short-CoT+Math}
\label{app:data-shortcot-math}

\texttt{Short-CoT+Math} pairs a short-reasoning general mix with a
dedicated \texttt{Math} source pool, giving a contrast point to
\texttt{Long-CoT}.  The short-reasoning source is Post-Training-v2
(English splits) used without any length filter; average assistant
traces in Post-Training-v2 are roughly $1$K tokens, an order of
magnitude shorter than Llama-Nemotron's long-CoT output.  The
\texttt{Math} source pool is built from the math split of
Llama-Nemotron (restricted to \texttt{reasoning=on}, roughly
$2.22$M samples) concatenated with all of Math-Proofs-v1 (roughly
$920$K samples after the unclosed-\texttt{<think>} filter drops
$6$K).  No length filter is applied to either component.  The two
components are passed to the trainer with equal weights
(see Table~\ref{tab:app-data-weights}).  The \texttt{Math} source
pool alone contains $3{,}140{,}111$ samples; at the $4{,}096$-token
packing cap it contributes approximately $11.4$B total tokens.

\subsubsection{Long-CoT+Math}
\label{app:data-longcot-math}

\texttt{Long-CoT+Math} is a direct intersection of the previous two
mixes: the \texttt{Long-CoT} source pool from
\S\ref{app:data-longcot} (all $4.6$M length-filtered samples) and
the \texttt{Math} source pool from \S\ref{app:data-shortcot-math}
(Llama-Nemotron math plus Math-Proofs-v1, $3.1$M samples total).
No length filter is applied to the \texttt{Math} source pool in
this mix; the two pools are passed to the trainer with equal
weights.  This isolates the effect of adding a math-only anchor on
top of the \texttt{Long-CoT} source pool without changing the
short-CoT baseline.

\subsubsection{Long-CoT+Math+IF}
\label{app:data-longcot-math-if}

\texttt{Long-CoT+Math+IF} adds a dedicated instruction-following
\texttt{IF} source pool to the previous mix.  The \texttt{IF}
source pool is built from four sources and deduplicated across the
merged pool:
\begin{enumerate}\itemsep1pt
\item Nemotron-Cascade-1-SFT-Data Stage-2 IF ($146$K samples),
which uses Tulu-3 SFT prompts and DeepSeek-R1-0528 responses.
\item Nemotron-Cascade-2-SFT-Data IF ($362$K samples, selected by
NVIDIA's constraint filter from a raw pool of roughly $820$K);
responses are generated by GPT-OSS-120B.
\item Instruction-Following-Chat-v1 samples tagged with the
\texttt{instruction\_following} capability target ($78$K), giving
a direct IF-labeled subset of Chat-v1.
\item Instruction-Following-Chat-v1 structured-outputs samples
($5$K), which ask the model to generate JSON or XML conforming to
a given schema.
\end{enumerate}
After preprocessing and deduplicating across the four parts on a
hash of the serialized message list (which removes about $11$K
duplicates), the \texttt{IF} source pool contains $591{,}718$
samples, approximately $0.66$B total tokens and $0.53$B assistant
tokens.  It is trained alongside the \texttt{Long-CoT} and
\texttt{Math} source pools with the weights of
Table~\ref{tab:app-data-weights}.

\subsection{Mixing weights, sampling, and packing}
\label{app:data-mixing-weights}

Table~\ref{tab:app-data-weights} lists the trainer-level mixing
weights used for the four mixes.  The rest of this subsection
describes how those weights are applied during data loading and how
samples are packed into fixed-length training sequences.

\paragraph{Source-local packing.}
Each source dataset in the mix is wrapped in its own SFT
data-loading iterator that produces \emph{already-packed}
$4{,}096$-token training sequences.  Raw samples from that source
are concatenated into a running buffer via greedy bin-packing: for
each raw sample we apply Qwen3's ChatML template to produce a pair
$(\text{tokens}, \text{labels})$ with $-100$ on user and system
positions (so only assistant tokens contribute to the loss); a
sample whose tokenized length exceeds $4{,}096$ is truncated, while
a sample that would merely overflow the current buffer is deferred
(the current buffer is flushed with padding up to $4{,}096$, and
the deferred sample starts the next packed sequence).  Each packed
sequence carries per-token document IDs; the trainer uses these
with a document-causal attention mask to prevent cross-document
attention within a packed sequence.  For the AR training variant,
the first label of each post-boundary document in a packed sequence
is additionally set to $-100$, so the shifted-label AR loss never
asks the last token of one document to predict the first token of
the next.

\paragraph{Per-sequence multinomial mixing.}
Because packing is source-local, the unit of draw from the mixture
is one fully-packed $4{,}096$-token sequence.  On each
\texttt{next} call, a mixer component is selected from the
multinomial distribution $(w_1, \ldots, w_k)$ of the current mix,
and that component's next packed sequence is yielded.  The
multinomial draw is with replacement over components, while each
component iterates its own packed sequences without replacement
within its own epoch; the realized per-batch composition therefore
matches $(w_1, \ldots, w_k)$ only in expectation.  If a component
is exhausted mid-epoch its weight is zeroed and the remaining
weights are re-normalized.  One consequence of this design matters
for interpretation: raw samples from different sources never
co-occur in the same packed sequence, so no cross-source attention
can happen even under the document-causal mask.  The sampling RNG
is seeded deterministically and persisted in the training
checkpoint, so a resumed run produces the same sample stream as an
uninterrupted run.

\begin{table}[h]
\centering
\small
\setlength{\tabcolsep}{4pt}
\caption{Mixing weights for the four transfer mixes.  Each row is
one paper-facing mix; the Components column lists, for each
component, the source pool and the sampling weight passed to the
trainer.  All four mixes are trained with a 4096-token context,
global batch size 256, and 9000 optimiser steps on the
Qwen3-1.7B seed.}
\label{tab:app-data-weights}
\begin{tabular}{ll}
\toprule
Mix & Components (source pool $\to$ weight) \\
\midrule
\texttt{Long-CoT}         & \texttt{Long-CoT} pool (\S\ref{app:data-longcot}) $\to 1.0$ \\
\texttt{Short-CoT+Math}   & Post-Training-v2 (English) $\to 0.5$; \texttt{Math} pool (\S\ref{app:data-shortcot-math}) $\to 0.5$ \\
\texttt{Long-CoT+Math}    & \texttt{Long-CoT} pool $\to 0.5$; \texttt{Math} pool $\to 0.5$ \\
\texttt{Long-CoT+Math+IF} & \texttt{Long-CoT} pool $\to 0.4$; \texttt{Math} pool $\to 0.4$; \texttt{IF} pool (\S\ref{app:data-longcot-math-if}) $\to 0.2$ \\
\bottomrule
\end{tabular}
\end{table}

\subsection{Automatic SFT data selection for AR-to-dLLM transfer}
\label{app:data-selection}

The four transfer mixes of \S\ref{app:data-mixes} select data
primarily by length and per-source weighting.  An orthogonal
question is whether automatic, instance-level quality selection on
the same raw SFT pool improves AR-to-dLLM transfer quality.  This
subsection documents such a pipeline: it scores individual
examples with an instruction-following-difficulty signal, applies
per-source quality gates, and rebalances the result by domain via
cluster-aware sampling, all without manual curation.  The
resulting mixture (which we call the \texttt{Auto-selected} mix
below) is not used by any of the paper-facing \model{} runs, but
its controlled comparison against an unfiltered baseline (under
the same Qwen3-1.7B seed and training budget) informed our final
decision to keep the \texttt{Long-CoT+Math+IF} mix lightly
filtered.

\paragraph{Source pools.}
The automatic selection pipeline starts from two public SFT pools that
together approximate the long-CoT plus instruction-following
coverage of \S\ref{app:data-mixes}: \texttt{Dolci-Think-SFT-32B}
($2.25$M raw rows, $23.58$B raw tokens; $1.17$M rows and $3.02$B
tokens at $\leq 8$K context) and \texttt{Nemotron-Cascade-2-SFT-Data}
($24.49$M rows, $162.94$B tokens; $19.06$M rows and $54.04$B tokens
at $\leq 8$K context).  Together the union has $26.74$M rows and
$186.52$B tokens raw, dropping to $20.23$M rows and approximately
$57$B tokens after the $8{,}192$-token Qwen3 ChatML length filter.
Cascade-2 overlaps in part with the
Nemotron-Cascade-2-SFT-Data instruction-following subset already
described in \S\ref{app:data-sources}; here we use the un-filtered
$\leq 8$K pool rather than NVIDIA's constraint-satisfaction
selection.

\paragraph{Six-step curation pipeline.}
The pipeline applies the following six steps in order.  Each step
only sees examples that survived the previous one.
\begin{enumerate}\itemsep1pt
\item \textbf{Structural validity.}  Drop malformed conversations:
empty \texttt{messages}, missing roles, zero-length content,
unmatched \texttt{<think>} tags, assistant turns that repeat the
prompt verbatim, and dangling final user turns.  Known
Cascade-2 tool-message leaks are normalized in place.
\item \textbf{Length filter.}  Tokenize with the Qwen3 ChatML
template and drop examples with more than $8{,}192$ tokens.  Long
reasoning traces are dropped rather than truncated.
\item \textbf{Deduplication and evaluation decontamination.}  Remove
near-duplicates with MinHash LSH over $5$-gram prompt shingles at
Jaccard $\geq 0.8$ for intra-source dedup, and drop rows with
Jaccard $\geq 0.5$ against the prompts of GSM8K, MATH-500, IFEval,
HumanEval, MBPP, ARC-C, and GPQA.  This step follows standard
large-corpus deduplication and decontamination practice.
\item \textbf{Quality scoring.}  Score each instruction-response
pair $(x, y)$ with the Instruction-Following Difficulty (IFD)
score, defined as $\mathrm{IFD}(x, y) = \mathrm{PPL}(y \mid x) /
\mathrm{PPL}(y)$, where $\mathrm{PPL}(y \mid x)$ is the response
perplexity under the scorer when conditioned on the prompt and
$\mathrm{PPL}(y)$ is the unconditional response perplexity.  Lower
IFD indicates that conditioning on the instruction substantially
reduces response perplexity, which we interpret as stronger
instruction--response alignment.  We use Qwen3-1.7B-Base as a cheap
weak scorer.  We additionally record rank-normalized IFD, response
PPL, and response length, drop extreme response-PPL outliers, and
do not score very short responses.
\item \textbf{Composite ranking and source-specific gating.}  Rank
rows within each source by a composite of normalized IFD,
length, and a diversity weight: $0.5 \cdot (1 - \mathrm{IFD}_{\text{norm}}) +
0.3 \cdot \mathrm{length}_z + 0.2 \cdot \mathrm{diversity}_{\text{wt}}$.
Apply source-specific keep rates (top $70\%$ for cleaner
math/science/IF sources, top $50\%$ for Dolci Python algorithms,
and top $40$--$60\%$ for noisier chat/tool sources).
\item \textbf{Cluster-balanced sampling.}  Embed instruction text
with \texttt{BAAI/bge-small-en-v1.5}, run $k$-means within each
domain bucket, and round-robin sample from clusters by composite
score until each bucket budget is met.  Each curated row carries
a \texttt{cluster\_id} and an inverse cluster-frequency weight.
\end{enumerate}

\paragraph{Target and realized domain mixture.}
The bucket budgets were chosen to be evaluation-driven rather than
proportional to raw source size, biased toward math, code,
science, and instruction following.  Table~\ref{tab:app-selection-mix}
lists the planned target shares and the realized bucket counts
after sampling.  The final mixture contains $1{,}627{,}706$ rows
and approximately $5$B tokens, against an original design target
of roughly $6$B curated tokens for three epochs ($\sim 18$B
effective training tokens).

\begin{table}[h]
\centering
\small
\setlength{\tabcolsep}{4pt}
\caption{Auto-selected SFT mixture: target domain shares
(eval-driven design) and realized bucket counts after the six-step
pipeline.
Targets are the budget passed to the cluster-balanced sampler;
realized counts are the number of unique prompts in the curated
split.}
\label{tab:app-selection-mix}
\begin{tabular}{lrrr}
\toprule
Bucket & Target share & Prompts & Realized share \\
\midrule
\texttt{math\_direct}      & $22\%$ & $252{,}060$ & $15.49\%$ \\
\texttt{math\_tool}        & $8\%$  & $85{,}120$  & $5.23\%$  \\
\texttt{code}              & $18\%$ & $100{,}328$ & $6.16\%$  \\
\texttt{precise\_if}       & $12\%$ & $476{,}567$ & $29.28\%$ \\
\texttt{science}           & $12\%$ & $335{,}153$ & $20.59\%$ \\
\texttt{chat}              & $15\%$ & $167{,}156$ & $10.27\%$ \\
\texttt{tool\_multi\_turn} & $8\%$  & $85{,}308$  & $5.24\%$  \\
\texttt{safety}            & $3\%$  & $66{,}368$  & $4.08\%$  \\
\texttt{multilingual}      & $2\%$  & $59{,}646$  & $3.66\%$  \\
\midrule
Total                      & $100\%$ & $1{,}627{,}706$ & $100\%$ \\
\bottomrule
\end{tabular}
\end{table}

The deviation between target and realized share, especially the
overshoot on \texttt{precise\_if} and \texttt{science} and the
undershoot on \texttt{math\_direct} and \texttt{code}, reflects how
many high-IFD examples each upstream source actually contributed
once the cluster-balanced sampler attempted to fill the bucket
budgets without collapsing diversity.

\paragraph{Controlled comparison with an unfiltered baseline.}
To assess whether the additional curation effort translated into
downstream quality, we ran a head-to-head SFT comparison on
Qwen3-1.7B at the same training budget against an unfiltered
baseline mixture used elsewhere in the project (referred to here
as the ``baseline'' mix).  Both mixes were trained on the same
seed checkpoint with the same optimizer settings and evaluated
under the same $16$K-token evaluation harness.
Table~\ref{tab:app-selection-eval} reports the per-benchmark
scores together with the parameter-matched Qwen3-1.7B reference
evaluated at $8$K context.

\begin{table}[h]
\centering
\small
\setlength{\tabcolsep}{4pt}
\caption{SFT comparison between the unfiltered baseline mixture
and the auto-selected mixture under the same Qwen3-1.7B seed,
training budget, and $16$K evaluation harness.  The first row is
the original Qwen3-1.7B checkpoint evaluated at $8$K context as a
reference.  Bold marks the better of the two SFT runs on each
benchmark.}
\label{tab:app-selection-eval}
\begin{tabular}{lcccccc}
\toprule
Run & GSM8K & MATH-500 & IFEval & HumanEval & MBPP & ARC-C \\
\midrule
Qwen3-1.7B reference ($8$K eval)   & $88.93$ & $85.80$ & $71.72$ & $81.71$ & $75.49$ & $89.16$ \\
Baseline SFT, $16$K eval           & $82.79$ & $\mathbf{84.40}$ & $\mathbf{67.10}$ & $66.46$ & $58.37$ & $82.51$ \\
Auto-selected SFT, $16$K eval      & $\mathbf{85.14}$ & $79.80$ & $63.59$ & $\mathbf{70.12}$ & $\mathbf{66.54}$ & $\mathbf{85.58}$ \\
\midrule
$\Delta$ Auto-selected $-$ baseline & $+2.35$ & $-4.60$ & $-3.51$ & $+3.66$ & $+8.17$ & $+3.07$ \\
\bottomrule
\end{tabular}
\end{table}

Under this evaluation setting, the auto-selected mixture improves
GSM8K, HumanEval, MBPP, and ARC-C relative to the unfiltered
baseline mixture, while regressing on MATH-500 and IFEval.  The
pattern suggests that the IFD-driven composite ranking and
cluster-balanced sampler effectively concentrate budget on cleaner
code and science instruction-following data, but that the realized
bucket undershoot on \texttt{math\_direct}
(Table~\ref{tab:app-selection-mix}) translates into a measurable
MATH-500 regression, and the heavy reweighting away from generic
chat hurts IFEval despite the high \texttt{precise\_if} share.

We did not pursue this automatic selection pipeline further for
the paper-facing \model{} runs.  Both SFT runs in
Table~\ref{tab:app-selection-eval} fall well short of the reference
Qwen3-1.7B at $8$K, indicating that under our fixed training
budget the bottleneck is not a shortage of high-IFD examples but
the same source-distribution shift discussed in \S\ref{sec:flare}:
continuing SFT on external long-CoT data drawn from non-Qwen
generators shifts the output distribution away from the post-trained
seed.  The \model{} \texttt{Long-CoT+Math+IF} mix in
\S\ref{app:data-mixes} therefore keeps filtering minimal and
relies on length-based selection plus per-source mixture weighting
rather than the heavier composite-ranking approach explored here.

\section{Additional Experimental Details and Results}
\label{app:per-benchmark}

In this section, we present training and inference details for \model, as well as additional experimental results for the reader's reference. 

\subsection{Inference and Training Hyperparameters}
\paragraph{Training} For all experiments in this paper, we use an AdamW optimizer with a learning rate of $10^{-5}$, a 500-step warm-up, and no learning rate decay. 
We always use a global batch size of 256 and a maximum sequence length of 4096 during training for 9000 optimizer steps.
 For Qwen3-related experiments, we use 32 NVIDIA RTX A100 80GB for experiments. For Qwen3.5-related experiments, we use 32 NVIDIA RTX H100 80GB for experiments. Each experiment requires a wall-clock time of 12 to 36 hours. 
 
\paragraph{Inference} We use SGLang for all the benchmark inference. We use a max response length of 32768 tokens, sampling temperature 1, top-p 0.95, top-k 50 for all the model checkpoints.

\subsection{Additional Experimental Results}
\label{app:extended-comparison}

\subsubsection{Throughput of \model}
\begin{table}[h]
\centering
\scriptsize
\setlength{\tabcolsep}{14pt}
\caption{\textbf{Fixed-output throughput (tokens/s).}  Measured on
1$\times$A100 80GB (bf16, SGLang serving stack, \texttt{max\_new\_tokens}
$=2048$, \texttt{ignore\_eos}$=$true, temperature $1.0$, top-$p$
$0.95$, top-$k$ $50$).  Columns give the concurrency level $C$.
Results other than \model are run under each model's recommended
decoding configuration.}
\label{tab:flare-tps}
\begin{tabular}{l|ccc|ccc|ccc}
\toprule
 & \multicolumn{3}{c|}{\textbf{GSM8K}} & \multicolumn{3}{c|}{\textbf{HumanEval}} & \multicolumn{3}{c}{\textbf{GPQA Diamond}} \\
\textbf{Model} & $C{=}1$ & $C{=}4$ & $C{=}8$ & $C{=}1$ & $C{=}4$ & $C{=}8$ & $C{=}1$ & $C{=}4$ & $C{=}8$ \\
\midrule
SDAR-1.7B       & 137.0 &  376.6 &  438.3 & 135.6 &  364.7 &  427.0 & 130.4 & 383.1 & 461.5 \\
SDAR-4B         & 102.8 &  327.3 &  426.9 & 105.1 &  326.1 &  398.3 &  95.7 & 313.6 & 441.7 \\
SDAR-8B         &  83.0 &  267.5 &  388.9 &  83.8 &  276.8 &  372.1 &  72.2 & 229.6 & 379.3 \\
SDAR-30B-A3B    &  43.9 &  124.6 &  216.4 &  39.8 &  122.3 &  214.1 &  43.2 & 120.6 & 210.2 \\
LLaDA-2.0-mini  & 185.6 &  408.7 &  518.7 & 234.2 &  496.7 &  626.1 & 182.6 & 264.2 & 373.2 \\
LLaDA-2.1-mini  & 524.5 & 1224.1 &  963.0 & 472.4 & 1471.4 & 1646.5 & 226.9 & 283.9 & 401.2 \\
\midrule
\model-2B & 487.0 & 1314.7 & 2087.0 & 373.5 & 1113.3 & 1763.9 & 330.7 & 949.0 & 1440.8 \\
\model-4B & 292.1 &  868.7 & 1293.2 & 268.2 &  812.2 & 1178.9 & 237.4 & 649.3 &  990.4 \\
\model-9B & 217.4 &  696.1 & 1096.5 & 203.2 &  642.8 & 1007.1 & 170.8 & 499.0 &  786.5 \\
\bottomrule
\end{tabular}
\end{table}

Table~\ref{tab:app-lsft-family} reports the full-family (2B/4B/9B) per-benchmark comparison of the released Qwen3.5 checkpoints, AR-SFT on the \texttt{Long-CoT+Math+IF} mix, and \model on the same mix, providing the source-model and AR-SFT references behind the main-text capability discussion in Section~\ref{sec:flare}.

\begin{table}[h]
\centering
\scriptsize
\setlength{\tabcolsep}{2.5pt}
\caption{\textbf{Full-family FLARE comparison on the
\texttt{Long-CoT+Math+IF} transfer mix.} \textsuperscript{*}: potentially under-reported due to answer extraction or length limits, consistent with the markers in Table~\ref{tab:flare-small}.}
\label{tab:app-lsft-family}
\resizebox{\linewidth}{!}{%
\begin{tabular}{lccccc|cccc|ccc}
\toprule
 & \multicolumn{5}{c|}{\textit{Knowledge \& Instruction Following}} & \multicolumn{4}{c|}{\textit{Math}} & \multicolumn{3}{c}{\textit{Code}} \\
\textbf{Model} & ARC-C & MMLU & MMLU-Pro & GPQA-D & IFEval & GSM8K & MATH-500 & AIME-24 & AIME-25 & HumanEval & MBPP & LCBv6 \\
\midrule
Qwen3.5-2B (released)         & 92.15 & 73.59 & 59.53 & 62.12 & 79.48 & 77.63\textsuperscript{*} & 72.20\textsuperscript{*} & 8.89\textsuperscript{*}  & 12.22\textsuperscript{*} & 48.17 & 53.31 & 17.71 \\
\quad + AR-SFT                & 88.74 & 68.63 & 56.67 & 47.98 & 67.65          & 83.93          & 85.80 & 27.78          & 22.22          & 67.07 & 71.60 & 21.14 \\
\model-2B                     & 85.07          & 67.60          & 53.57          & 37.37          & 68.95 & 84.46 & 84.40          & 31.11 & 26.67 & 64.02          & 68.09          & 15.43 \\
\midrule
Qwen3.5-4B (released)         & 96.33 & 85.22 & 77.88 & 80.30 & 90.02 & 89.16 & 95.40 & 63.33 & 48.89 & 87.80 & 82.49 & 50.86 \\
\quad + AR-SFT                & 93.86 & 80.75 & 71.70 & 61.11          & 72.46          & 91.66 & 94.40 & 62.22 & 46.67 & 92.68          & 91.83 & 45.71 \\
\model-4B                     & 93.52          & 78.73          & 71.14          & 63.64 & 73.20 & 91.05          & 94.20          & 58.89          & 43.33          & 93.29 & 89.11          & 41.71 \\
\midrule
Qwen3.5-9B (released)         & 97.70 & 88.21 & 81.39 & 80.30 & 91.31 & 89.16 & 96.60 & 65.56 & 60.00 & 95.12 & 89.11 & 49.71 \\
\quad + AR-SFT                & 96.25          & 86.07 & 77.66 & 71.21 & 76.52 & 94.01 & 95.00          & 66.67 & 61.11 & 96.34 & 93.77 & 46.29 \\
\model-9B                     & 96.33 & 84.80          & 77.39          & 71.21 & 71.35          & 93.33          & 95.20 & 63.33          & 54.44          & 92.07          & 91.05          & 49.71 \\
\bottomrule
\end{tabular}
}
\end{table}

\subsubsection{Study Summary Results}
\label{app:aggregate-study}

\begin{table}[h]
\centering
\footnotesize
\setlength{\tabcolsep}{5pt}
\caption{Ablation on the algorithmic design space. Rows (b)--(e) add one ingredient at a time: (a) AR next-token finetune baseline; (b) block diffusion with block-causal clean stream and no clean-stream loss; (c) adds a token-causal clean stream attention mask on top of (b); (d) adds the clean-stream NTP loss; (e) adds logit-shift on the noisy stream (\model setting). All rows share the same Qwen3-1.7B seed and the \texttt{Long-CoT} data.}
\label{tab:study-ingredients}
\begin{tabular}{lccc}
\toprule
Setting & Math + Reasoning & Knowledge + IF & Code \\
\midrule
(a) AR-SFT               & 51.93 & 57.29 & 54.81 \\
(b) Block Diffusion      & 40.98 & 31.29 & 26.47 \\
(c) \;\;+ causal context & 43.09 & 48.42 & 49.19 \\
(d) \;\;+ NTP loss       & 52.70 & 54.53 & 51.41 \\
(e) \;\;+ logit shift (\model) & 51.00 & 56.91 & 51.93 \\
\bottomrule
\end{tabular}
\end{table}

\begin{table}[h]
\centering
\footnotesize
\setlength{\tabcolsep}{6pt}
\caption{Training mask sampling distribution and logit-shift ablation on \texttt{Long-CoT}, built on top of (c)+(d) of Table~\ref{tab:study-ingredients}. All-masked (AM) represents diffusion loss computed with deterministic pure mask sampling; Diffusion (D) represents diffusion loss computed with random mask sampling. Per-benchmark numbers are in Table~\ref{tab:app-shift}.}
\label{tab:study-shift}
\begin{tabular}{llccc}
\toprule
Noisy loss & Shift & Math + Reasoning & Knowledge + IF & Code \\
\midrule
All-masked (AM) & \xmark & 52.83 & 55.67 & 51.59 \\
All-masked (AM) & \cmark & 52.05 & 56.44 & 52.45 \\
\midrule
Diffusion (D)   & \xmark & 52.70 & 54.53 & 51.41 \\
Diffusion (D; full \model) & \cmark & 51.00 & 56.91 & 51.93 \\
\bottomrule
\end{tabular}
\end{table}

\begin{table}[h]
\centering
\footnotesize
\setlength{\tabcolsep}{5pt}
\caption{Ablation of data recipe. Methods: \textbf{AR-SFT}, continued finetuning in standard AR fashion; \textbf{AM-Causal}, all-mask noisy stream with pure causal attention both for clean stream and within noisy blocks; \textbf{AM-Bidir}, all-mask noisy stream with pure causal attention for clean stream and bidirectional attention within noisy blocks; \model. Per-benchmark numbers are in Table~\ref{tab:app-data-full}.}
\label{tab:study-data}
\begin{tabular}{llccc}
\toprule
Data mix & Method & Math + Reasoning & Knowledge + IF & Code \\
\midrule
(reference) & Qwen3-1.7B & 64.70 & 66.74 & 65.60 \\
\midrule
\multirow{4}{*}{\texttt{Long-CoT}}
  & AR-SFT    & 51.93 & 57.29 & 54.81 \\
  & AM-Causal & 50.81 & 57.57 & 53.86 \\
  & AM-Bidir  & 52.05 & 56.44 & 52.45 \\
  & \model    & 51.00 & 56.91 & 51.93 \\
\midrule
\multirow{4}{*}{\texttt{Short-CoT+Math}}
  & AR-SFT    & 48.64 & 57.25 & 50.69 \\
  & AM-Causal & 54.71 & 56.35 & 33.99 \\
  & AM-Bidir  & 54.30 & 55.92 & 36.64 \\
  & \model    & 54.03 & 56.70 & 41.17 \\
\midrule
\multirow{4}{*}{\texttt{Long-CoT+Math}}
  & AR-SFT    & 52.55 & 55.84 & 53.62 \\
  & AM-Causal & 53.23 & 54.73 & 51.16 \\
  & AM-Bidir  & 51.80 & 54.95 & 51.91 \\
  & \model    & 51.48 & 55.54 & 54.17 \\
\midrule
\multirow{4}{*}{\texttt{Long-CoT+Math+IF}}
  & AR-SFT    & 53.31 & 59.35 & 54.43 \\
  & AM-Causal & 47.88 & 57.57 & 50.35 \\
  & AM-Bidir  & 49.09 & 59.14 & 52.06 \\
  & \model    & 53.09 & 59.19 & 52.66 \\
\bottomrule
\end{tabular}
\end{table}

\subsubsection{Per-benchmark Detailed Results of Main Tables}
\label{app:per-benchmark-study}
In this section, we include the per-benchmark score of each model we trained that is included in the main study and depicted in
Tables~\ref{tab:study-shift}--\ref{tab:study-data}. Every non-AR checkpoint is decoded under the AR-Trust sampling path through speculative decoding unless it's non-applicable. 
mode (clean-stream trust path); AR baselines use native AR
decoding.  Benchmark columns group by capability (Knowledge \&
Instruction Following, Math, Code).

\begin{table}[h]
\centering
\scriptsize
\setlength{\tabcolsep}{3.2pt}
\caption{Per-benchmark numbers behind Table~\ref{tab:study-shift}}
\label{tab:app-shift}
\resizebox{\linewidth}{!}{%
\begin{tabular}{ll|ccccc|cccc|ccc}
\toprule
Loss & Shift & ARC-C & MMLU & MMLU-Pro & GPQA & IFEval & GSM8K & MATH500 & AIME24 & AIME25 & HumanEval & MBPP & LCBv6 \\
\midrule
AM & \xmark & 80.89 & 62.23 & 45.63 & 30.30 & 59.15 & 83.78 & 84.20 & 16.67 & 21.11 & 70.12 & 66.93 & 17.71 \\
AM & \cmark & 83.02 & 62.62 & 47.37 & 24.75 & 59.33 & 83.09 & 84.80 & 17.78 & 18.89 & 71.95 & 66.54 & 18.86 \\
\midrule
D  & \xmark & 82.59 & 62.63 & 45.88 & 30.81 & 55.08 & 83.24 & 81.80 & 20.00 & 17.78 & 67.68 & 67.70 & 18.86 \\
D  & \cmark & 84.04 & 62.70 & 45.92 & 26.26 & 62.11 & 83.02 & 83.80 & 10.00 & 18.89 & 71.34 & 67.32 & 17.14 \\
\bottomrule
\end{tabular}%
}
\end{table}

\begin{table}[h]
\centering
\scriptsize
\setlength{\tabcolsep}{2.2pt}
\renewcommand{\arraystretch}{1.05}
\caption{Per-benchmark numbers behind Table~\ref{tab:study-data}: the
four transfer data mixes $\times$ four methods, all decoded under
AR-mode.  Each row block is one data mix; within a block, the four
rows are AR-SFT / AM-Causal / AM-Bidir / \model respectively.}
\label{tab:app-data-full}
\resizebox{\linewidth}{!}{%
\begin{tabular}{ll|ccccc|cccc|ccc}
\toprule
Data mix & Method & ARC-C & MMLU & MMLU-Pro & GPQA & IFEval & GSM8K & MATH500 & AIME24 & AIME25 & HumanEval & MBPP & LCBv6 \\
\midrule
\multirow{4}{*}{\makecell[l]{\texttt{Long-}\\\texttt{CoT}}}
 & AR-SFT    & 84.90 & 63.77 & 48.77 & 28.79 & 59.33 & 84.31 & 85.80 & 14.44 & 13.33 & 76.83 & 71.60 & 16.00 \\
 & AM-Causal & 80.89 & 62.83 & 47.39 & 27.78 & 62.48 & 83.40 & 85.00 & 13.33 & 14.44 & 75.00 & 68.87 & 17.71 \\
 & AM-Bidir  & 83.02 & 62.62 & 47.37 & 24.75 & 59.33 & 83.09 & 84.80 & 17.78 & 18.89 & 71.95 & 66.54 & 18.86 \\
 & \model    & 84.04 & 62.70 & 45.92 & 26.26 & 62.11 & 83.02 & 83.80 & 10.00 & 18.89 & 71.34 & 67.32 & 17.14 \\
\midrule
\multirow{4}{*}{\makecell[l]{\texttt{Short-CoT}\\\texttt{+Math}}}
 & AR-SFT    & 81.40 & 63.82 & 51.56 & 31.31 & 56.38 & 83.93 & 85.20 &  4.44 &  5.56 & 71.34 & 70.43 & 10.29 \\
 & AM-Causal & 84.13 & 64.85 & 50.03 & 30.30 & 54.16 & 84.69 & 83.60 & 26.67 & 18.89 & 42.07 & 49.03 & 10.86 \\
 & AM-Bidir  & 84.73 & 64.31 & 50.76 & 28.28 & 52.68 & 84.15 & 84.20 & 21.11 & 23.33 & 46.34 & 52.14 & 11.43 \\
 & \model    & 82.94 & 64.36 & 50.28 & 31.82 & 55.45 & 83.40 & 83.80 & 22.22 & 20.00 & 54.88 & 57.20 & 11.43 \\
\midrule
\multirow{4}{*}{\makecell[l]{\texttt{Long-CoT}\\\texttt{+Math}}}
 & AR-SFT    & 82.59 & 62.46 & 47.19 & 31.82 & 57.86 & 83.40 & 86.40 & 16.67 & 14.44 & 71.95 & 71.21 & 17.71 \\
 & AM-Causal & 81.48 & 61.07 & 45.82 & 29.80 & 57.30 & 83.78 & 83.20 & 18.89 & 22.22 & 71.34 & 66.15 & 16.00 \\
 & AM-Bidir  & 82.25 & 61.47 & 45.52 & 26.26 & 57.86 & 81.73 & 85.00 & 17.78 & 17.78 & 70.73 & 66.15 & 18.86 \\
 & \model    & 81.14 & 61.57 & 46.10 & 29.29 & 58.96 & 83.93 & 83.40 & 12.22 & 18.89 & 73.78 & 70.43 & 18.29 \\
\midrule
\multirow{4}{*}{\makecell[l]{\texttt{Long-CoT}\\\texttt{+Math+IF}}}
 & AR-SFT    & 82.76 & 62.91 & 48.40 & 31.82 & 66.73 & 76.72 & 83.00 & 22.22 & 23.33 & 71.95 & 68.48 & 22.86 \\
 & AM-Causal & 82.25 & 62.75 & 45.45 & 24.24 & 64.51 & 75.82 & 81.60 &  6.67 & 16.67 & 70.12 & 62.65 & 18.29 \\
 & AM-Bidir  & 83.02 & 63.22 & 47.29 & 21.21 & 66.91 & 79.00 & 82.40 & 13.33 & 15.56 & 71.34 & 66.54 & 18.29 \\
 & \model    & 83.19 & 64.21 & 49.58 & 30.30 & 63.77 & 83.55 & 84.80 & 18.89 & 17.78 & 69.51 & 68.48 & 20.00 \\
\bottomrule
\end{tabular}%
}
\end{table}

\end{document}